\documentclass{sjtudenglab}
\usepackage{amsmath}
\usepackage{enumerate} 
\usepackage{algorithm}
\usepackage{algpseudocode}
\usepackage{amsfonts}
\usepackage{amsthm}
\usepackage{newtxtt}
\usepackage{diagbox} 
\usepackage{colortbl}
\usepackage{amssymb}
\usepackage{xspace}
\usepackage{wrapfig}
\usepackage{adjustbox}
\usepackage{tabularx}
\usepackage{booktabs}
\usepackage{mathtools}
\usepackage{tikz}
\usepackage{enumitem}
\usepackage{silence}
\usepackage{dsfont}
\usepackage[table]{xcolor}
\usepackage[dvipsnames]{xcolor}
\usepackage{multirow}
\usepackage{makecell}
\usepackage{xfakebold}
\usepackage{natbib}
\usepackage{siunitx}
\usepackage{cleveref}
\usepackage{thmtools}

\usepackage{amsmath,amsfonts,bm}

\def\eqref#1{equation~\ref{#1}}

\def\1{\bm{1}}

\DeclareMathAlphabet{\mathsfit}{\encodingdefault}{\sfdefault}{m}{sl}
\SetMathAlphabet{\mathsfit}{bold}{\encodingdefault}{\sfdefault}{bx}{n}

\usepackage{parskip}

\usepackage{natbib}
\usepackage{amsfonts,bm}

\usepackage{amsmath, amssymb,mathrsfs, amsthm, mathtools}
\usepackage{xspace}
\usepackage{enumitem}
\usepackage{lipsum}
\usepackage{xpatch}
\usepackage{mathtools}
\usepackage{dsfont}
\usepackage{pgf,tikz}
\usetikzlibrary{positioning,matrix,patterns}
\usepackage{caption}
\usepackage{graphicx}
\usepackage{makecell}
\usepackage{multirow}
\usepackage[mathscr]{euscript}
\definecolor{hanblue}{rgb}{0.27, 0.42, 0.81}
\definecolor{deepred}{HTML}{900C3F}
\definecolor{deepgreen}{HTML}{2F6960}
\hypersetup{
    colorlinks=true,
    linkcolor=hanblue,
    urlcolor=hanblue,
    citecolor=hanblue,
    anchorcolor=hanblue}

\declaretheoremstyle[
  headfont=\sffamily\bfseries,
]{sansserif}

\theoremstyle{sansserif}

\theoremstyle{definition}

\theoremstyle{sansserif}

\theoremstyle{remark}

\DeclarePairedDelimiter\abs{\lvert}{\rvert}%
\DeclarePairedDelimiter\norm{\lVert}{\rVert}%

\makeatletter
\let\oldabs\abs
\def\abs{\@ifstar{\oldabs}{\oldabs*}}
\let\oldnorm\norm
\def\norm{\@ifstar{\oldnorm}{\oldnorm*}}
\makeatother

\definecolor{textgray}{HTML}{6E6E73}
\usetikzlibrary{positioning, calc}
\usetikzlibrary{decorations.pathmorphing}

\makeatletter
\patchcmd{\wrong@fontshape}{\@gobbletwo}{}{}{}
\makeatother
\numberwithin{equation}{section} 
\tcbuselibrary{minted}
\setminted[python]{
  linenos,
  breaklines,
  fontsize=\footnotesize,
  xleftmargin=2em
}

\makeatletter
\AtBeginDocument{
  \urlstyle{sf}
  
}
\makeatother

\floatstyle{plain}
\newfloat{algorithmgroup}{tbp}{loag}
\floatname{algorithmgroup}{Algorithm Group}

\renewcommand{\eqref}[1]{\textup{(\ref{#1})}}

\definecolor{light}{RGB}{125, 125, 125}
\crefname{tcb@cnt@pbox}{code}{code}
\Crefname{tcb@cnt@pbox}{Code}{Code}
\crefname{assumption}{assumption}{assumption}
\Crefname{assumption}{Assumption}{Assumptions}

\newtcolorbox[auto counter]{pbox}[2][]{
  colback=white,
  title=Code~\thetcbcounter: #2,
  #1,fonttitle=\sffamily,
  fontupper=\sffamily,
  arc=2pt,
  colframe=bgcolor,
  coltitle=fgcolor,
  colbacktitle=bgcolor,
  toptitle=0.25cm,
  bottomtitle=0.125cm
}

\makeatletter
\newcommand\applefootnote[1]{%
  \begingroup
  \renewcommand\thefootnote{}%
  \renewcommand\@makefntext[1]{\noindent##1}%
  \footnote{#1}%
  \addtocounter{footnote}{-1}%
  \endgroup
}
\makeatother

\definecolor{cverbbg}{gray}{0.90}

\title{AdvDMD: Adversarial Reward Meets DMD For High-Quality Few-Step Generation}

\author[1]{Xu Wang}
\author[2]{Zexian Li} 
\author[2]{Litong Gong}
\author[2]{Tiezheng Ge}
\author[1]{Zhijie Deng}

\affiliation[1]{Shanghai Jiao Tong University}
\affiliation[2]{Alimama Tech}

\abstract{
    Diffusion models offer superior generation quality at the expense of extensive sampling steps. 
    Distillation methods, with Distribution Matching Distillation (DMD) as a popular example, can mitigate this issue, but performance degradation remains pronounced when sampling steps are limited.
    Reinforcement learning (RL) has been leveraged to improve the few-step generation quality during distillation, with the potential to even surpass the performance of the teacher model. However, existing approaches are combinatorial in nature, merely integrating an RL process with the distillation process, which introduces unnecessary complexities. To address this gap, we propose AdvDMD, a method that seamlessly unifies DMD distillation and RL.  
    Specifically, AdvDMD employs the adversarially trained discriminator from DMD2 as the reward model, which assigns low scores to generated images and high scores to real ones. 
    It is trained on both intermediate and final states of the denoising process and updated online with the distilled model, enabling a holistic supervision of the sampling trajectories and mitigating reward hacking.
    We adopt a unified SDE backward simulation and a different training schedule for DMD and RL to enable a more stable and efficient training. 
    Experimental results demonstrate that the 4-step AdvDMD outperforms the original 40-step model for SD3.5 on DPG-Bench, while achieving significant performance gains for SD3 on the GenEval.
    On Qwen-Image, our 2-step AdvDMD achieves superior performance over TwinFlow.

}

\metadata[Code]{\url{{https://github.com/SJTU-DENG-Lab/AdvDMD}}} 
\metadata[Correspondence]{Zhijie Deng: \url{zhijied@sjtu.edu.cn}}
\date{\sffamily\today}

\AtBeginDocument{ 
  \fancyheadoffset[R]{-0.3cm}
  \fancyhead[R]{\raisebox{9pt}{\includegraphics[height=35pt]{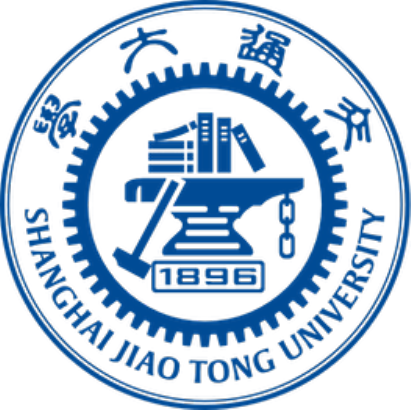}}}
}

\begin{document}

\maketitle
\begin{figure}[H] 
    \begin{minipage}{0.5\textwidth}
    \centering
    \includegraphics[width=.8\textwidth,height=0.5\textwidth]{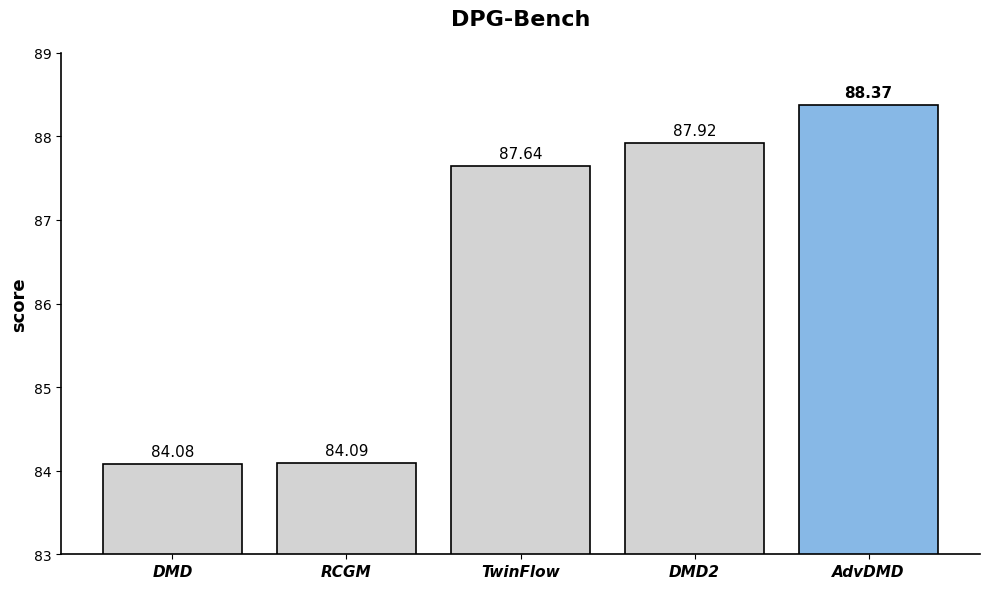}

    \label{fig:trajectory_framework}
    \end{minipage}
    \begin{minipage}{0.5\textwidth}
    \centering
    \includegraphics[width=.8\textwidth,height=0.5\textwidth]{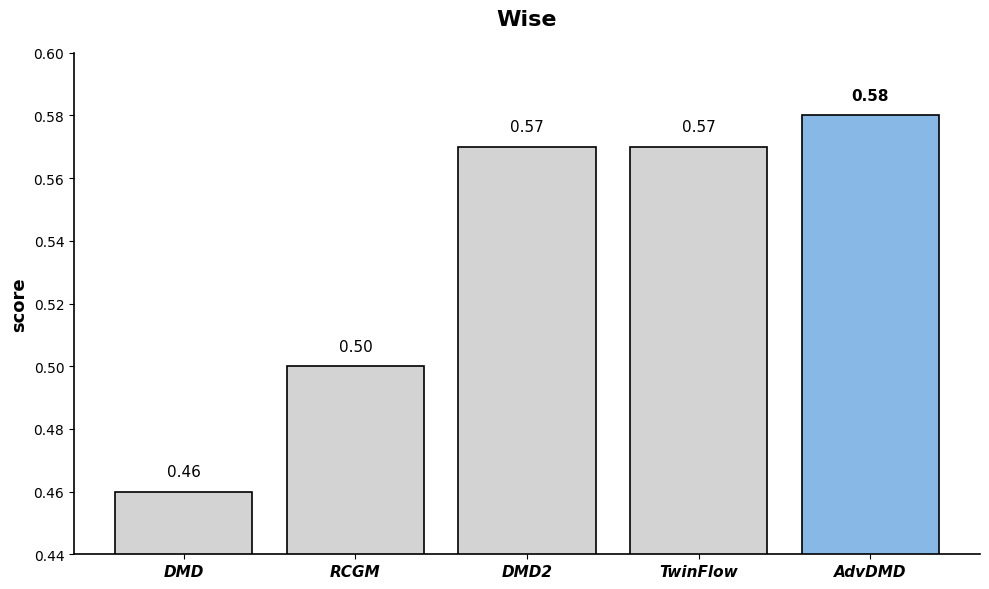}

    \label{fig:trajectory_framework}
    \end{minipage}
    \caption{\textbf{DPG-Bench and Wise score for different methods on Qwen-Image with 2 inference steps.}}
\end{figure}
\section{Introduction}
\label{sec:intro}

\begin{figure}[t!]
    \centering
    \begin{minipage}{0.20\textwidth}
        \includegraphics[width=\linewidth]{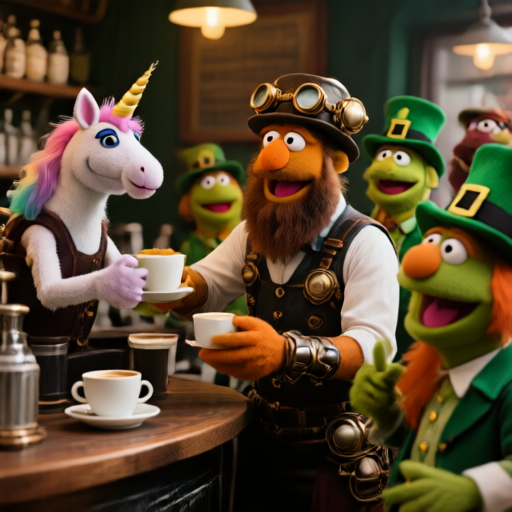}
    \end{minipage}%
    \begin{minipage}{0.20\textwidth}
        \includegraphics[width=\linewidth]{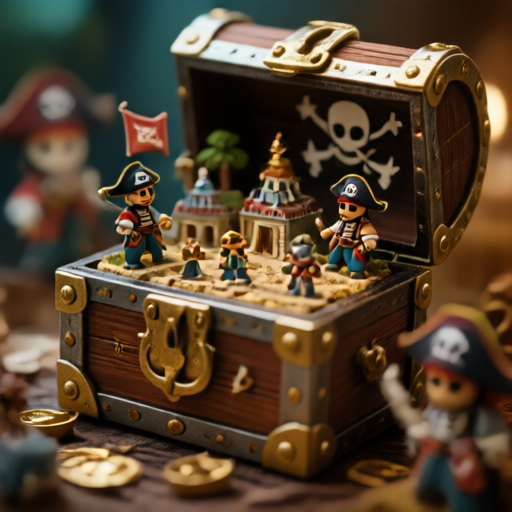}
    \end{minipage}%
    \begin{minipage}{0.20\textwidth}
        \includegraphics[width=\linewidth]{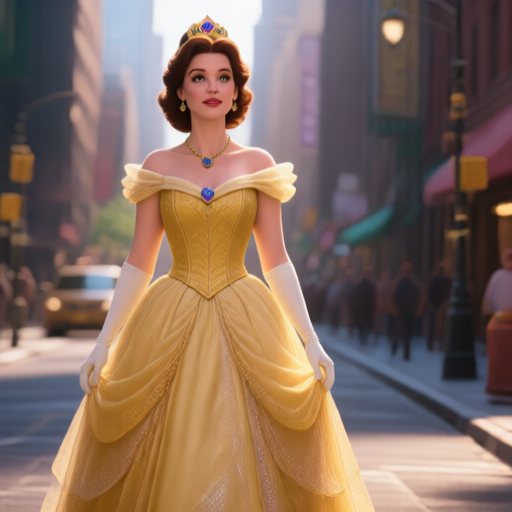}
    \end{minipage}%
    \begin{minipage}{0.20\textwidth}
        \includegraphics[width=\linewidth]{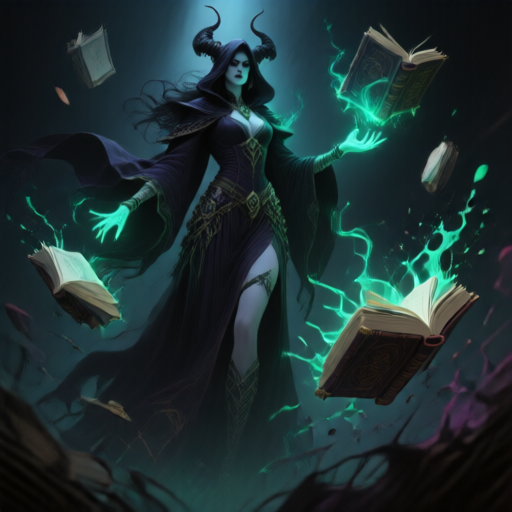}
    \end{minipage}%
    \begin{minipage}{0.20\textwidth}
        \includegraphics[width=\linewidth]{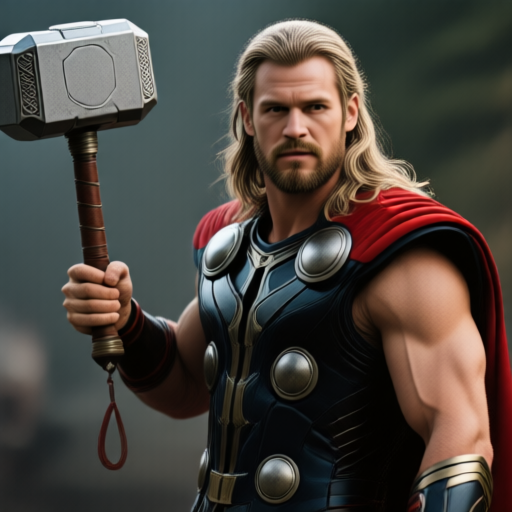}
    \end{minipage}%

    \begin{minipage}{0.2\textwidth}
        \includegraphics[width=\linewidth]{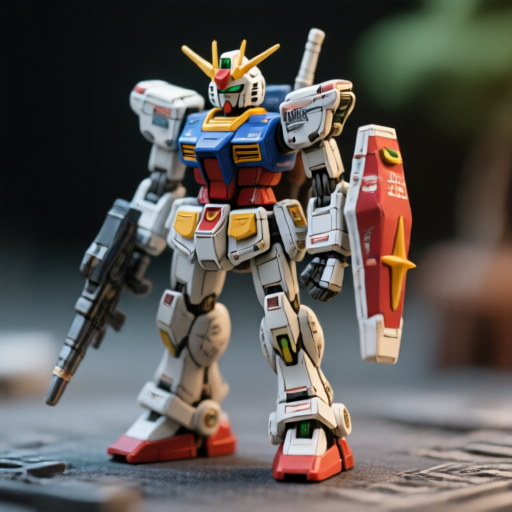}
    \end{minipage}%
    \begin{minipage}{0.2\textwidth}
        \includegraphics[width=\linewidth]{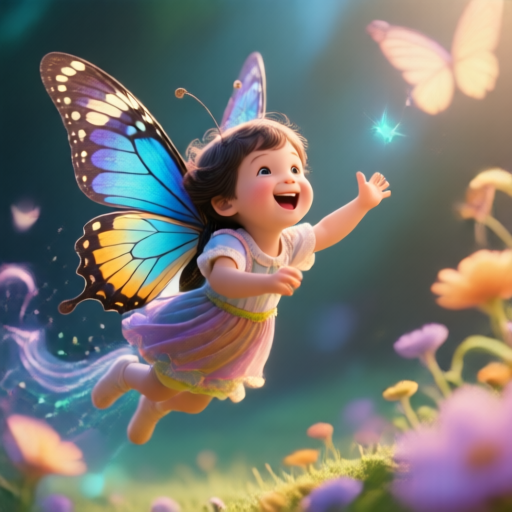}
    \end{minipage}%
    \begin{minipage}{0.2\textwidth}
        \includegraphics[width=\linewidth]{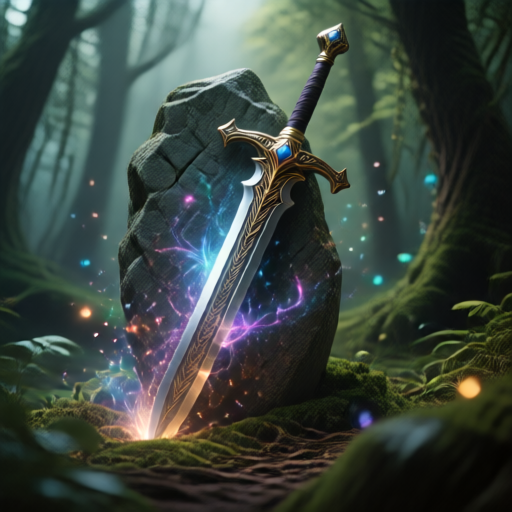}
    \end{minipage}%
    \begin{minipage}{0.2\textwidth}
        \includegraphics[width=\linewidth]{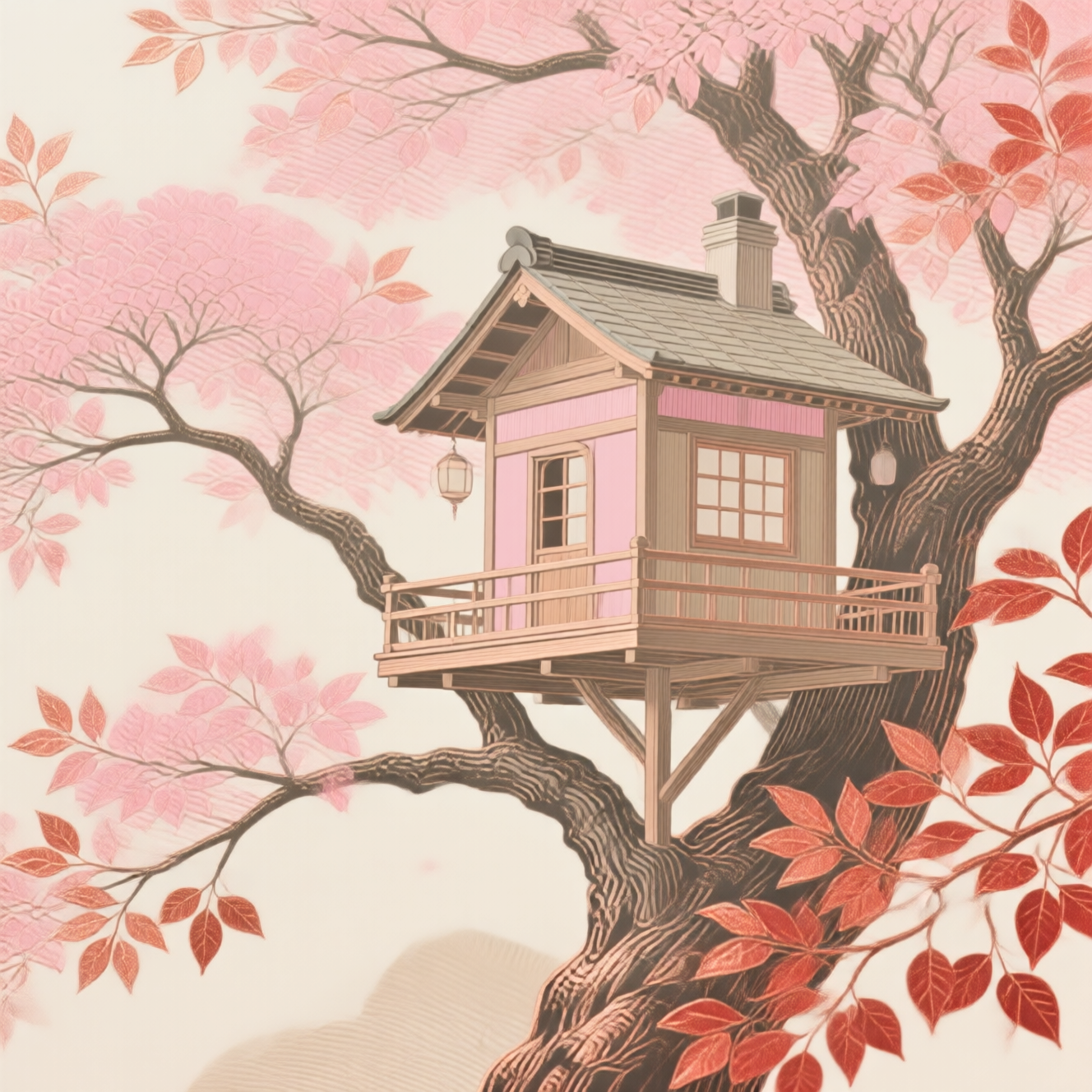}
    \end{minipage}%
    \begin{minipage}{0.20\textwidth}
        \includegraphics[width=\linewidth]{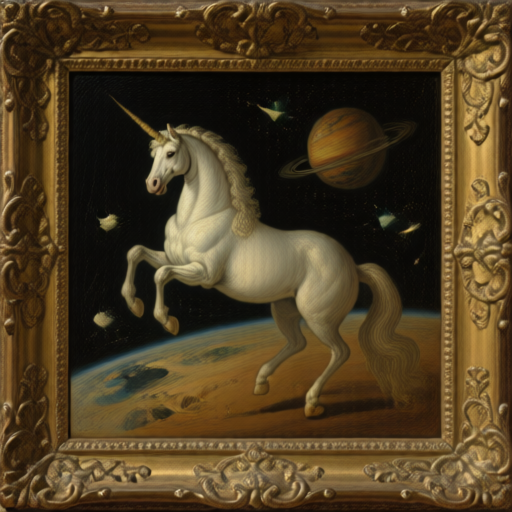}
    \end{minipage}%

    \begin{minipage}{0.2\textwidth}
        \includegraphics[width=\linewidth]{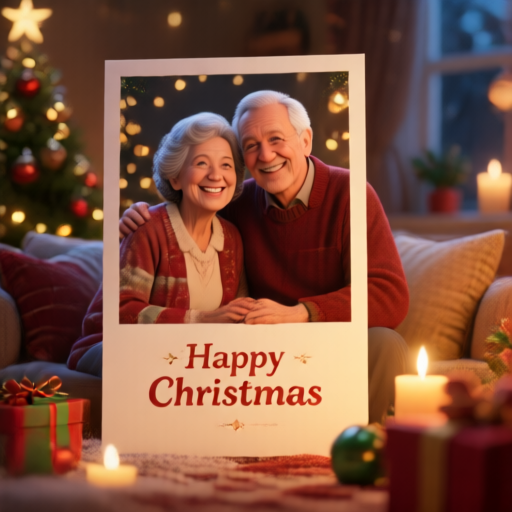}
    \end{minipage}%
    \begin{minipage}{0.2\textwidth}
        \includegraphics[width=\linewidth]{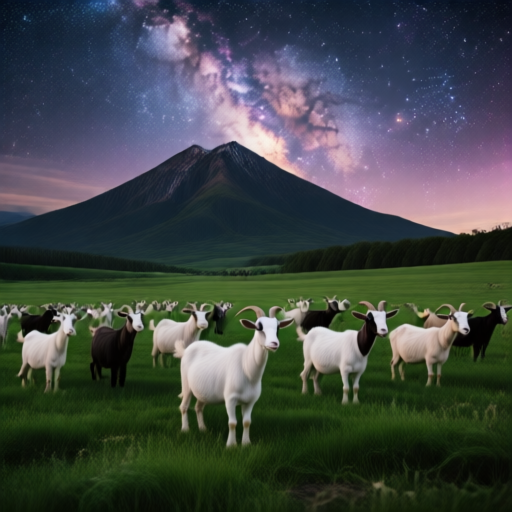}
    \end{minipage}%
    \begin{minipage}{0.2\textwidth}
        \includegraphics[width=\linewidth]{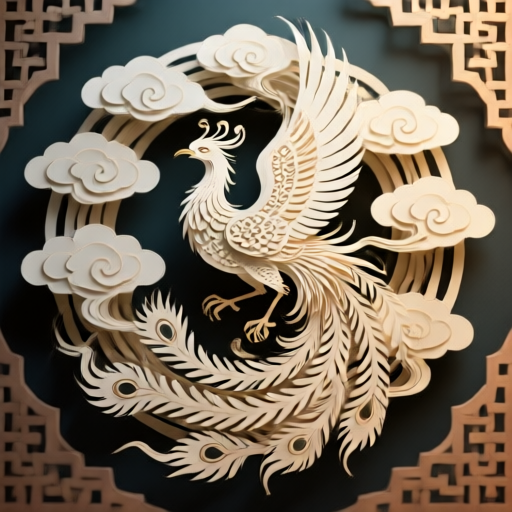}
    \end{minipage}%
    \begin{minipage}{0.2\textwidth}
        \includegraphics[width=\linewidth]{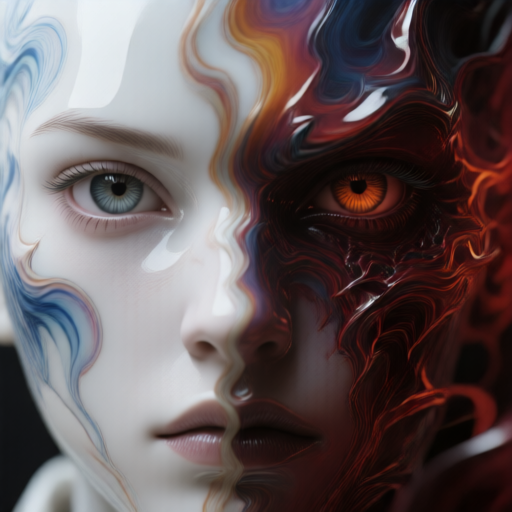}
    \end{minipage}%
    \begin{minipage}{0.20\textwidth}
        \includegraphics[width=\linewidth]{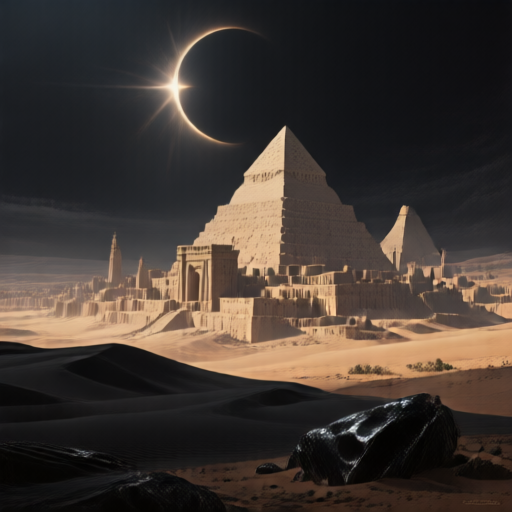}
    \end{minipage}%
    \vspace{0cm}
    \begin{minipage}{0.20\textwidth}
        \includegraphics[width=\linewidth]{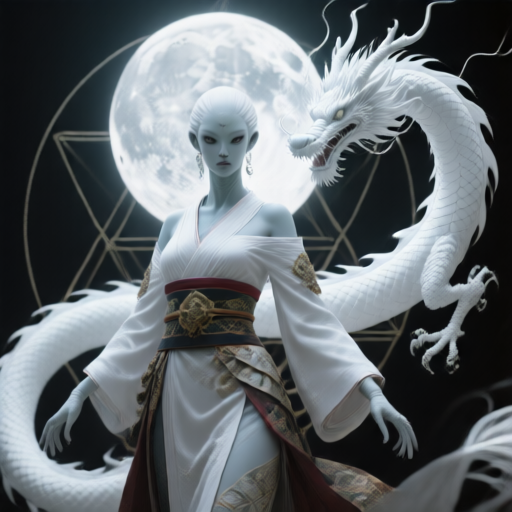}
    \end{minipage}%
    \begin{minipage}{0.20\textwidth}
        \includegraphics[width=\linewidth]{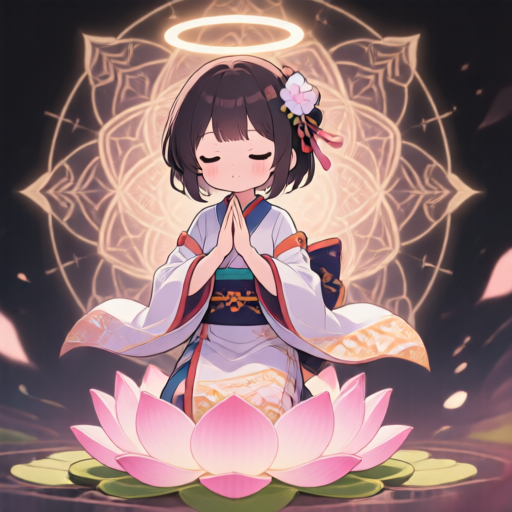}
    \end{minipage}%
    \begin{minipage}{0.2\textwidth}
        \includegraphics[width=\linewidth]{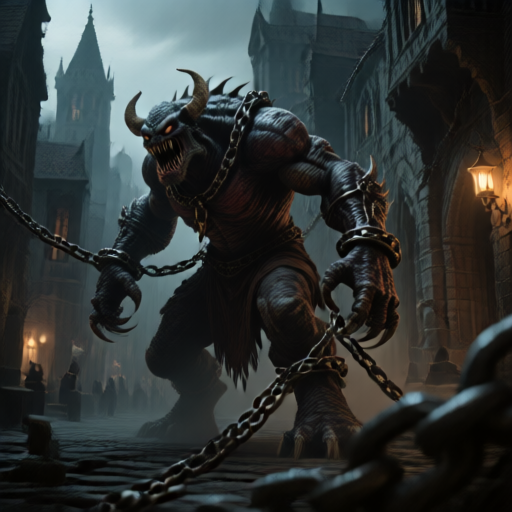}
    \end{minipage}%
    \begin{minipage}{0.2\textwidth}
        \includegraphics[width=\linewidth]{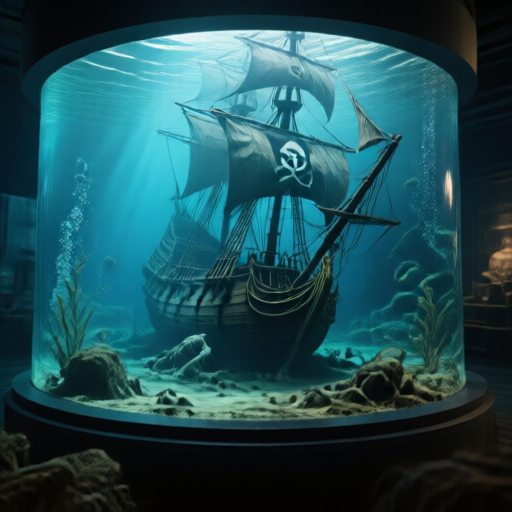}
    \end{minipage}%
        \begin{minipage}{0.20\textwidth}
        \includegraphics[width=\linewidth]{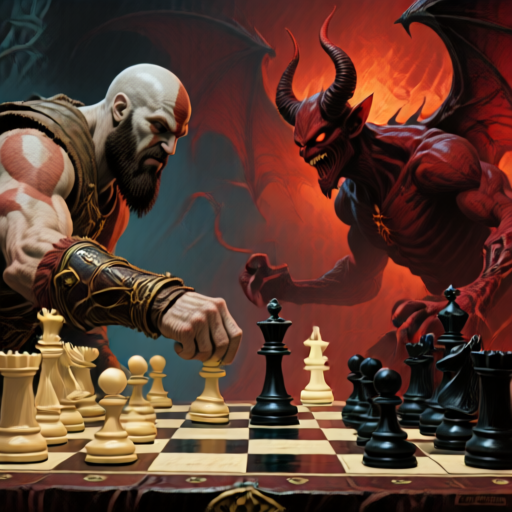}
    \end{minipage}%
\vspace{0cm}
    \begin{minipage}{0.20\textwidth}
        \includegraphics[width=\linewidth,height=\linewidth]{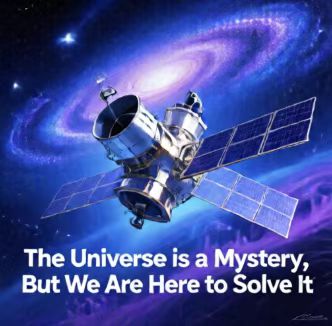}
    \end{minipage}%
    \begin{minipage}{0.20\textwidth}
        \includegraphics[width=\linewidth,height=\linewidth]{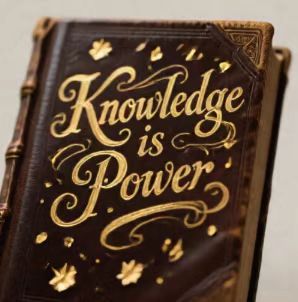}
    \end{minipage}%
    \begin{minipage}{0.2\textwidth}
        \includegraphics[width=\linewidth]{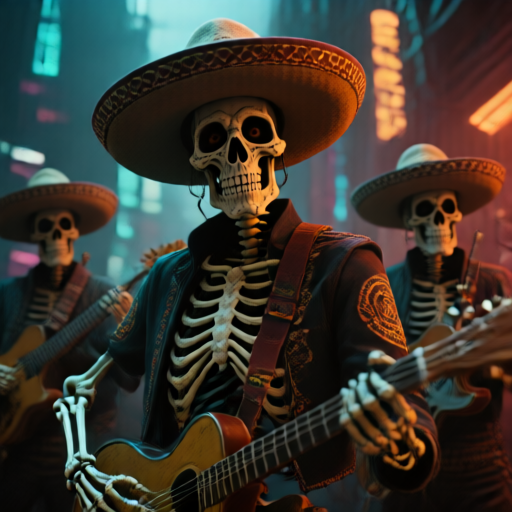}
    \end{minipage}%
    \begin{minipage}{0.2\textwidth}
        \includegraphics[width=\linewidth]{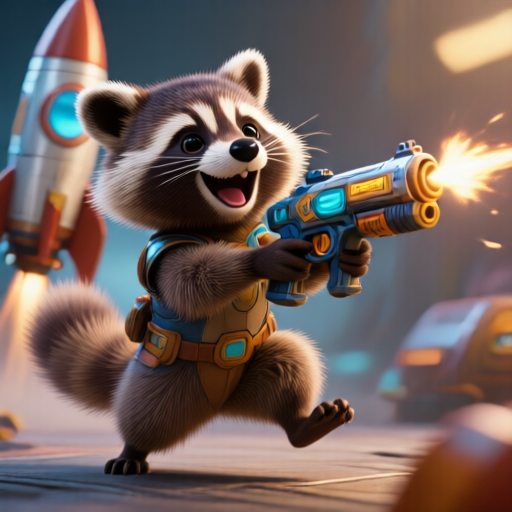}
    \end{minipage}%
        \begin{minipage}{0.20\textwidth}
        \includegraphics[width=\linewidth]{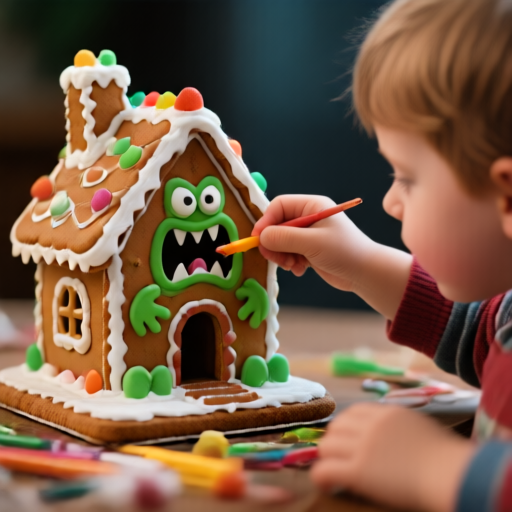}
    \end{minipage}%
    \caption{Visual generations produced by our AdvDMD method under 2 sampling steps on Qwen-Image. All results are obtained without applying classifier-free guidance~\citep{ho2021classifier}.}
    \label{fig:show_qwen}
\end{figure}

Diffusion models have achieved tremendous advancements in generating high fidelity images~\citep{cai2025z,wu2025qwen,labs2025flux1kontextflowmatching,esser2024scaling} and videos~\citep{yang2024cogvideox,wan2025wan,zhang2025waver,gao2025seedance,gao2025wan}.
However, their standard inference procedure typically requires hundreds of denoising steps, leading to low generation efficiency and considerable computational overhead.
Existing acceleration techniques for diffusion models are predominantly based on distillation strategies.
Distribution Matching Distillation (DMD)~\citep{yin2024one,liu2025decoupled,bandyopadhyay2025sd3,yin2024improved} achieves few-step generation by aligning the distribution of the student model with that of a pre-trained teacher model. 
Consistency distillation~\citep{chen2025sana,song2023consistency,zheng2025rcm,lu2024simplifying,luo2023latent,song2023improved,wang2024phased,kim2023consistency} trains the student model to map any two points along the sampling trajectory of a teacher model to a shared terminal point for reduced sampling steps.
However, 
when performing inference with extremely few sampling steps, they often yield inferior performance compared to the original teacher model. 

Reinforcement learning methods~\citep{wang2025coefficients,wang2025grpoguardmitigatingimplicitoveroptimization,zheng2025diffusionnft,black2023training,liu2025improving} like Flow-GRPO~\citep{liu2025flow} and Dance-GRPO~\citep{xue2025dancegrpo} hold the promise to improve generation quality for diffusion models. 
For example, Flow-GRPO samples diverse trajectories via a SDE formulation and leverages external rewards to guide the model toward distributions with higher rewards.
Some works like DMDR~\citep{jiang2025distribution} attempt to directly integrate RL with distillation but require a separate pre-distillation phase and the combination causes substantial complexities. 
Besides, existing methods rely on a fixed reward model, which tends to induce reward hacking --- for instance, HPS~\citep{wu2023human} and PickScore~\citep{kirstain2023pick} favor overly bright images, while OCR-grounded reward models often overemphasize textual content. 

To address these issues, we propose AdvDMD, a novel framework that unifies distillation and reinforcement learning for reliably learning few-step high-quality diffusion models. 
Inspired by Adv-GRPO~\citep{mao2025image} and LRM~\citep{zhang2025diffusion}, we reformulate the original discriminator in DMD2~\citep{yin2024improved} as a reward model within the GRPO training paradigm. 
The reward model is trained on both intermediate and final denoising steps, so as to effectively guide the student model toward more realistic image synthesis. 
It is updated online with the distilled model, thus mitigating reward hacking during training.
Furthermore, to mitigate unreliable rewards caused by insufficient discriminator optimization and inferior generation quality of the few-step generator in early training stages, we employ a decoupled update frequency for DMD and GRPO training --- updating the DMD module multiple times with its own loss between GRPO updates. 
In addition, we replace the ODE-based backward simulation in DMD with an SDE-based formulation to eliminate redundant sampling across DMD and GRPO training, which strengthens distribution alignment and expedites the overall optimization procedure. 
These improvements effectively shorten the training cycle and significantly accelerate the convergence speed while ensuring generation quality.

We conduct comprehensive experiments on SD series~\citep{esser2024scaling} and Qwen-Image~\citep{wu2025qwen}.
We adopt GenEval~\citep{ghosh2023geneval} and DPG-Bench~\citep{hu2024ella} as primary evaluation benchmarks. 
Experimental results validate the efficacy of our method: 4-step AdvDMD achieves a score of 84.65 on DPG-Bench, surpassing both the 4-step DMD2~\citep{yin2024improved} and the original 40-step baseline on SD3.5.
Furthermore, even for the extremely challenging 2-step inference setting, AdvDMD still achieves a competitive score of 84.11 on DPG-Bench on SD3.5, demonstrating substantial performance improvements. 

In summary, our main contributions are as follows:
\begin{itemize}
    \item We propose AdvDMD, a novel unified framework that integrates distillation and RL for few-step generation, where a dynamically updated discriminator serves as the reward model to provide precise guidance for the model. 
    \item  We employ a decoupled update frequency for DMD and GRPO and replace the ODE-based backward simulation with an SDE-based formulation to mitigate unreliable rewards in early training and redundant sampling across DMD and GRPO. 
    \item The 4-step AdvDMD not only outperforms the DMD2, but also achieves superior performance over the original 40-step teacher model method across various evaluation metrics.

\end{itemize}

\section{Related Work}
\label{sec:related}

\noindent \textbf{Distillation for Diffusion Models.}
The distillation of diffusion models~\citep{sauer2024adversarial,sauer2024fast,lu2025adversarial,cheng2025twinflow,xie2024tlcm,heek2024multistep} aims to accelerate sampling by transferring knowledge from a multi-step teacher model to a student model that generates high-quality samples with significantly fewer inference steps.
Distribution Matching Distillation (DMD)~\citep{yin2024one,cheng2025twinflow,yin2024improved} has emerged as a highly promising approach for distilling diffusion models into efficient one-step generators by aligning the output distribution of the student with the teacher, without requiring explicit correspondence at the trajectory level.
DMD2~\citep{yin2024improved} introduces a GAN-style training objective to distinguish between real images and synthesized samples, enabling direct learning from real data and alleviating inaccuracies in score estimation derived from the teacher model.
TwinFlow~\citep{cheng2025twinflow} further reduces memory consumption based on DMD by adopting a self-adversarial objective and using the interval $[-1,0]$ to learn the image generation trajectory. 
This design unifies the entire training process within a single model, substantially reducing the training overhead.
DMDR~\citep{jiang2025distribution} explores the integration of reinforcement learning into DMD by leveraging existing aesthetic preference models, with the aim of further enhancing generation performance.
However, existing distillation methods are often constrained by pretrained teacher models or reward models with fixed, predefined preferences, which limit the generation capability of low-step models.

\noindent \textbf{Reinforcement learning for Diffusion Models.}
Reinforcement learning~\citep{black2023training,wallace2024diffusion,shao2025anchoring,deng2026densegrpo,li2025mixgrpo,wu2025rewarddance,yang2024using} has been widely validated as an effective strategy for enhancing the performance of diffusion models.
Flow-DPO~\citep{liu2025improving} extends direct preference optimization to flow-matching models by directly optimizing the model on paired preference data without explicit reward estimation, thereby enabling stable alignment for diffusion models.
Flow-GRPO~\citep{liu2025flow} extends the GRPO algorithm to the image generation domain by leveraging stochastic differential equation (SDE) sampling to enhance exploration, thereby facilitating group-wise reward computation.
DiffusionNFT~\citep{zheng2025diffusionnft} is an online RL framework that optimizes diffusion models via flow matching on the forward process, using positive-negative generation contrast to enable stable, efficient, and solver-agnostic fine-tuning without likelihood estimation.
However, these approaches presuppose that the model already exhibits sufficiently strong generative capabilities, rendering it ill-suited for direct application to generative models that operate with a limited number of inference steps. 
Moreover, existing reward models often emphasize specific aspects of generation, which can lead to reward hacking, where optimization artificially inflates reward scores without improvement in overall output quality.

\section{Method}

This section presents AdvDMD, a novel framework that unifies denoising diffusion model distillation with reinforcement learning under an adversarial reward mechanism.
We first review existing DMD~\citep{yin2024one,yin2024improved} methods and the representative reinforcement learning method Flow-GRPO~\citep{liu2025flow}. 
We then elaborate on how the discriminator is adapted to function as a reward model within the GRPO framework. 
We further introduce a unified optimization objective and a suite of training strategies, including SDE backward simulation and staged update schedules, to improve convergence speed and training stability.

\subsection{Preliminary: DMD and GRPO }
\noindent \textbf{Distribution Matching Distillation.} DMD has been widely validated in numerous studies~\citep{yin2024one,cheng2025twinflow,yin2024improved,zheng2025rcm,jiang2025distribution,bandyopadhyay2025sd3} as an effective distillation approach for training few-step diffusion models. 
The ideal objective of DMD is to minimize the Kullback–Leibler (KL) divergence $ D_{\mathrm{KL}}(p_{\text{fake}} \parallel p_{\text{real}})$ between the generated image distribution and the real image distribution. 
Direct computation of this divergence is intractable, 
but its gradient can be estimated by:  
\begin{equation}
\label{eq:dmd_loss}
\nabla_{\theta} \mathcal{L}_{\text{dmd}}= \mathbb{E}_{{t\sim [0,1]}} \left[ -\int\left(  s_{\text{real}}(x_t,t) - s_{\text{fake}}(x_t,t) \right) \frac{\partial G_{\theta}(z)}{\partial \theta}dz \right],
\end{equation}
where $z\sim \mathcal{N}(0, I)$, $x_t$ denotes the noisy version of the generated image $x = G_{\theta}(z)$ at timestep $t$, and $s_{\text{real}}(x_t,t)$ and $s_{\text{fake}}(x_t,t)$ represent the score functions of the real and fake data distributions, which can be approximated by a pre-trained teacher diffusion model and an additionally trained diffusion model (referred to as the fake model), respectively.

As the generator’s distribution evolves during training, the fake model is continuously updated via a standard denoising objective:
\begin{equation}  
\label{eq:diff}
\mathcal{L}_{\text{diff}} = \mathbb{E}_{t, z}  \left\| \mu_{\text{fake}}(x_t, t) - x \right\|_2^2 ,
\end{equation}
where $x = G_\theta(z) $ denotes the image generated by the generator and $\mu_{\text{fake}}$ can be easily reparameterized by $s_{\text{real}}$. 

\noindent \textbf{GRPO for Diffusion Models.} Flow-GRPO~\citep{liu2025flow} is a reinforcement learning framework devised to align flow-matching generative models with task-specific objectives or human preferences via online policy gradient optimization.
It formulates the deterministic flow trajectory, usually governed by an ordinary differential equation (ODE), as a stochastic process by transforming it into an equivalent stochastic differential equation (SDE) that preserves marginal distributions across all timesteps:
\begin{equation}
\label{eq:sde}
d\mathbf{x}_t = \left[ \mathbf{v}_\theta(\mathbf{x}_t, t) + \frac{\sigma_t^2}{2t} \left( \mathbf{x}_t + (1-t)\hat{\mathbf{v}}_\theta(\mathbf{x}_t, t) \right) \right] dt + \sigma_t \sqrt{dt}\ {\epsilon},
\end{equation}
where $ \epsilon \sim \mathcal{N}(0,I) $,  $\sigma_t=\eta\sqrt{\frac{t}{1-t}}$, and $ \eta $ controls the stochastic exploration strength.
The generative process is formulated as a Markov Decision Process (MDP), in which the policy $ \pi_\theta(\cdot \mid x_t, c) $ corresponds to the conditional flow distribution $ p_\theta(x_{t-1} \mid x_t, c) $, where $x_t$ denotes the noisy latent state at timestep $t$ and $ c $ represents a conditioning signal such as a text prompt.
During training, a batch of $G$ samples $ \{x_0^i\}_{i=1}^G $ is generated from the current policy $ \pi_{\theta_{\text{old}}}$, and each sample is assigned a scalar reward $R(x_0^i, c)$.
The group-wise advantage is then computed via intra-group normalization:  
\begin{equation}
    \label{eq:grpo_advantage}
\hat{A}^i_t = \frac{R(x_0^i, c) - \operatorname{mean}(\{R(x_0^j, c)\}_{j=1}^G)}{\operatorname{std}(\{R(x_0^j, c)\}_{j=1}^G)}.
\end{equation}
The policy is updated by optimizing the following objective:
\begin{equation}  
\label{eq:grpo}
\mathcal{L}_{\text{grpo}} = \frac{1}{G} \sum_{i=1}^G \frac{1}{T} \sum_{t=0}^{T-1} \min\left( r_t^i(\theta) \hat{A}^i_t,\; \operatorname{clip}(r_t^i(\theta), 1-\epsilon, 1+\epsilon) \hat{A}^i_t \right) - \beta D_{\mathrm{KL}}( \pi_\theta \,\|\, \pi_{\text{ref}}) ,
\end{equation}
where $ r_t^i(\theta) = p_\theta(x_{t-1}^i \mid x_t^i, c) / p_{\theta_{\text{old}}}(x_{t-1}^i \mid x_t^i, c) $ denotes the importance ratio, $ \epsilon $ controls the clipping threshold to constrain the policy update magnitude, and $ \beta $ is the coefficient that weights the KL regularization term.
\label{sec:method}
\subsection{From Discriminator to Adversarial Reward}
We aim to combine DMD~\citep{cheng2025twinflow,liu2025decoupled,zheng2025rcm} and GRPO~\citep{deng2026densegrpo,wang2025coefficients,liu2025flow} to achieve high-quality few-step generation, 
where the design of the reward model is critical. 
However, existing reward models often exhibit biases toward specific attributes and struggle to deliver the holistic improvements typically achieved by distillation. 
We observe that the GAN-style objective in DMD2~\citep{yin2024improved}, where a discriminator is trained to distinguish between generated and real images, effectively guides the generator toward the reference data distribution, thereby preventing mode collapse or deviation from the true data manifold. 
This insight motivates our use of the discriminator as the basis for a more robust and distribution-aware reward signal in the RL framework.
Instead of employing a pixel-level discriminator as in Adv-GRPO~\citep{mao2025image}, we extend the discriminator to operate in the latent space of noisy image states~\citep{sauer2024fast,Lin2025DiffusionAP,mi2025video,zhang2025diffusion}, enabling it to assess generation quality at a more semantically meaningful and computationally efficient level.

\noindent \textbf{Architecture.}
Pretrained diffusion models have been demonstrated to serve as effective feature extractors for discriminators~\citep{Lin2025DiffusionAP,mi2025video,zhang2025diffusion}, providing rich and structured representations that capture both low-level details and high-level semantics of images.
Following prior work~\citep{yin2024improved}, we adopt the pretrained diffusion model $ \mu_{\text{fake}} $ as the backbone feature extractor for the discriminator, which is followed by lightweight trainable convolutional heads that produce realism predictions.
Specifically, given an input noisy image $x_t$, the diffusion-based backbone $ \mu_{\text{fake}} $ yields multi-scale spatial feature maps from different layers. 
These features are then fed into $ K $ independent convolutional heads $ \{h^{k}_\phi(\cdot)\}_{k=1}^K $, each generating a local binary prediction of realism. 
The final discriminator output is computed as the average of all head predictions, providing a stable and semantically informed signal for guiding generator learning:
\begin{equation}
\label{eq:discriminator}
D_\phi(x_t,c) =  \frac{1}{K} \sum_{k=1}^K \sigma (h^{k}_\phi(\mu_{\text{fake}}^{i}(x_t,t,c))) ,
\end{equation}
where $ \sigma(\cdot) $ denotes the sigmoid function, $\mu_{\text{fake}}^{(i)}$ denotes the output of the $i$-th layer of the fake model, 
$ \phi $ encompasses the parameters of the convolutional heads, while the backbone is kept frozen, and $c$ presents the conditioning signal, such as a text prompt or class label, which guides the generation process. 
This design improves training stability and generalization by aggregating multi-scale realism judgments.

\noindent \textbf{Training.}
The discriminator is trained by minimizing the standard adversarial loss:
\begin{equation}
\label{eq:dis}
\mathcal{L}_{\text{dis}} =\mathbb{E}_{x \sim p_{\text{real}},y \sim p_{\text{fake}},t\in[0,1]}[-\log D_\phi(x_t,c) - \log(1 - D_\phi(y_t,c))], 
\end{equation}
where $ p_{\text{real}} $ denotes the distribution of reference images, $ p_{\text{fake}}$ is the images generated by the generator, and $ x_t,y_t $ represents the forward diffusion process that adds noise to an image at timestep $ t \in [0, 1] $. 
The discriminator is continuously trained to distinguish between generated images and reference images, producing a realism score that quantifies the perceived authenticity of the input with respect to the data distribution.
In prior distillation approaches, the generator is optimized to produce samples that maximize the discriminator’s realism score, effectively aiming to "fool" the discriminator by minimizing the following the gan loss:
\begin{equation}
\label{eq:gan}
 \mathcal{L}_{\text{gan}} =  \mathbb{E}_{y \sim p_{\text{fake}}, t\in[0,1] } \big[ -\log D_\phi(y_t,c) \big],   
\end{equation}
where $p_{\text{fake}}$ denotes the distribution of images generated by the model $G_\theta$.

Conventional GANs optimize in a pointwise manner, ignoring the collective quality of conditional trajectories.
We want to extend this to the group level, steering entire ensembles toward higher realism to achieve more effective generator optimization.
Inspired by Adv-GRPO~\citep{mao2025image}, we employ the discriminator as a reward model, guiding the generator to approach the true data distribution more closely.

\noindent \textbf{Reward.} Instead of assigning the same reward derived from the final generated image to all intermediate denoising steps, we leverage the discriminator’s ability to process noisy inputs and compute distinct rewards $R_{t}^{i}(x_t,c)=D_\phi(x_t,c)$ where c represents the condition for each intermediate step $x_t$. 
Accordingly, following~\cref{eq:grpo_advantage}, the advantage estimation is computed independently at each individual timestep:
\begin{equation}
    \label{eq:grpo_advantage_T}
\hat{A}^i_t = \frac{R(x_t^i, c) - \operatorname{mean}(\{R(x_t^j, c)\}_{j=1}^G)}{\operatorname{std}(\{R(x_t^j, c)\}_{j=1}^G)}.
\end{equation}
This per-timestep reward scheme provides more informative and stable learning signals, facilitating fine-grained policy updates that align intermediate generations with the target data distribution.

\subsection{AdvDMD: Adversarial Reward Meets DMD}
Following the above observations, we adapt the discriminator designed to distinguish between real and generated images and repurpose it as a reward model for GRPO training. 
Without introducing any extra models, we combine the original DMD2~\citep{yin2024improved} distillation framework with the GRPO algorithm, resulting in a new method called AdvDMD. 
AdvDMD leverages adversarial feedback throughout the generation process and leads to stronger few-step diffusion models with improved sample quality and fidelity.
We present a detailed description of the algorithm in the following.

\begin{figure*}[t] 
    \centering
    \includegraphics[width=\textwidth,keepaspectratio]{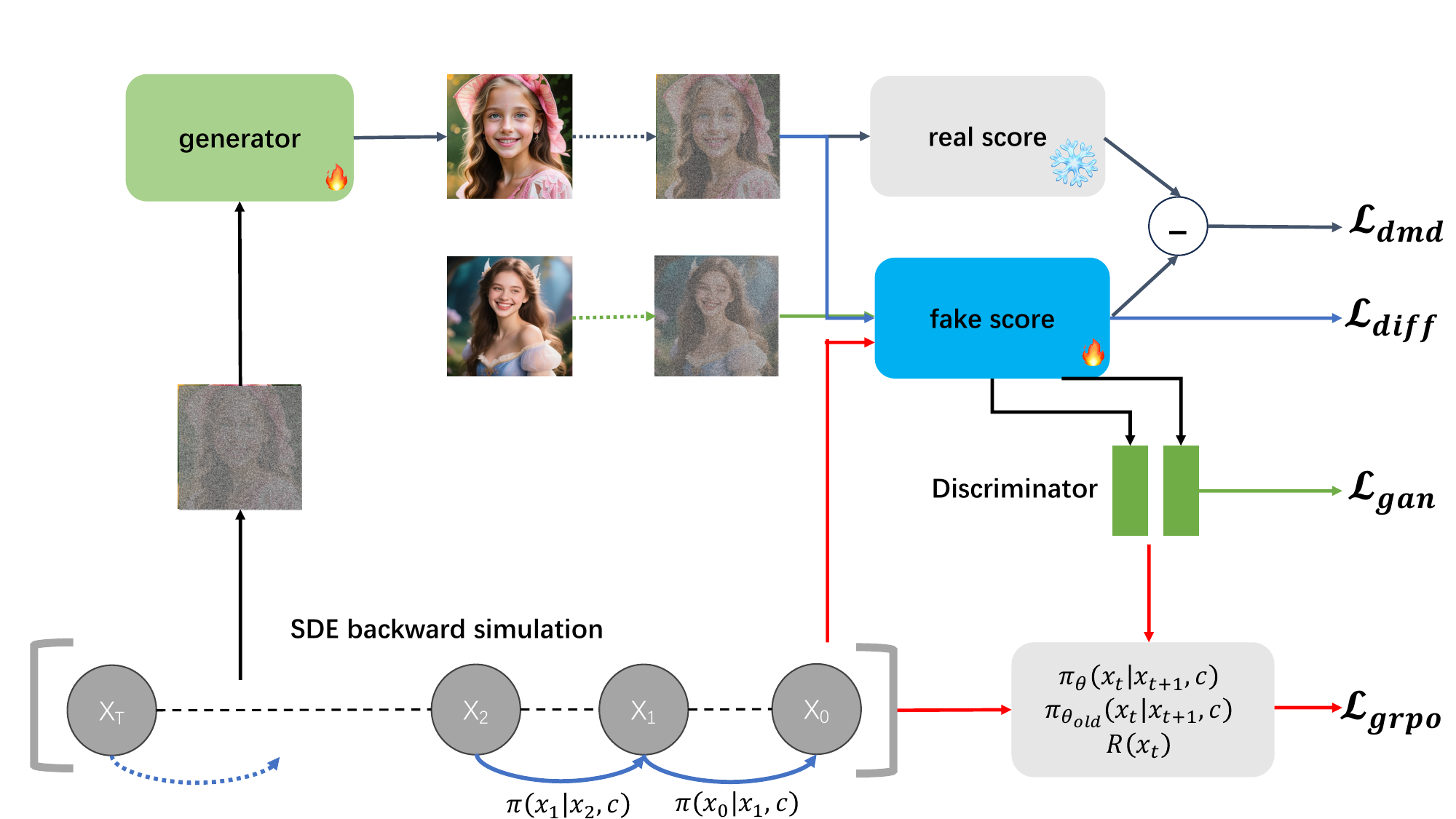}
    \caption{\textbf{Illustration of the AdvDMD Training Pipeline.}The red line illustrates the \textbf{GRPO training pipeline}, where the intermediate state \(x_t\) obtained from the SDE backward
    simulation is fed into the discriminator to yield a corresponding score, which is then used to compute the GRPO loss.}
    \label{fig:trajectory_framework}
\end{figure*}

\noindent \textbf{SDE backward simulation.}
In the DMD training framework, the generator performs an ODE backward simulation to produce noisy images that serve as inputs for subsequent training stages.
Similarly, GRPO requires the model to execute all generation steps sequentially to obtain the final sample and compute its associated reward.
A key difference lies in the underlying dynamics: DMD typically employs deterministic ODE-based sampling, while GRPO utilizes stochastic SDE trajectories to promote exploration during policy updates. 
Notably, we find that substituting the ODE solver with SDE-based sampling in DMD does not degrade performance. 
Leveraging this compatibility, we design a training pipeline in which DMD directly reuses the samples generated during the GRPO sampling phase, eliminating redundant computations and substantially improving training efficiency without compromising generation quality.

\noindent \textbf{Decoupled Update Frequencies.}
The reward model typically requires the generator to possess a minimal level of synthesis competence in order to provide informative feedback. 
Similarly, the discriminator must be updated sufficiently to maintain calibration and remain effective throughout training. 
We observe that reward signals are unstable and poorly calibrated during the phase of training. 
Given the high computational cost of GRPO updates in early stages, we prioritize more frequent optimization of the DMD loss at the outset to rapidly establish baseline generation and discrimination capabilities, deferring full GRPO updates until these components are sufficiently stabilized.

\noindent \textbf{Training Objective.}
For the updates of the fake model and discriminator, we adopt the training configuration of DMD2 and employ the reconstruction loss $ \mathcal{L}_{\text{diff}} $ defined in~\cref{eq:diff} together with the adversarial loss $ \mathcal{L}_{\text{dis}} $ specified in~\cref{eq:dis}.
For the generator, we optimize a composite objective that combines the DMD loss and the GRPO loss. The final training objective for the generator is given by:
\begin{equation}
\label{eq:gen}
     \mathcal{L}_{\text{gen}} = \alpha\mathcal{L}_{\text{dmd}} + \gamma \mathcal{L}_{\text{gan}}+\mathcal{L}_{\text{grpo}}
\end{equation}
where $\alpha$ balances the DMD and GRPO losses to ensure stable optimization, and $\gamma $ controls the strength of the adversarial term $ \mathcal{L}_{\text{gan}} $.
\section{Experiments}
\label{sec:exp}


This section presents the detailed experimental configurations employed to evaluate the effectiveness of the proposed AdvDMD method.
\subsection{Implementation Details}
We train our models using the image–caption pairs provided in Adv-GRPO~\citep{mao2025image}. 
We adopt SD3.5-medium and SD3-medium~\citep{esser2024scaling} as the backbone architectures for the generator. 
The discriminator consists of independent lightweight convolutional heads that operate on multi-layer features extracted from the pretrained diffusion model $ \mu_{\text{fake}} $.
During training, the generator, the fake score model, and the discriminator heads are optimized with learning rates of $5 \times 10^{-6}$, $1 \times 10^{-6}$, and $1 \times 10^{-4}$, respectively.
During the DMD training phase, the real score model employs classifier-free guidance~\citep{ho2021classifier} with a guidance scale of 3.5. 
To stabilize optimization in the early stages of training, we follow the strategy of DMD2~\citep{yin2024improved} and perform five joint updates of the \( \mu_{\text{fake}} \) model and the discriminator for every single generator update.
For the GRPO component, we adopt the configuration from Flow-GRPO~\citep{liu2025flow} with a group size of 32. 
The hyperparameters $ \alpha $ and $\gamma$ are set to $ 1 \times 10^{-1}$ and  $ 1 \times 10^{-2}$, respectively, to balance the adversarial and distillation losses and ensure stable optimization.
Specifically, we train both 4-step and 2-step variants for SD3.5-medium, and a 4-step variant for SD3-medium. 
All models are trained at a resolution of 512 × 512 on 8 NVIDIA H20 GPUs, with a total training duration of approximately 18 hours.
\subsection{Main Results}
\noindent \textbf{Benchmarks.}
We evaluate the effectiveness of AdvDMD on multiple established benchmarks. 
To mitigate the risk of reward hacking, where models might exploit artifacts in simplistic metrics rather than improving genuine output quality, we avoid conventional aesthetic scores such as HPS~\citep{wu2023human} or ClipScore~\citep{hessel2022clipscorereferencefreeevaluationmetric}. 
Instead, we adopt more holistic and behavior-aware evaluation protocols, specifically DPG-Bench~\citep{hu2024ella} and GenEval~\citep{ghosh2023geneval}, which assess generated images across multiple dimensions including visual fidelity, semantic alignment, compositional correctness, and adherence to human preferences. 
This evaluation strategy ensures a more reliable and comprehensive assessment of model performance.

\begin{table*}[t]
    \centering    \small
    \setlength{\tabcolsep}{1pt} 
    \definecolor{lightgray}{RGB}{230,230,230}
    \begin{tabular}{l c cccccc}
        \toprule
        Model & Step & \fontsize{8pt}{9pt}\selectfont Relation $\uparrow$ & \fontsize{8pt}{9pt}\selectfont Other $\uparrow$ & \fontsize{8pt}{9pt}\selectfont Attribute $\uparrow$ & \fontsize{8pt}{9pt}\selectfont Entity $\uparrow$ & \fontsize{8pt}{9pt}\selectfont Global $\uparrow$ & \fontsize{8pt}{9pt}\selectfont Overall $\uparrow$ \\
        \midrule
        \rowcolor{lightgray}
         SD3.5-medium &    &  &  &  &  &  &  \\
        Base (CFG=3.0)   & 40   & 91.38 & \textbf{90.67} & 89.73 & 88.14 & \textbf{92.07} & 84.55 \\
        DMD2*   & 4    & \textbf{91.55} & 87.87 & 88.97 & 86.90 & 86.11 & 82.38 \\
        Turbo & 4    & 85.33 & 84.89 & 84.11 & 84.66 & 81.77 & 74.44 \\
        AdvDMD & 4    & 91.45 & 84.94 & \textbf{90.43} & \textbf{89.86} & 89.82 & \textbf{84.65} \\
        DMD2*    & 2    & 88.59 & 81.46 & 87.76 & \textbf{89.93} & 88.39 & 82.01 \\
        AdvDMD & 2    & \textbf{88.79} & \textbf{87.89} & \textbf{90.33} & 87.92 & \textbf{92.42} & \textbf{84.11} \\
        \midrule
        \rowcolor{lightgray}
         SD3-medium &    &  &  &  &  &  &  \\
        Base (CFG=7.0) & 25    & 80.70 & 88.68 & 88.83 & \textbf{91.01} & 87.90 & 84.08 \\
        DMD2*  & 4    & 89.58 & 87.40 & 89.60 & 89.77 & 88.13 & 83.64 \\
        Hyper-SD (CFG=3.0)  & 4   & \textbf{91.02} & 86.96 & 88.07 & 88.50 & 85.37 & 82.60 \\
        TDM  & 4    & 87.00 & 86.09 & 89.74 & 90.32 & 87.00 & 84.11 \\
        AdvDMD  & 4    & 90.41 & \textbf{89.16} & \textbf{90.32} & 89.67 & \textbf{89.35} & \textbf{84.25} \\

        \bottomrule
    \end{tabular}
    \caption{\textbf{Comparison of DPG-Bench performance across different model variants and steps.}The corresponding best results are highlighted in bold, and * denotes that the method is re-produced by us.}
    \label{tab:dpg_bench_performance_with_step}
\end{table*}



\begin{table*}[t]
    \centering
    \small
    \setlength{\tabcolsep}{1pt} 
     \definecolor{lightgray}{RGB}{230,230,230}
    \begin{tabular}{l c ccccccc}
        \toprule
        \fontsize{8pt}{9pt}\selectfont Model & \fontsize{8pt}{9pt}\selectfont Step & \fontsize{8pt}{9pt}\selectfont Single $\uparrow$ & \fontsize{8pt}{9pt}\selectfont Two  $\uparrow$ & \fontsize{8pt}{9pt}\selectfont Counting $\uparrow$ & \fontsize{8pt}{9pt}\selectfont Colors $\uparrow$ & \fontsize{8pt}{9pt}\selectfont Position $\uparrow$ & \fontsize{8pt}{9pt}\selectfont Attr $\uparrow$ & \fontsize{8pt}{9pt}\selectfont Overall $\uparrow$ \\
        \midrule
         \rowcolor{lightgray}
         SD3.5-medium &    &  &  &  &  &  &  &\\
        Base     & 40   & \textbf{1.0} & 0.84 & 0.64 & \textbf{0.83} & 0.25 & 0.57 & 0.69 \\
        Turbo& 4    & 0.98 & 0.81 & 0.65 & 0.80 & 0.24 & 0.50 & 0.66 \\
        Flash& 4    & - & - & - & - & - & - & \textbf{0.70} \\
        DMD2*  & 4    & 0.98 & 0.81 & 0.65 & 0.80 & 0.24 & 0.50 & 0.66 \\
      
        AdvDMD    & 4    & 0.98 & \textbf{0.90} & \textbf{0.66} & 0.80 & \textbf{0.29} & \textbf{0.58} & \textbf{0.70} \\
      DMD2*       & 2    & 0.98 & 0.83 & 0.58 & 0.79 & \textbf{0.17} & \textbf{0.55} & 0.65 \\
        AdvDMD    & 2    & \textbf{0.99} & \textbf{0.90} & \textbf{0.60} & \textbf{0.84} & 0.16 & 0.53 & \textbf{0.67} \\
        \midrule
         \rowcolor{lightgray}
         SD3-medium &    &  &  &  &  &  & & \\
        Base(CFG=7.0) & 25   & 0.98 & 0.74 & 0.63 & 0.67 & \textbf{0.34} & 0.36 & 0.62 \\
        DMD2*& 4    & 0.98 & 0.82 & 0.58 & 0.80 & 0.18 & 0.52 & 0.64 \\
        Hyper-SD& 4    & 0.99 & 0.74 & 0.55 & 0.83 & 0.18 & 0.45 & 0.62 \\
        TDM & 4    & \textbf{1.00} & 0.85 & \textbf{0.68} & 0.80 & 0.28 & 0.50 & \textbf{0.68} \\
        DMDR&4& 0.99&0.84&0.54&0.81&0.25&0.44 &0.64\\
        AdvDMD& 4    & 0.99 & \textbf{0.87} & 0.55 & \textbf{0.84} & 0.27 & \textbf{0.54} & \textbf{0.68} \\
        \bottomrule
    \end{tabular}
    \vspace{0.2cm}
    \caption{\textbf{Comparison of GenEval performance across different model variants.}
    The corresponding best results are highlighted in bold, and * denotes that the method is reproduced by us.
    }
    \label{tab:geneval_performance}
\end{table*}
\begin{table}[t]
    \centering    \small
    \setlength{\tabcolsep}{1pt} 
    \definecolor{lightgray}{RGB}{230,230,230}
    \begin{tabular}{l c ccc}
        \toprule
        Model & \ \ \  Step & \ \ \ GenEval &\ \ \  DPG-Bench &\ \ \ Wise \\
        \midrule
        \rowcolor{lightgray}
         Qwen-Image &    &  &  &\\
        Base    &\ \ \  50*2  & \ \ \ \textbf{0.87}& \ \ \ 88.32 &\ \ \ \textbf{0.62}\\
        DMD & \ \ \ 2&\ \  \ 0.80&\ \ \  84.08& \ \ \ 0.46  \\
        DMD2*   &\ \ \  2    & \ \ \ \textbf{0.87} & \ \ \ 87.92 & \ \ \ 0.57 \\
        RCGM   &\ \ \ 2 &\ \ \ 0.82&\ \ \  84.09&\ \ \ 0.50 \\
        TwinFlow &\ \ \  2 &\ \ \ \textbf{0.87} & \ \ \ 87.64& \ \ \ 0.57 \\
        AdvDMD &\ \ \  2    & \ \ \ \textbf{0.87} &\ \ \  \textbf{88.37} &\ \ \ 0.58 \\
        \bottomrule
    \end{tabular}
    \caption{\textbf{Comparison of performance across different methods on Qwen-Image.}The corresponding best results are highlighted in bold, and * denotes that the method is re-produced.}
    \label{tab:qwen_performance_with_step}
\end{table}
\begin{figure}[t!]
    \centering
    \begin{minipage}{0.2\textwidth}  
        \centering
        \textbf{SD3}  
    \end{minipage}%
    \begin{minipage}{0.2\textwidth}  
        \centering
        \textbf{TDM}  
    \end{minipage}%
    \begin{minipage}{0.2\textwidth}  
               \centering
        \textbf{DMD2}
    \end{minipage}%
    \begin{minipage}{0.2\textwidth}
                      \centering
        \textbf{Hyper-SD}
    \end{minipage}%
    \begin{minipage}{0.2\textwidth}
           \centering
        \textbf{AdvDMD}
    \end{minipage}%
\vspace{0.1cm}
    \begin{minipage}{0.2\textwidth}
     \centering
        \includegraphics[width=\linewidth]{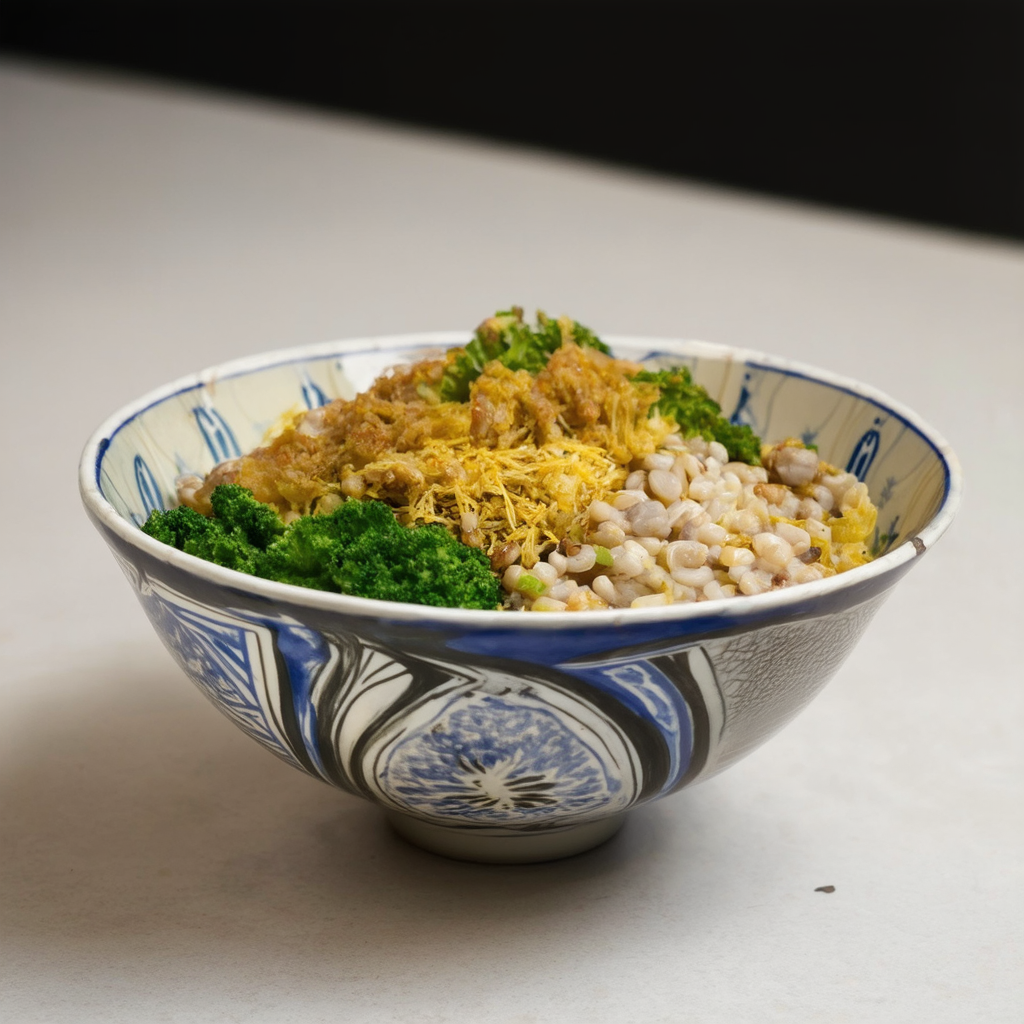}
    \end{minipage}%
    \begin{minipage}{0.2\textwidth}
     \centering
        \includegraphics[width=\linewidth]{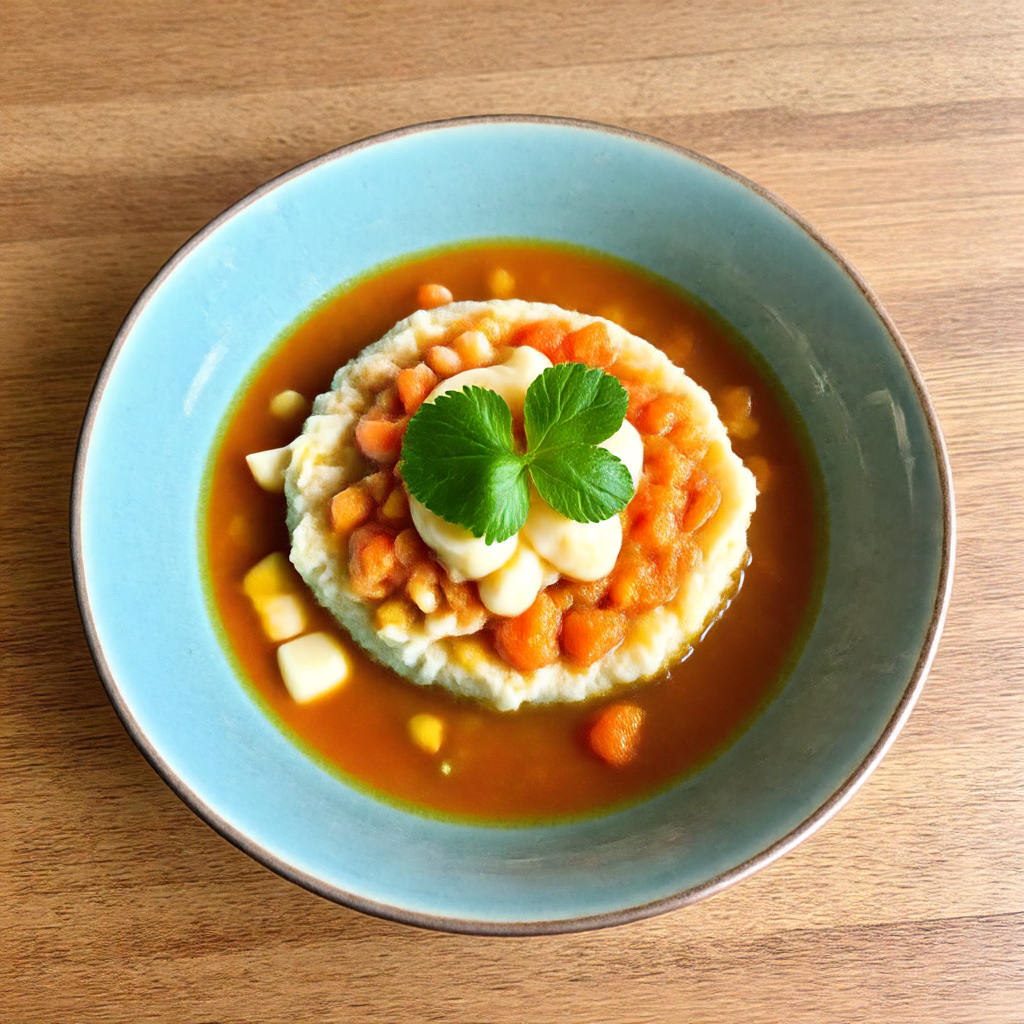}
    \end{minipage}%
    \begin{minipage}{0.2\textwidth}
     \centering
        \includegraphics[width=\linewidth]{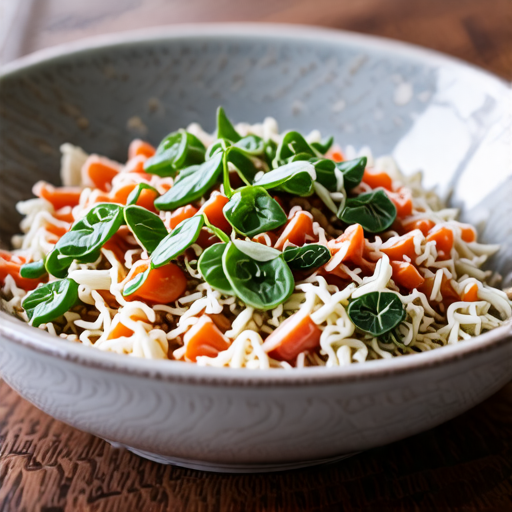}
    \end{minipage}%
    \begin{minipage}{0.2\textwidth}
     \centering
        \includegraphics[width=\linewidth]{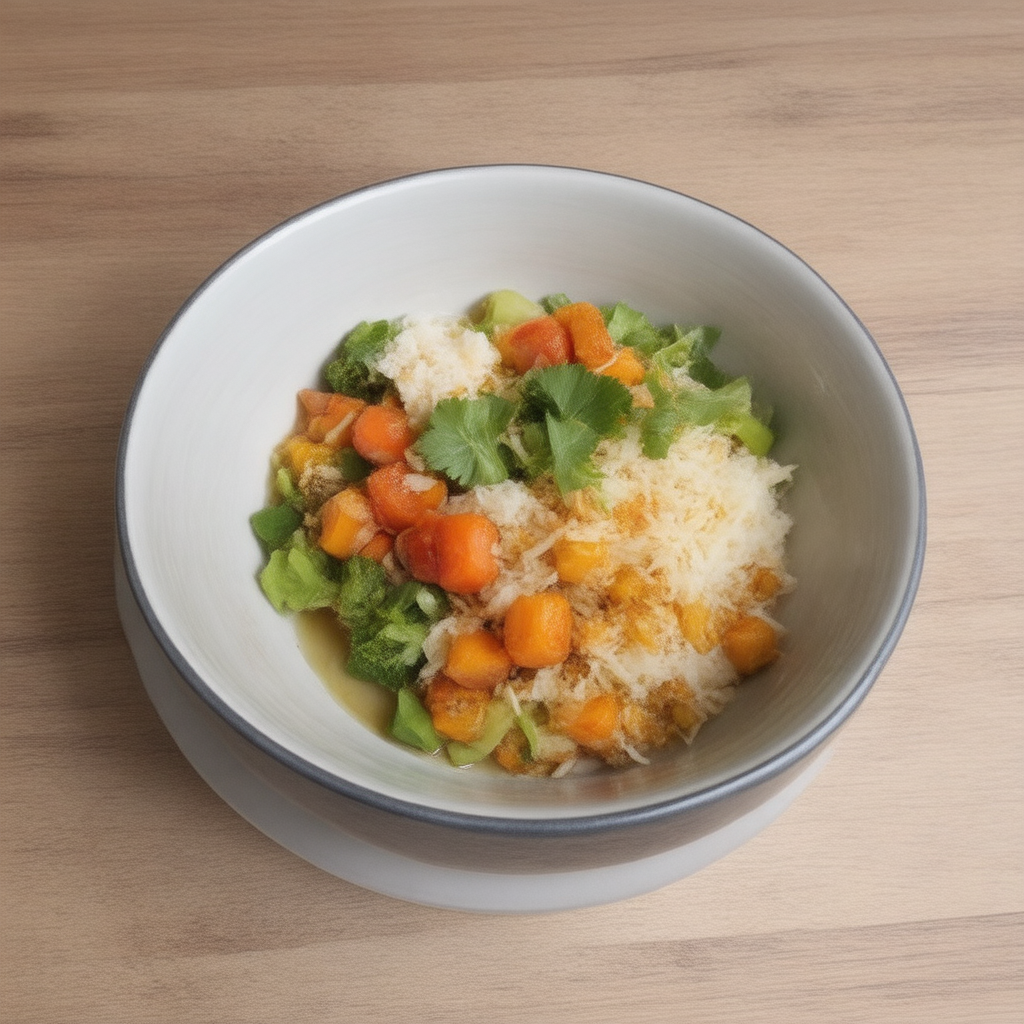}
    \end{minipage}%
    \begin{minipage}{0.2\textwidth}
     \centering
        \includegraphics[width=\linewidth]{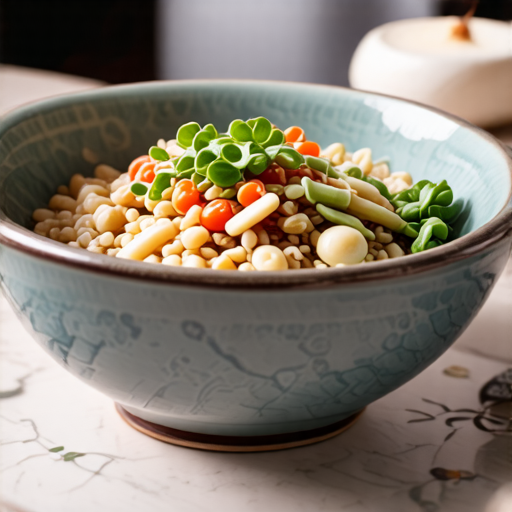}
    \end{minipage}%
\vspace{0.1cm}
    \begin{minipage}{0.2\textwidth}
     \centering
        \includegraphics[width=\linewidth]{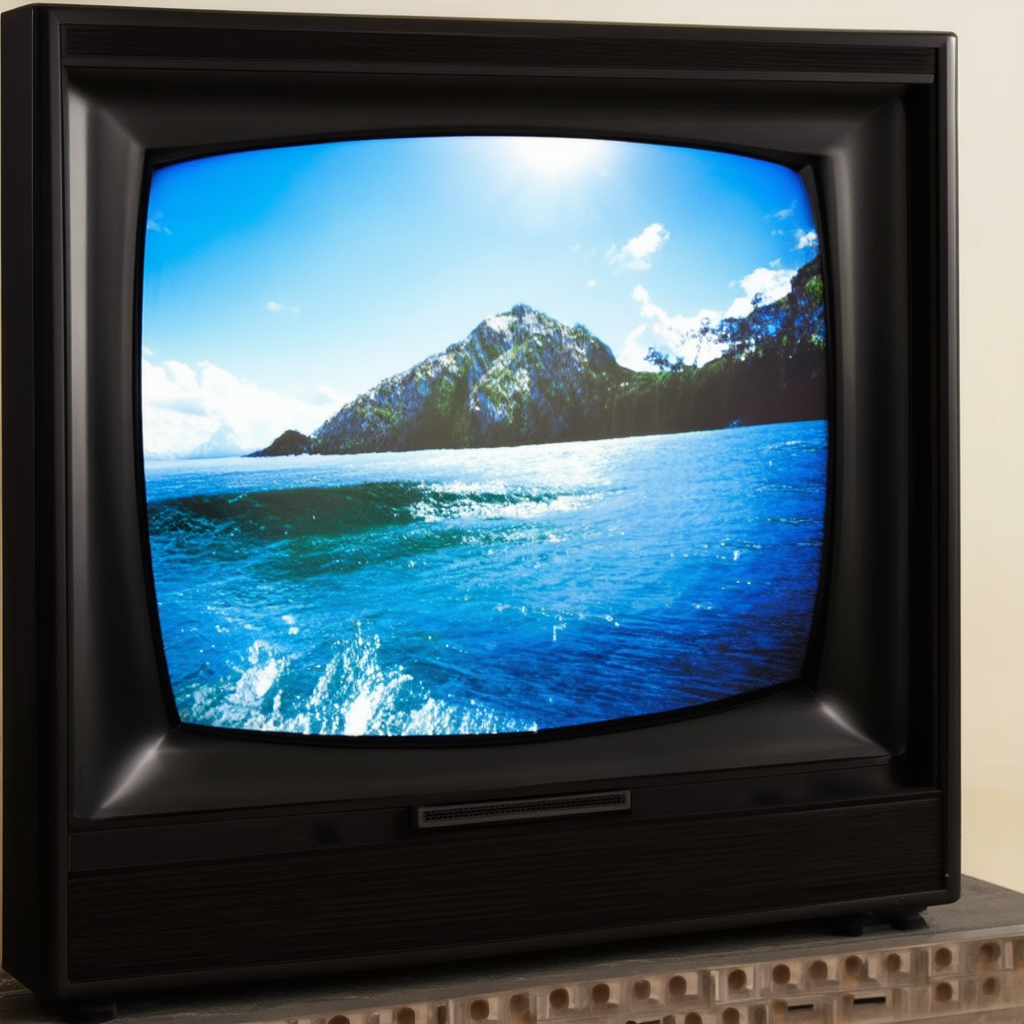}
    \end{minipage}%
      \begin{minipage}{0.2\textwidth}
       \centering
        \includegraphics[width=\linewidth]{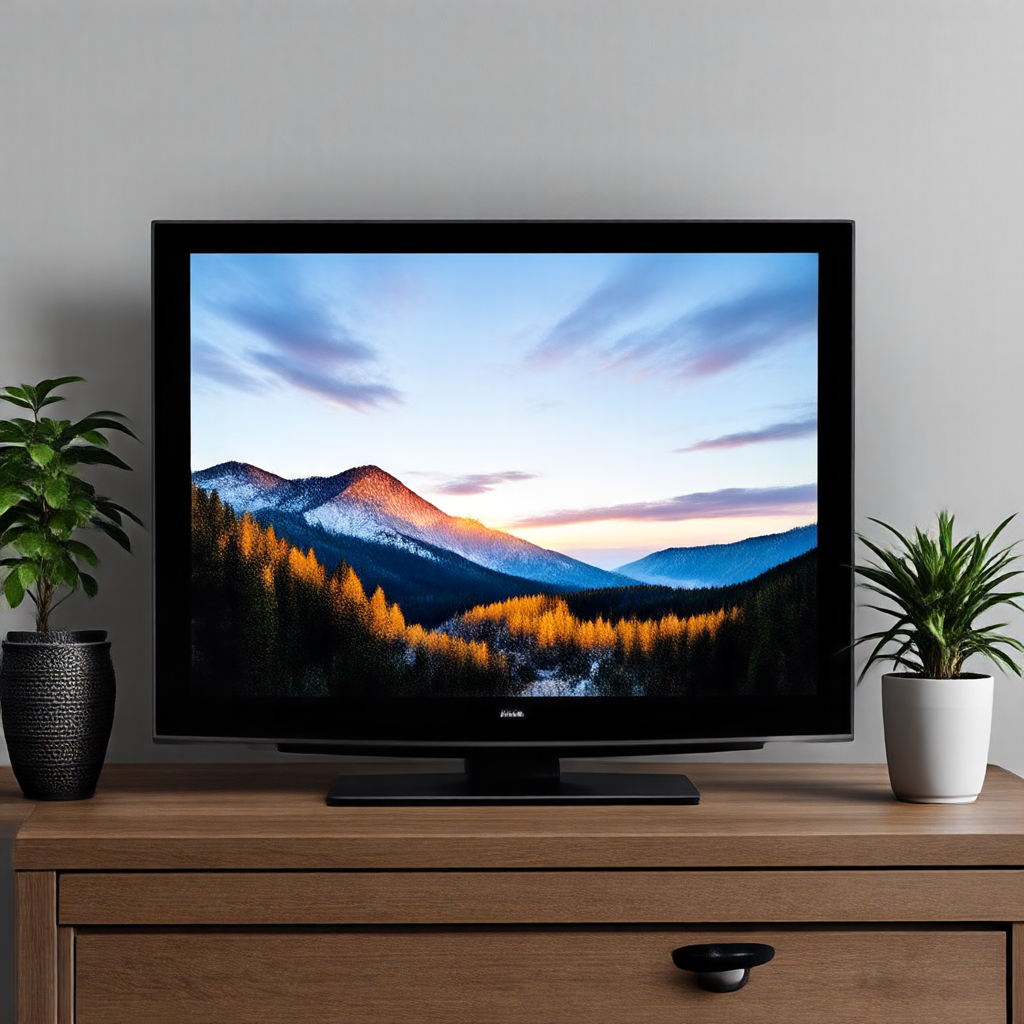}
    \end{minipage}%
    \begin{minipage}{0.2\textwidth}
     \centering
        \includegraphics[width=\linewidth]{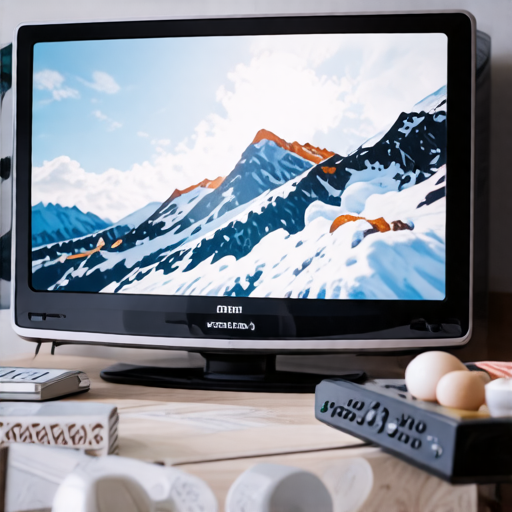}
    \end{minipage}%
    \begin{minipage}{0.2\textwidth}
     \centering
        \includegraphics[width=\linewidth]{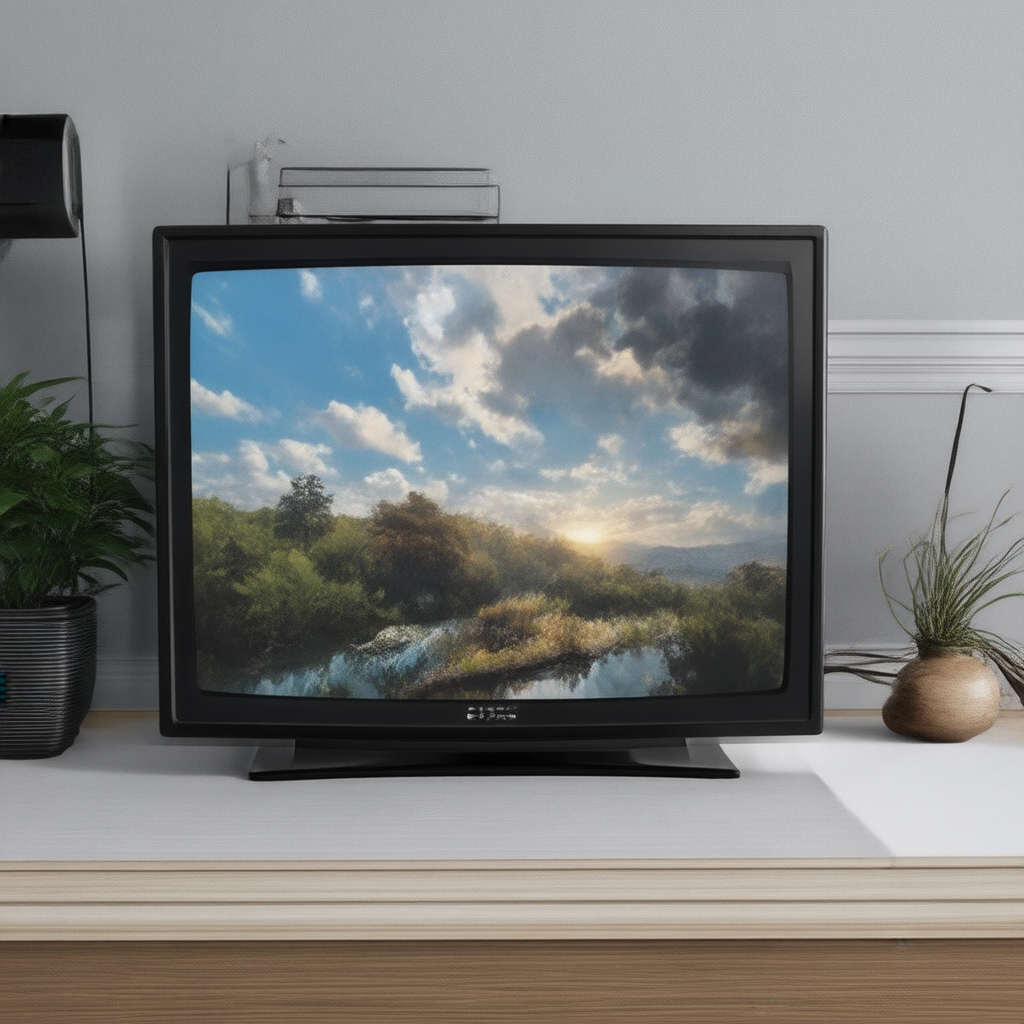}
    \end{minipage}%
    \begin{minipage}{0.2\textwidth}
     \centering
        \includegraphics[width=\linewidth]{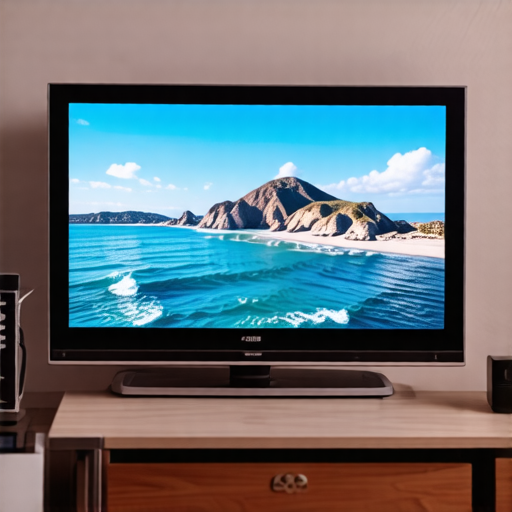}
    \end{minipage}%
    \vspace{0.1cm}
    \begin{minipage}{0.2\textwidth}
     \centering
        \includegraphics[width=\linewidth]{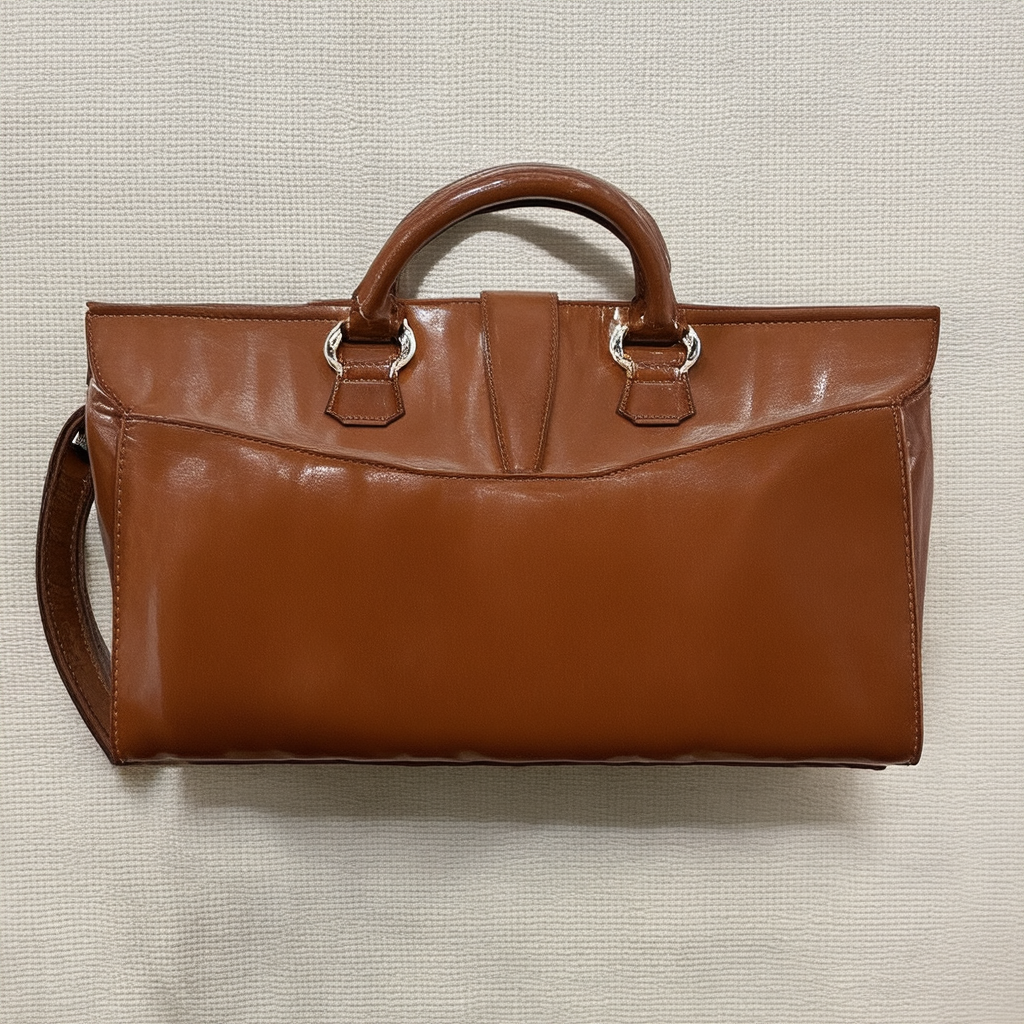}
    \end{minipage}%
      \begin{minipage}{0.2\textwidth}
       \centering
        \includegraphics[width=\linewidth]{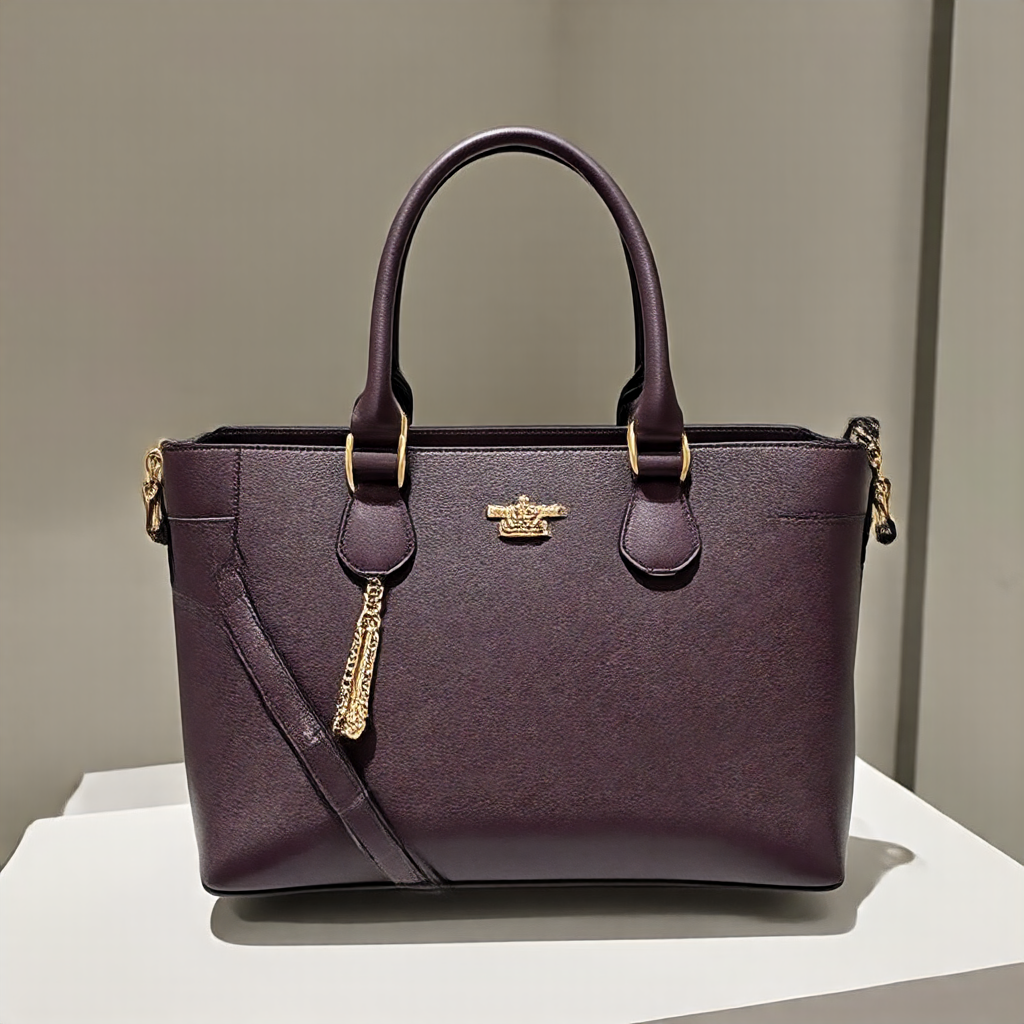}
    \end{minipage}%
    \begin{minipage}{0.2\textwidth}
     \centering
        \includegraphics[width=\linewidth]{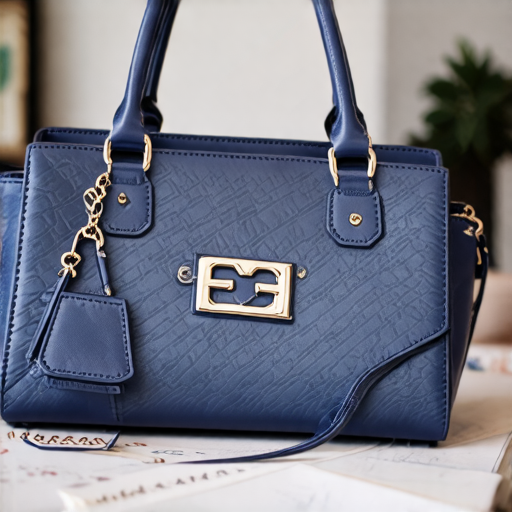}
    \end{minipage}%
    \begin{minipage}{0.2\textwidth}
     \centering
        \includegraphics[width=\linewidth]{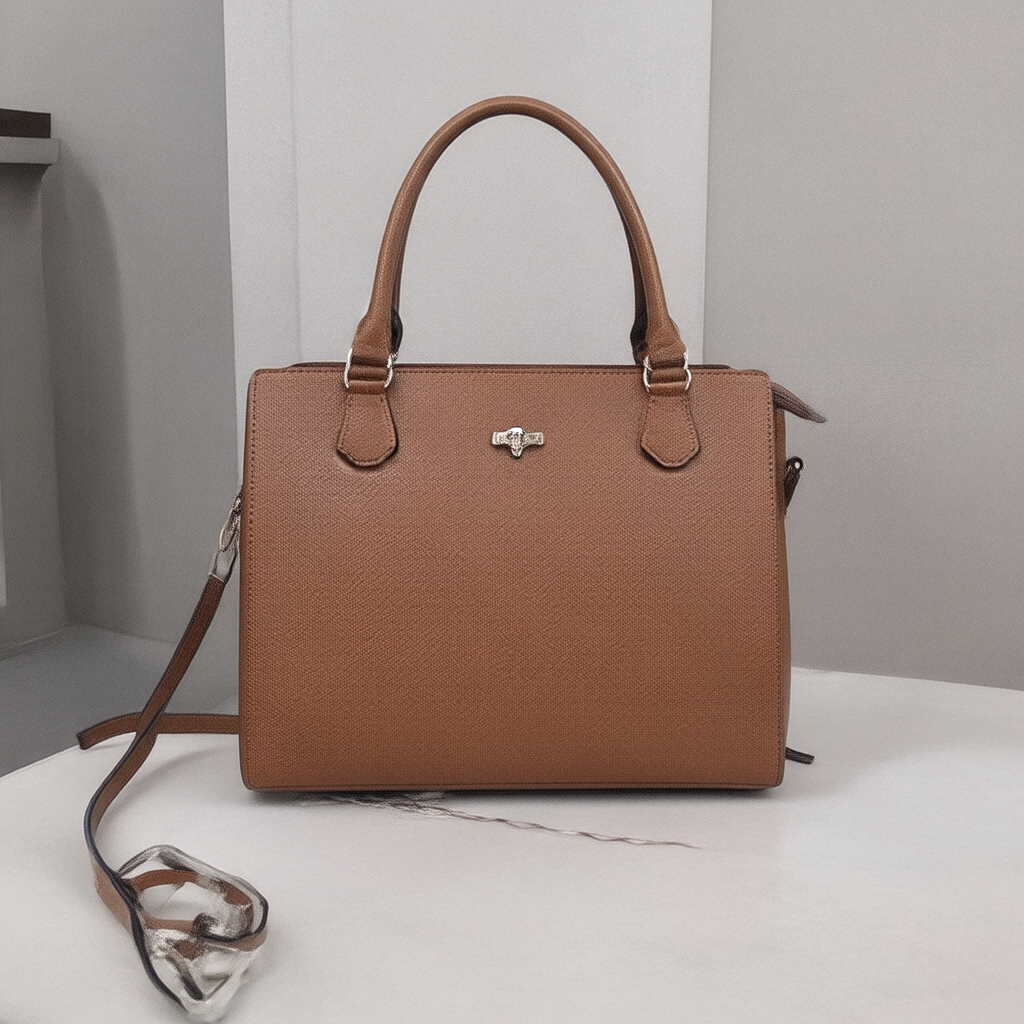}
    \end{minipage}%
    \begin{minipage}{0.2\textwidth}
     \centering
        \includegraphics[width=\linewidth]{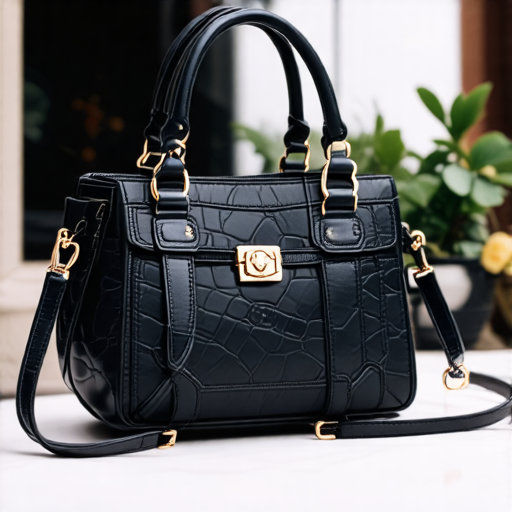}
    \end{minipage}%
   ‘\vspace{0.1cm}
    \begin{minipage}{0.2\textwidth}
     \centering
        \includegraphics[width=\linewidth]{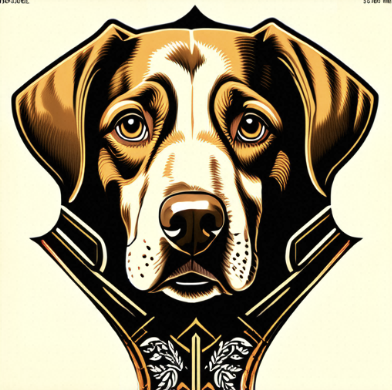}
    \end{minipage}%
      \begin{minipage}{0.2\textwidth}
       \centering
        \includegraphics[width=\linewidth]{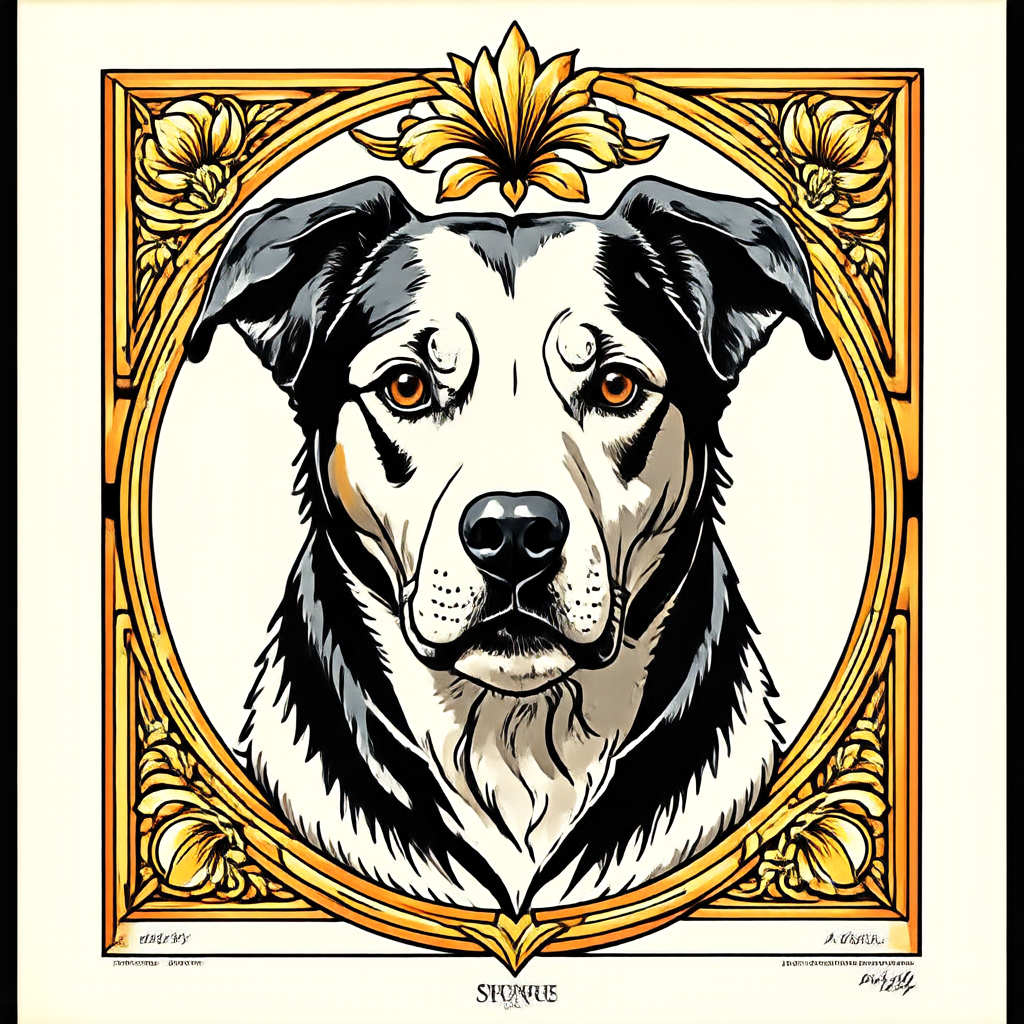}
    \end{minipage}%
    \begin{minipage}{0.2\textwidth}
     \centering
        \includegraphics[width=\linewidth]{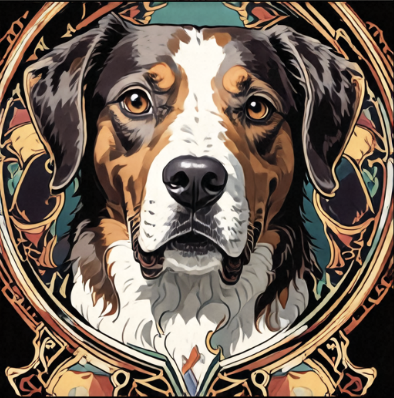}
    \end{minipage}%
    \begin{minipage}{0.2\textwidth}
     \centering
        \includegraphics[width=\linewidth]{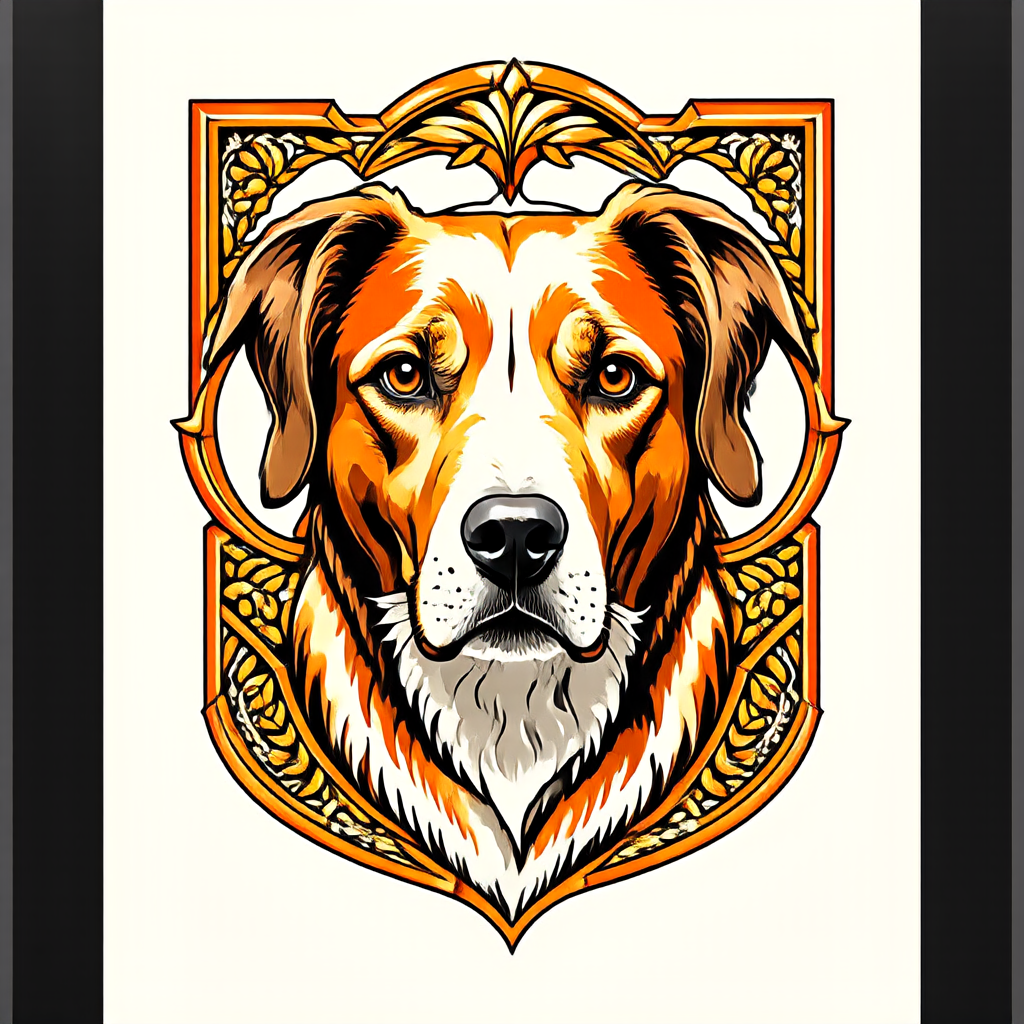}
    \end{minipage}%
    \begin{minipage}{0.2\textwidth}
     \centering
        \includegraphics[width=\linewidth]{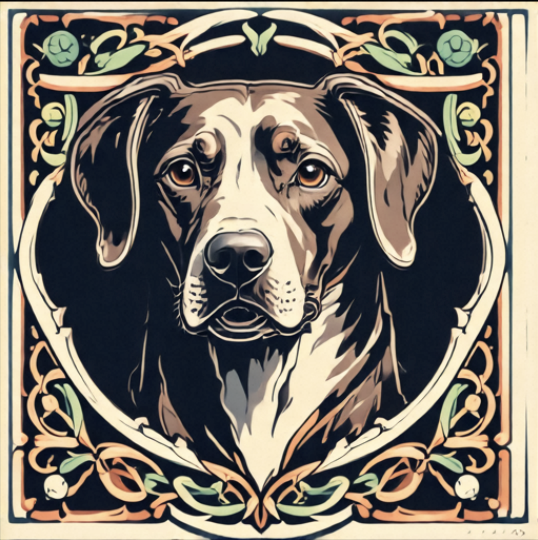}
    \end{minipage}%
    \vspace{0.1cm}
   ‘\vspace{0.1cm}
    \begin{minipage}{0.2\textwidth}
     \centering
        \includegraphics[width=\linewidth]{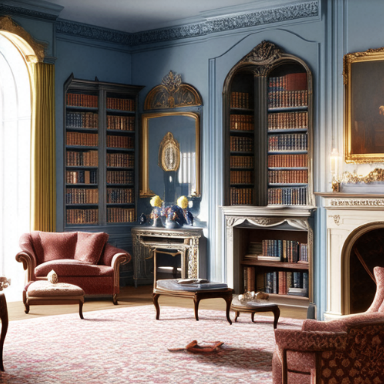}
    \end{minipage}%
      \begin{minipage}{0.2\textwidth}
       \centering
        \includegraphics[width=\linewidth]{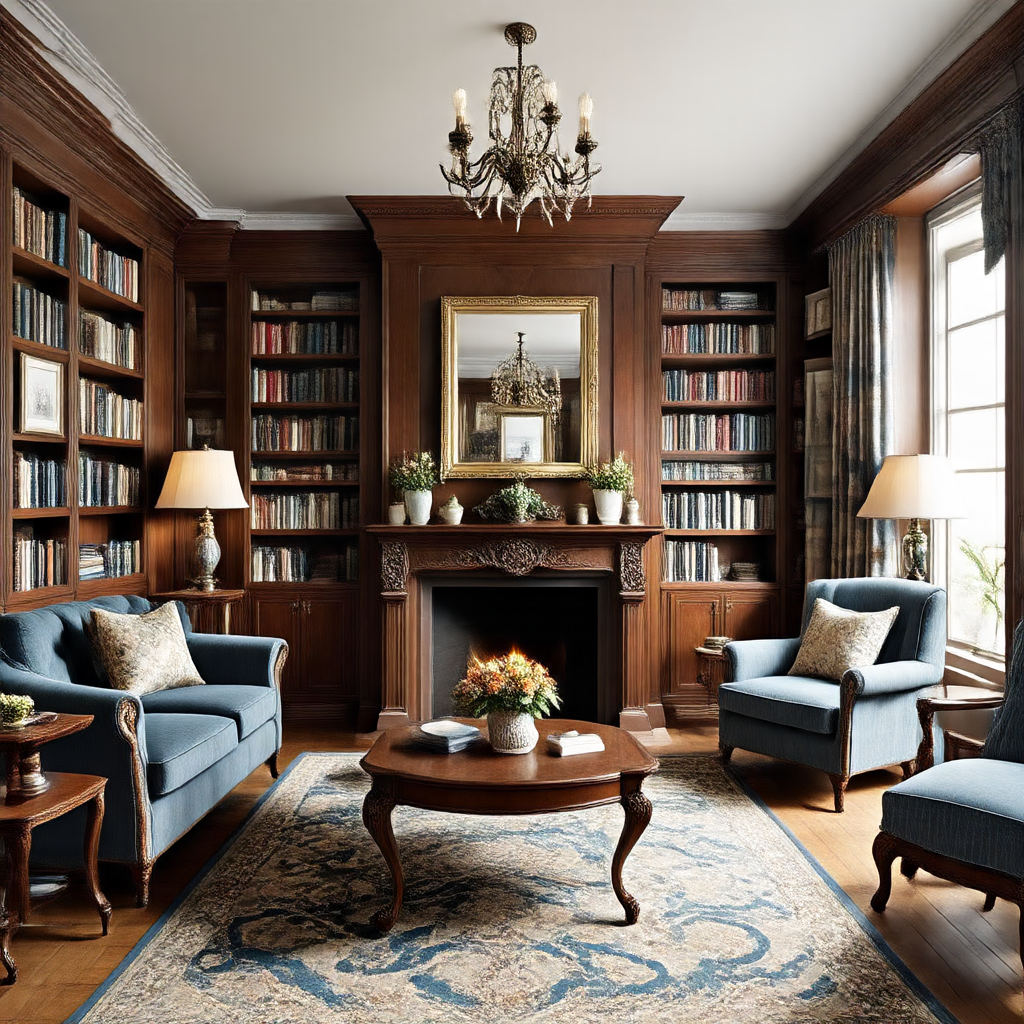}
    \end{minipage}%
    \begin{minipage}{0.2\textwidth}
     \centering
        \includegraphics[width=\linewidth]{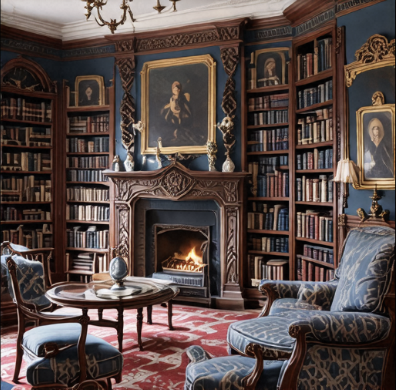}
    \end{minipage}%
    \begin{minipage}{0.2\textwidth}
     \centering
        \includegraphics[width=\linewidth]{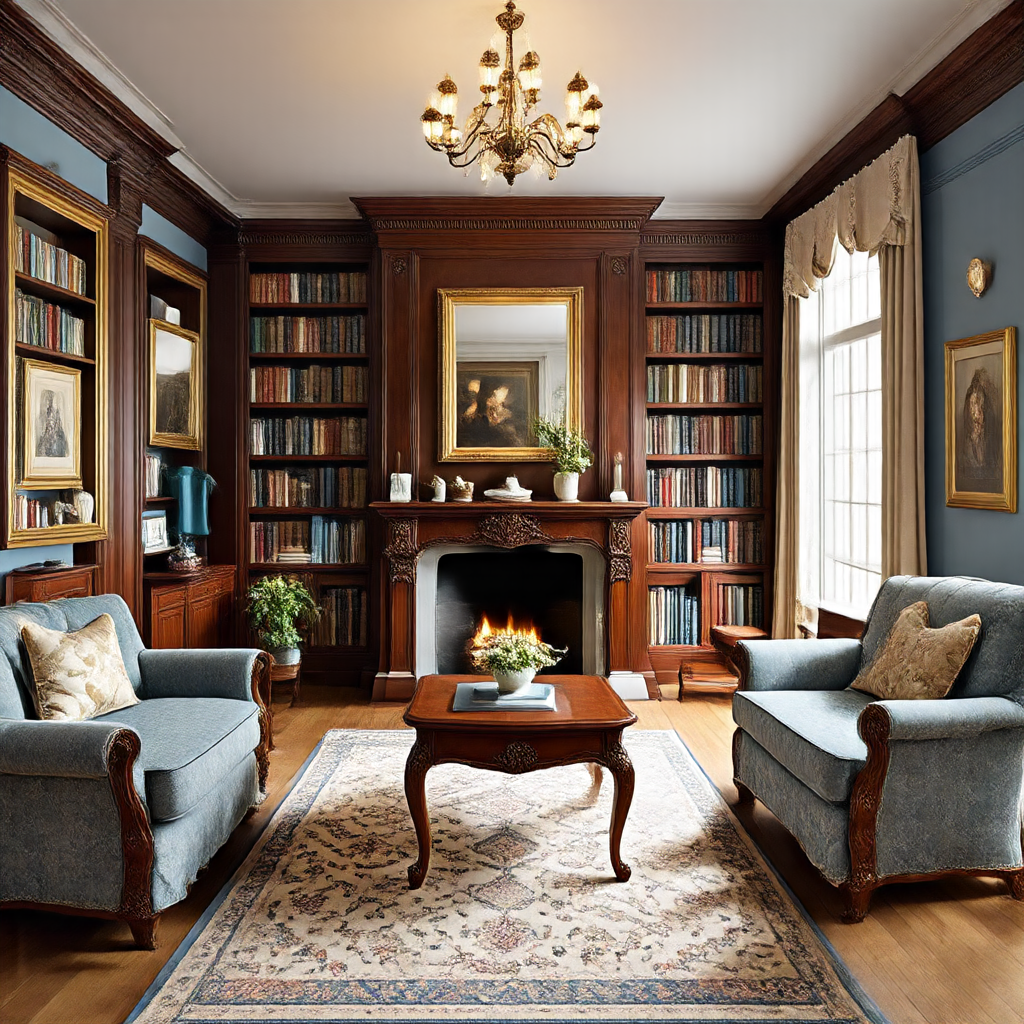}
    \end{minipage}%
    \begin{minipage}{0.2\textwidth}
     \centering
        \includegraphics[width=\linewidth]{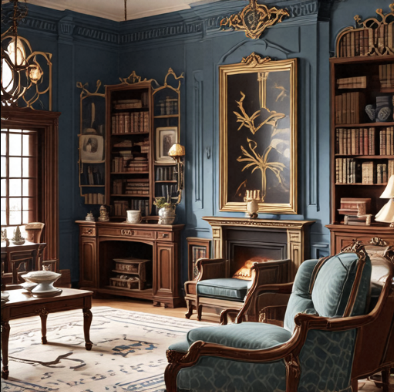}
    \end{minipage}%
    \vspace{0.1cm}
    \caption{\textbf{ Comparison between SD3, TDM, DMD2, Hyper-SD and AdvDMD}}
    \label{fig:compare}
\end{figure}

\noindent \textbf{Baselines.} 
We validate the effectiveness of AdvDMD on both SD3-medium and SD3.5-medium base models by comparing their performance against the original multi-step teacher models. 
Specifically, for SD3, we compare the 4-step AdvDMD model against existing distillation methods including Hyper-SD~\citep{ren2024hyper}, TDM~\citep{luo2025tdm} and DMD2~\citep{yin2024improved} under the same step budget. 
For SD3.5, we benchmark our AdvDMD-trained 4-step model against SD3.5-Flash~\citep{bandyopadhyay2025sd3} and SD3.5-Turbo, which represent current state-of-the-art fast inference variants. 
This comprehensive comparison demonstrates the superiority of AdvDMD in preserving generation quality at extremely low inference budgets while maintaining strong alignment.

\noindent \textbf{Quantitative Results.}
The \cref{tab:dpg_bench_performance_with_step} presents a comprehensive comparison on DPG-Bench across different model variants and sampling steps. 
Overall, our proposed AdvDMD in SD3-medium and SD3.5-medium consistently achieve competitive performance relative to their respective baselines and prior acceleration methods, and in several cases demonstrate superior results.
For the SD3.5 backbone, AdvDMD with only 4 sampling steps achieves the highest overall score of 84.65, slightly outperforming the full-step SD3.5 baseline, which employs 40 sampling steps and a classifier-free guidance scale of 3.0 to obtain a score of 84.55. 
When further reduced to 2 sampling steps, AdvDMD maintains a strong overall score of 84.11, outperforming its DMD2* counterpart at the same step count and achieving the highest Global score of 92.42 among all SD3.5 variants. 
A similar trend is observed for the SD3 backbone.
AdvDMD with 4 sampling steps achieves the highest overall score of 84.25 among all 4-step SD3-based methods and substantially improves upon the original SD3 baseline, which uses 25 sampling steps and a classifier-free guidance scale of 7.0. 
In comparison with DMD2*, Hyper-SD~\citep{ren2024hyper}, and TDM~\citep{luo2025tdm}, our method achieves more balanced performance across all fine-grained metrics. 

Table \ref{tab:geneval_performance} reports the GenEval results across different model variants and sampling steps. 
Overall, our method AdvDMD demonstrates competitive and stable performance under aggressive step reduction. 
For the SD3.5 backbone, AdvDMD with 4 sampling steps achieves the highest overall score of 0.70, matching or slightly surpassing the 40-step SD3.5 baseline, which obtains a score of 0.69, while using only one-tenth of the sampling steps. 
At 2 sampling steps, AdvDMD maintains a higher overall score of 0.67 compared to its DMD2* counterpart, which achieves 0.65. 
It also shows notable improvements in the Two-object and Colors metrics, demonstrating superior robustness in few step regimes.
AdvDMD for SD3 with 4 sampling steps achieves an overall score of 0.68, outperforming the original SD3 baseline that uses 25 sampling steps and obtains a score of 0.62, while remaining competitive with other 4-step acceleration methods.
These results suggest that AdvDMD effectively preserves compositional and attribute-level fidelity under substantial step reduction, achieving comparable or superior overall quality relative to full-step baselines and existing acceleration approaches.

\begin{figure}[t]
    \centering
    \begin{minipage}{0.2\textwidth}  
        \centering
        \textbf{PickScore}  
    \end{minipage}%
    \begin{minipage}{0.2\textwidth}  
     \centering
        \includegraphics[width=0.8\linewidth]{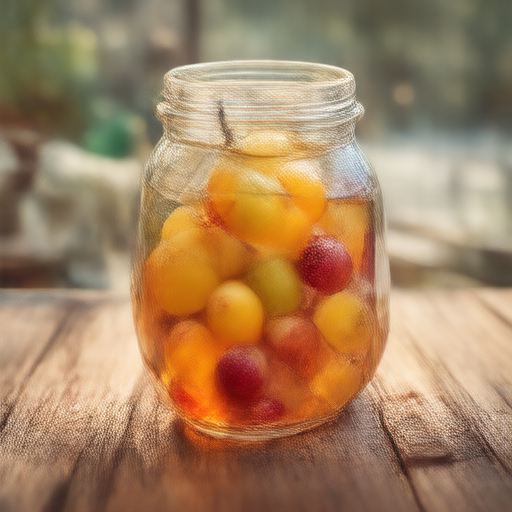}
    \end{minipage}%
    \begin{minipage}{0.2\textwidth}
     \centering
        \includegraphics[width=0.8\linewidth]{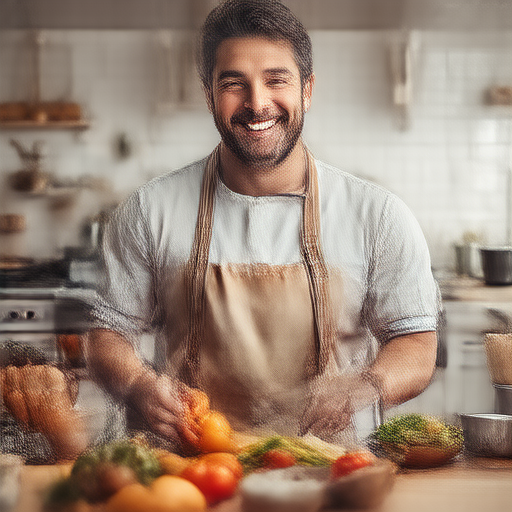}
    \end{minipage}%
    \begin{minipage}{0.2\textwidth}
     \centering
        \includegraphics[width=0.8\linewidth]{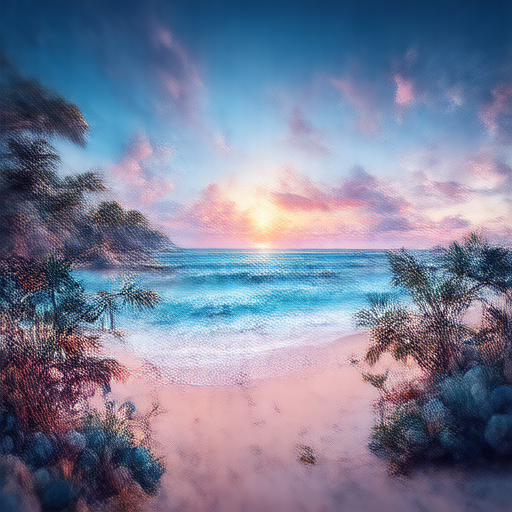}
    \end{minipage}%
    \begin{minipage}{0.2\textwidth}
     \centering
        \includegraphics[width=0.8\linewidth]{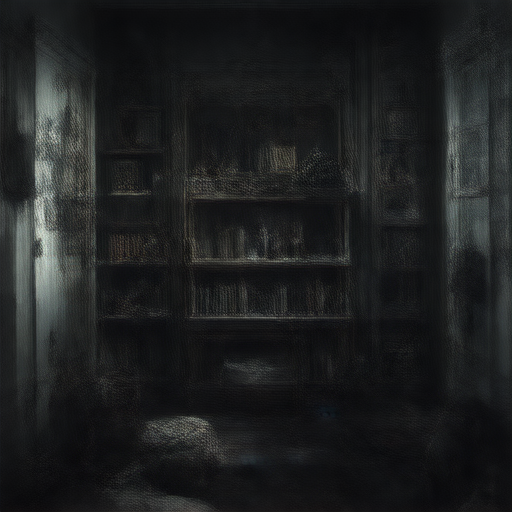}
    \end{minipage}%

    \vspace{-0.1cm}
    \begin{minipage}{0.2\textwidth}
        \centering
        \textbf{HPS}  
    \end{minipage}%
    \begin{minipage}{0.2\textwidth}
     \centering
        \includegraphics[width=0.8\linewidth]{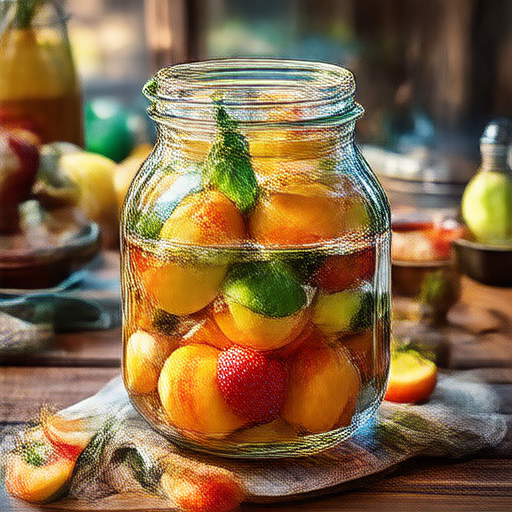}
    \end{minipage}%
    \begin{minipage}{0.2\textwidth}
     \centering
        \includegraphics[width=0.8\linewidth]{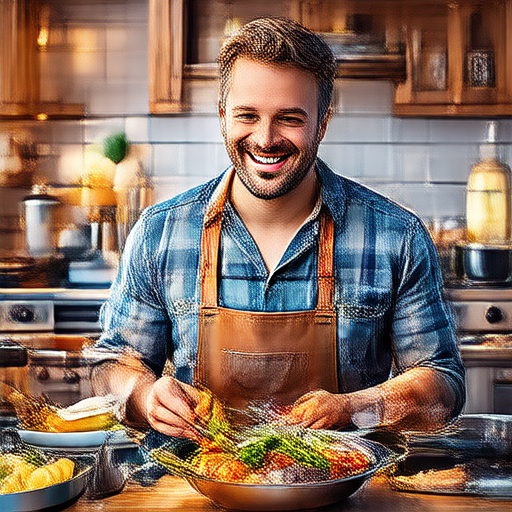}
    \end{minipage}%
    \begin{minipage}{0.2\textwidth}
     \centering
        \includegraphics[width=0.8\linewidth]{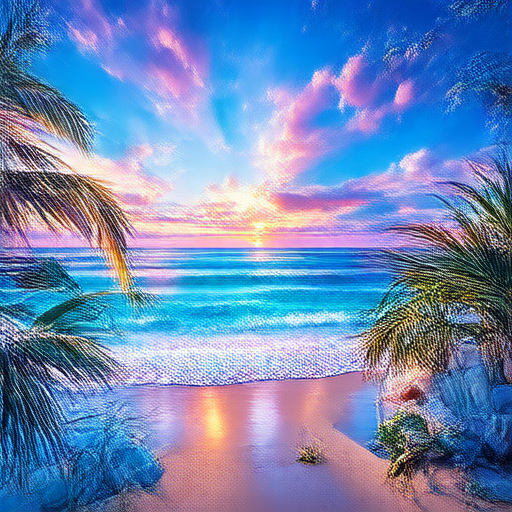}
    \end{minipage}%
    \begin{minipage}{0.2\textwidth}
     \centering
        \includegraphics[width=0.8\linewidth]{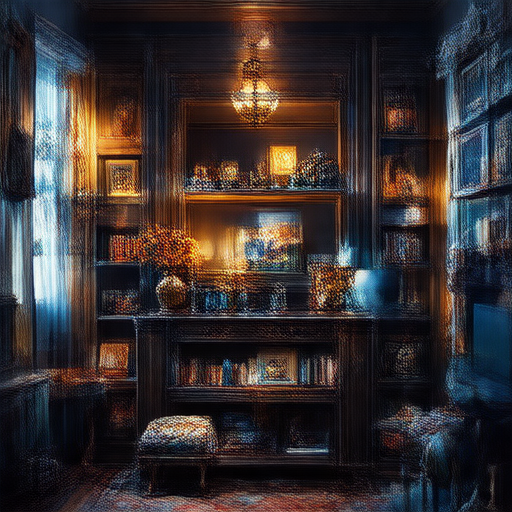}
    \end{minipage}%
    \vspace{-0.1cm}
    \begin{minipage}{0.2\textwidth}
        \centering
        \textbf{AdvDMD}  
    \end{minipage}%
    \begin{minipage}{0.2\textwidth}
     \centering
        \includegraphics[width=0.8\linewidth]{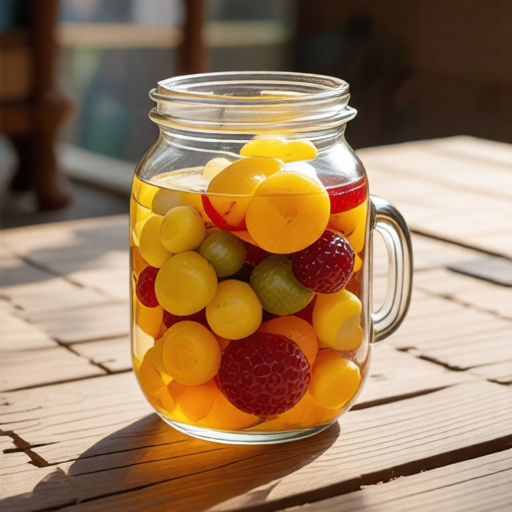}
    \end{minipage}%
    \begin{minipage}{0.2\textwidth}
     \centering
        \includegraphics[width=0.8\linewidth]{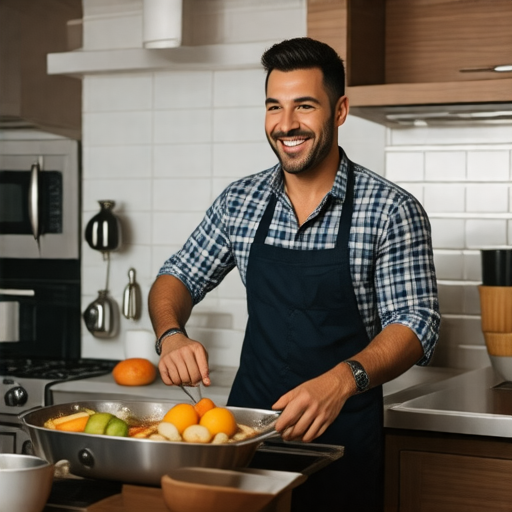}
    \end{minipage}%
    \begin{minipage}{0.2\textwidth}
     \centering
        \includegraphics[width=0.8\linewidth]{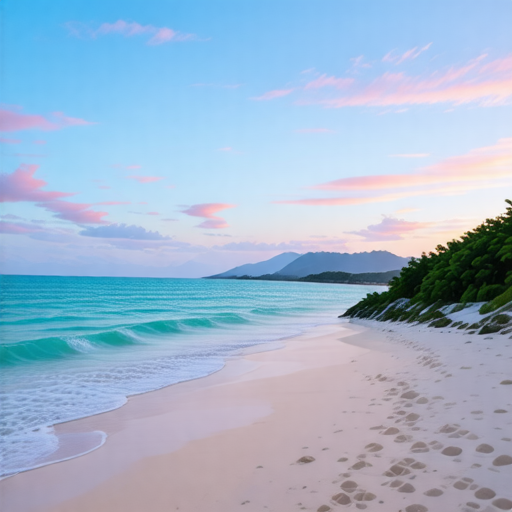}
    \end{minipage}%
    \begin{minipage}{0.2\textwidth}
     \centering
        \includegraphics[width=0.8\linewidth]{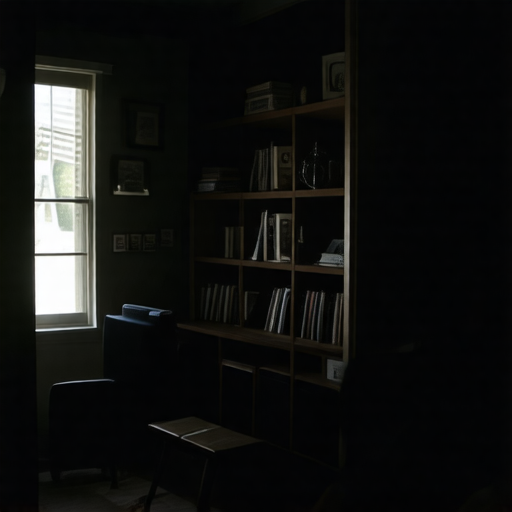}
    \end{minipage}%
 \vspace{-0.1cm}
    \caption{\textbf{4-step generation results of Flow-GRPO trained using the corresponding reward model.}The noticeable blurring and color bias observed in the generated images are indicative of apparent reward hacking.}
    \label{fig:grpo_reward}
\end{figure}

\begin{table*}[t]
    \centering
    \small
    \setlength{\tabcolsep}{1pt} 
    \definecolor{lightgray}{RGB}{230,230,230} 
    \begin{tabular}{lccccc}
        \toprule
        Reward  & Step&ClipScore $\uparrow$ & HPS $\uparrow$ &PickScore$\uparrow$ & DEQA $\uparrow$ \\
        \midrule
        Base   &40 & 0.389  & 0.282  & 0.223  & 4.14 \\
        HPS   &4 & 0.369  & 0.324  & 0.223  & 3.91 \\
        PickScore & 4 & 0.369 & 0.258  & 0.224  & 3.84 \\
        \bottomrule
    \end{tabular}
    \vspace{0.2cm}
    \caption{\textbf{Quantitative evaluation of aesthetic metrics for Flow-GRPO in the training of few-step generative models.}
    Here, Base denotes the original multi-step model, which is compared with the 4-step model trained using the corresponding reward model.
    }
    \label{tab:aes_score}
\end{table*}
\subsection{Ablation Studies}
\noindent \textbf{Fixed reward model for low steps.} 
In contrast to traditional reinforcement learning, the model lacks basic generation capabilities at few step counts.
Therefore, directly applying reinforcement learning strategies fails to yield comprehensive performance improvements.
Here we present the results under standalone Flow-GRPO~\citep{liu2025flow} training at 4 steps. 

Table \ref{tab:aes_score} presents the aesthetic evaluation results of Flow-GRPO when optimizing different reward models for few-step generation. As shown, training with a specific reward model tends to improve the corresponding metric but does not consistently enhance overall perceptual quality. For example, optimizing with HPS at 4 steps increases the HPS score from 0.282 (Base, 40 steps) to 0.324, yet results in a noticeable drop in ClipScore (0.369 vs. 0.389) and DEQA (3.91 vs. 4.14). A similar pattern is observed when optimizing PickScore: although PickScore slightly improves (0.224 vs. 0.223), other metrics, including HPS and DEQA, decline. This discrepancy indicates a clear reward hacking phenomenon, where the model overfits to the optimized reward signal without achieving holistic quality improvements. In contrast to the balanced performance of the original multi-step Base model, reward-specific optimization in the few-step regime may bias the generation toward narrow metric gains at the expense of broader aesthetic fidelity, highlighting the limitations of directly aligning training objectives with a single reward proxy.

In \cref{fig:grpo_reward}, we present the results of Flow-GRPO using HPS and PickScore individually as reward models, without incorporating knowledge distillation. We observe that although the main content of the images can be generated with vivid colors, the overall outputs suffer from severe blurring artifacts in multiple regions.
In the last column, we present the result from the prompt "a very dark room......", which typically fails to align well with the reward model.
It can be observed that employing HPS as the reward model leads to a degradation in the model’s instruction-following capability.
This indicates that directly using a reward model in low-step training leads to severe reward hacking issues.
We only present results at 4 inference steps here.
Under even fewer steps (e.g., 1 or 2 steps), the model completely lacks the necessary generation capability and thus fails to produce valid images.

\begin{table*}[t]
    \centering
    \small
    \setlength{\tabcolsep}{1pt} 
    \begin{tabular}{l ccccccc}
        \toprule
            Reward & \fontsize{8pt}{9pt}\selectfont relation $\uparrow$ & \fontsize{8pt}{9pt}\selectfont other $\uparrow$ & \fontsize{8pt}{9pt}\selectfont attribute $\uparrow$ & \fontsize{8pt}{9pt}\selectfont entity $\uparrow$ & \fontsize{8pt}{9pt}\selectfont global $\uparrow$ & \fontsize{8pt}{9pt}\selectfont all $\uparrow$ \\
        \midrule
        Discriminator           & 0.9104 & 0.8969 & 0.9009 & 0.8894 & 0.8960 & 0.8406 \\
        +HPS, ClipScore  & \textbf{0.9144} & 0.8494 & \textbf{0.9043} & \textbf{0.8986} & \textbf{0.8982} & \textbf{0.8465} \\
        \bottomrule
    \end{tabular}
    \vspace{0.2cm}
    \caption{\textbf{Impact of Combined Reward on Performance Improvement across DPG-bench Metrics.}Here '+'
   indicates that an additional corresponding reward model is incorporated on top of the original discriminator-based reward.
    }
    \label{tab:dpg_bench_performance_new}
\end{table*}
\noindent \textbf{} 

\begin{table*}[t]
    \centering
    \small
    \setlength{\tabcolsep}{1pt} 
    \begin{tabular}{l cccccccc}
        \toprule
        Rewards & \fontsize{8pt}{9pt}\selectfont Single  $\uparrow$ & \fontsize{8pt}{9pt}\selectfont Two  $\uparrow$ & \fontsize{8pt}{9pt}\selectfont Count $\uparrow$ & \fontsize{8pt}{9pt}\selectfont Colors $\uparrow$ & \fontsize{8pt}{9pt}\selectfont Position $\uparrow$ & \fontsize{8pt}{9pt}\selectfont Attr $\uparrow$ & \fontsize{8pt}{9pt}\selectfont Overall $\uparrow$ \\
        \midrule
       Discriminator            & 0.9844 & \textbf{0.9040} & 0.6625 & 0.8059 & \textbf{0.2975} & \textbf{0.5843} & 0.70642 \\
        +HPS, ClipScore  & \textbf{0.9969} & 0.8842 & \textbf{0.6719} & \textbf{0.8457} & 0.2789 & 0.5625 & \textbf{0.7070} \\
        \bottomrule
    \end{tabular}
    \vspace{0.2cm}
    \caption{\textbf{Performance improvement achieved by the combined reward on Geneval metrics.}
   Here '+'
   indicates that an additional corresponding reward model is incorporated on top of the original discriminator-based reward.
    }
    \label{tab:geneval_performance_new}
\end{table*}
\noindent \textbf{Hybrid Training with Other Reward Models.} 
Tables \ref{tab:dpg_bench_performance_new} and \ref{tab:geneval_performance_new} analyze the impact of incorporating additional aesthetic rewards on top of the discriminator-based objective. 
Notably, using the discriminator alone already yields strong and well-balanced performance across both DPG-Bench and GenEval. 
On DPG-Bench, the discriminator achieves competitive scores across all fine-grained metrics, with an overall score of 0.8406.
When further combined with HPS and ClipScore, the overall performance improves to 0.8465, with gains in relation, attribute, entity, and global metrics, indicating that auxiliary rewards can provide complementary signals without disrupting the core semantic alignment learned by the discriminator.

A similar trend is observed on GenEval. The adversarial reward setting attains a strong Overall score of 0.7064, achieving the best performance on Two-object, Position, and Attr metrics. After incorporating HPS and ClipScore, the Overall score further increases to 0.7070.
Although certain dimensions slightly decrease, the combined reward leads to a marginal but consistent overall gain. These results suggest that the discriminator itself serves as a powerful and stable reward signal, capable of driving high-quality generation independently, while additional reward models can offer modest complementary improvements.
\section{Conclusion}
\label{sec:conclusion}
In this paper, we propose AdvDMD for training few step generation models with high quality.
AdvDMD employs the discriminator from the gan objective as a reward model within the reinforcement learning framework,  which provides reward signals for intermediate steps.
The adversarial reward serves as a continuous learning signal that guides the generator toward the true data distribution by providing discriminative feedback at each training step, thereby promoting perceptual realism and structural fidelity in the generated samples.
In future work, we will further explore how to better integrate distillation and reinforcement learning for high-performance few-step generation.
\setlength{\bibsep}{5pt}
\bibliography{reference}
\bibliographystyle{plainnat}

\newpage
\appendix
\section{Ablation Studies}
\noindent \textbf{SDE backward simulation.}Here we present the corresponding comparative results of ODE and SDE backward simulation.
We conduct our experiments on SD3.5 with a 4-step training setup to validate its impact on model performance.
The results on DPG-bench demonstrate that the randomness introduced by SDE can slightly improve image quality.
\begin{table}[H]
    \centering
    \small
    \setlength{\tabcolsep}{1pt} 
    \begin{tabular}{l ccccccc}
        \toprule
            Methods & Relation $\uparrow$ &  Other $\uparrow$ &  Attribute $\uparrow$ & Entity $\uparrow$ &  Global $\uparrow$ &  Overall $\uparrow$ \\
        \midrule
        ODE          & 0.8772 & 0.8700 & 0.8968 & 0.9141 & 0.9052 & 0.8385 \\
        SDE   & 0.9104 & 0.8969 & 0.9009 & 0.8894 & 0.8960 & 0.8406 \\
        \bottomrule
    \end{tabular}
    \vspace{0.2cm}
    \caption{Impact of SDE backward simulation on Performance Improvement across DPG-bench Metrics.
    }
    \label{tab:dpg_bench_performance_new_}
\end{table}

\section{Algorithm}
\begin{algorithm}
    \caption{AdvDMD Distillation Procedure}
    \label{alg:dmd}
    \begin{algorithmic}[1]
        \Require 
            Generator $G_{\theta}$; real score $\mu_{\text{real}}$; fake score $\mu_{\text{fake}}$; 
            Discriminator $D_\phi$; Image-prompt dataset $\mathcal{D}$; Number of sampling steps $N$
        \While{Training not converged}
            \State Sample an image-prompt pair $(x, c) \sim \mathcal{D}$
            \State Generate images with $G_{\theta}$ conditioned on $c$ over $N$ steps, 
                  yielding intermediate noisy samples $\{y_t^i\}_{t=1}^{N}$ for each image 
                  via backward SDE simulation
            \State Compute step-wise rewards $\{R_t^i\}_{t=1}^{N}$ for the $i$-th image using 
                  discriminator $D_\phi$ (see \cref{eq:discriminator})
            \State Select a subset of noisy samples $\{y_t^i\}_{i=1}^{G}$ at timestep $t$
            \State Denoise $\{y_t^i\}_{i=1}^{G}$ to obtain clean images $\{y_0^i\}_{i=1}^{G}$ via $G_{\theta}$
            \State Add noise to $\{y_0^i\}_{i=1}^{G}$ to generate new noisy samples 
                  $\{y_{t_{\text{new}}}^i\}_{i=1}^{G}$ where $t_{\text{new}} \in [0, 1]$
            
            \If{Training generator}
                \State Calculate DMD loss $\mathcal{L}_{\text{dmd}}$ and GAN loss $\mathcal{L}_{\text{gan}}$ 
                      following \cref{eq:dmd_loss,eq:gan}, respectively
                \State Compute advantage values for each step using \cref{eq:grpo_advantage}
                \State Calculate GRPO loss $\mathcal{L}_{\text{grpo}}$ based on \cref{eq:grpo}
                \State Update generator parameters $\theta$ via total loss $\mathcal{L}_{\text{gen}}$ 
                      (defined in \cref{eq:gen})
            \ElsIf{Updating $\mu_{\text{fake}}$ and discriminator $D_\phi$}
                \State Compute discriminator loss $\mathcal{L}_{\text{dis}}$ and diffusion loss $\mathcal{L}_{\text{diff}}$ 
                      following \cref{eq:dis,eq:diff}
                \State Update $\mu_{\text{fake}}$ and discriminator parameters $\phi$ via gradient descent
            \EndIf
        \EndWhile
    \end{algorithmic}
\end{algorithm}
\section{Additional Image Results}
\begin{figure}[H]
    \centering
    \begin{minipage}{0.20\textwidth}
        \includegraphics[width=\linewidth]{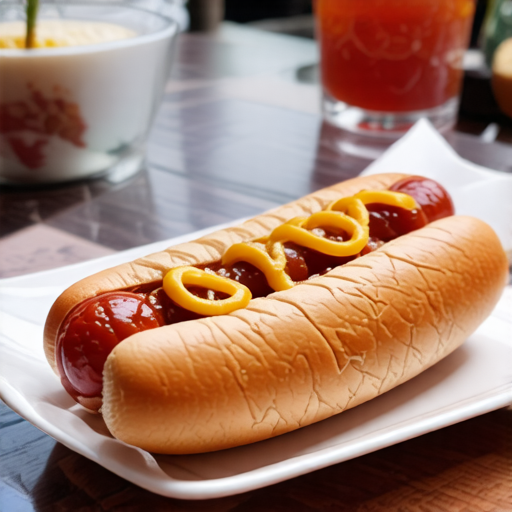}
    \end{minipage}%
    \begin{minipage}{0.20\textwidth}
        \includegraphics[width=\linewidth]{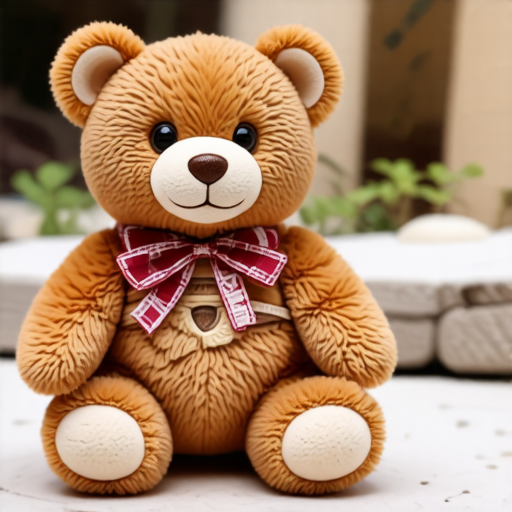}
    \end{minipage}%
    \begin{minipage}{0.20\textwidth}
        \includegraphics[width=\linewidth]{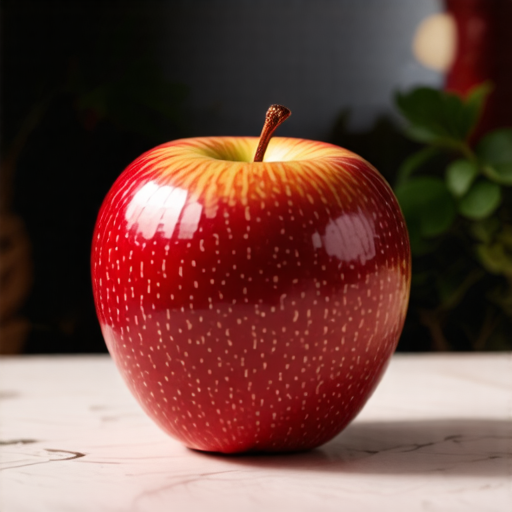}
    \end{minipage}%
    \begin{minipage}{0.20\textwidth}
        \includegraphics[width=\linewidth]{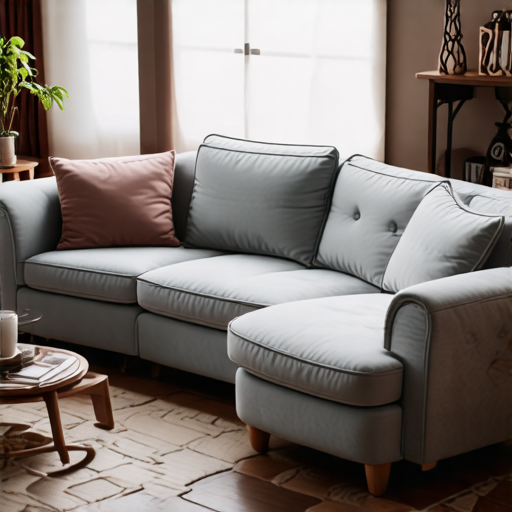}
    \end{minipage}%
    \begin{minipage}{0.20\textwidth}
        \includegraphics[width=\linewidth]{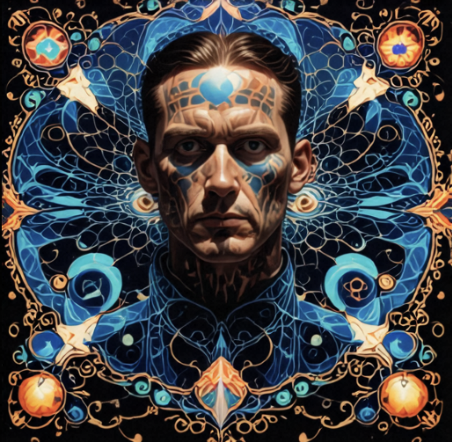}
    \end{minipage}%

    \begin{minipage}{0.2\textwidth}
        \includegraphics[width=\linewidth]{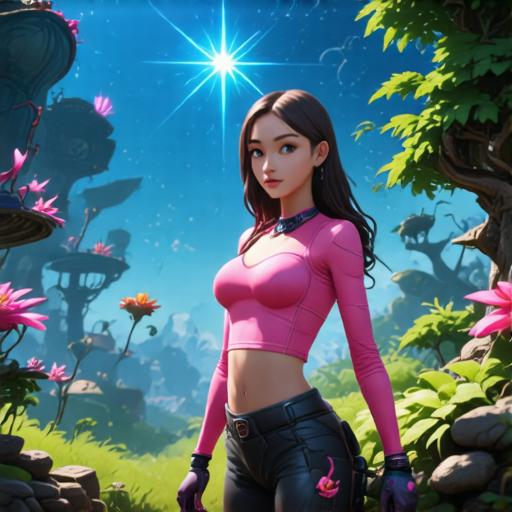}
    \end{minipage}%
    \begin{minipage}{0.2\textwidth}
        \includegraphics[width=\linewidth]{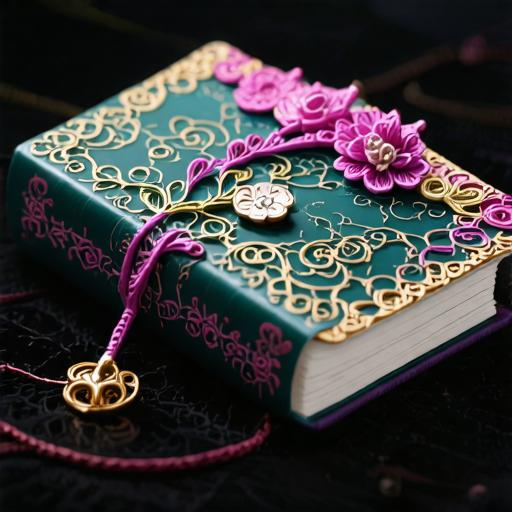}
    \end{minipage}%
    \begin{minipage}{0.2\textwidth}
        \includegraphics[width=\linewidth]{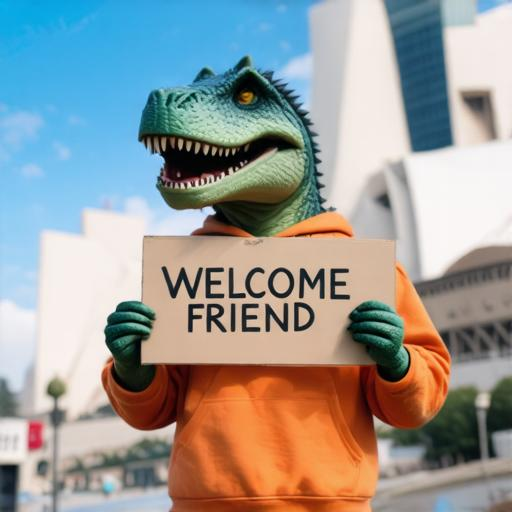}
    \end{minipage}%
    \begin{minipage}{0.2\textwidth}
        \includegraphics[width=\linewidth]{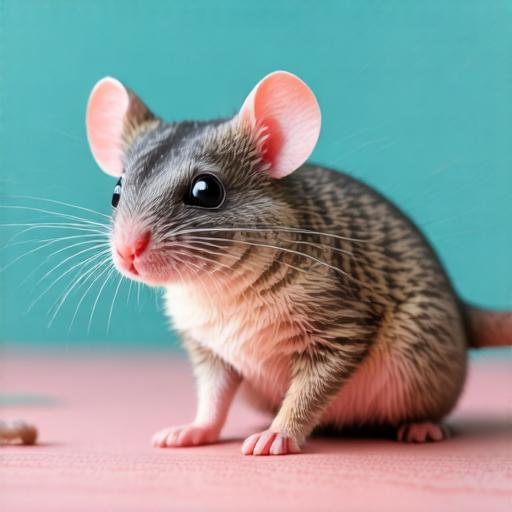}
    \end{minipage}%
    \begin{minipage}{0.20\textwidth}
        \includegraphics[width=\linewidth]{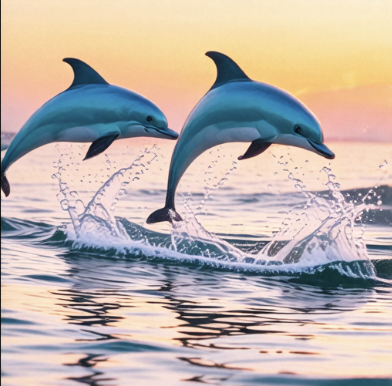}
    \end{minipage}%

    \begin{minipage}{0.2\textwidth}
        \includegraphics[width=\linewidth]{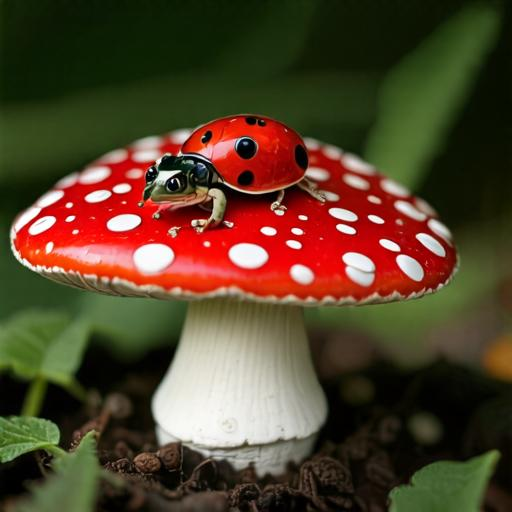}
    \end{minipage}%
    \begin{minipage}{0.2\textwidth}
        \includegraphics[width=\linewidth]{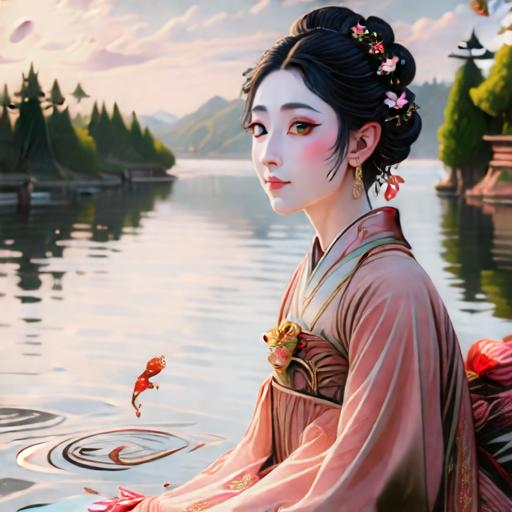}
    \end{minipage}%
    \begin{minipage}{0.2\textwidth}
        \includegraphics[width=\linewidth]{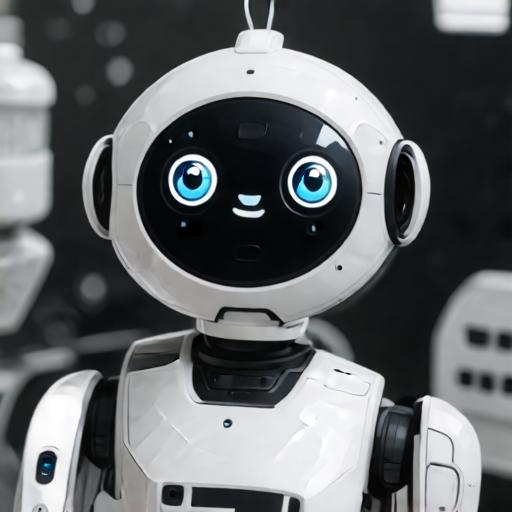}
    \end{minipage}%
    \begin{minipage}{0.2\textwidth}
        \includegraphics[width=\linewidth]{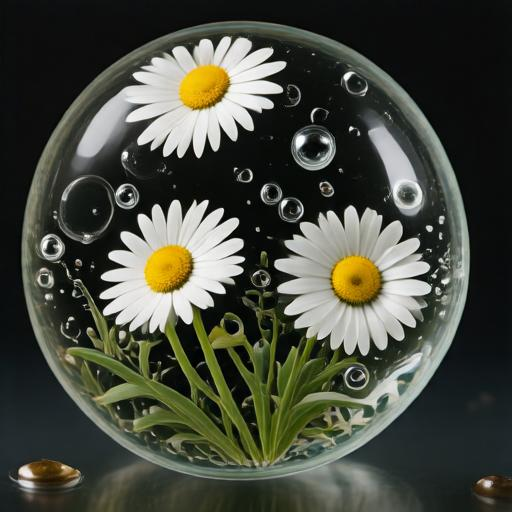}
    \end{minipage}%
    \begin{minipage}{0.20\textwidth}
        \includegraphics[width=\linewidth]{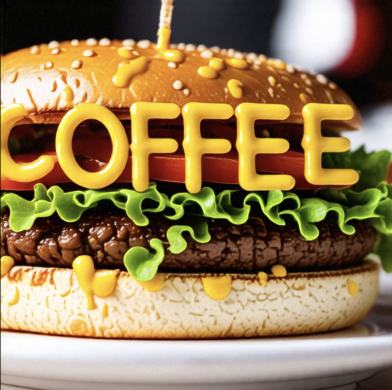}
    \end{minipage}%
    \vspace{0cm}
    \begin{minipage}{0.20\textwidth}
        \includegraphics[width=\linewidth]{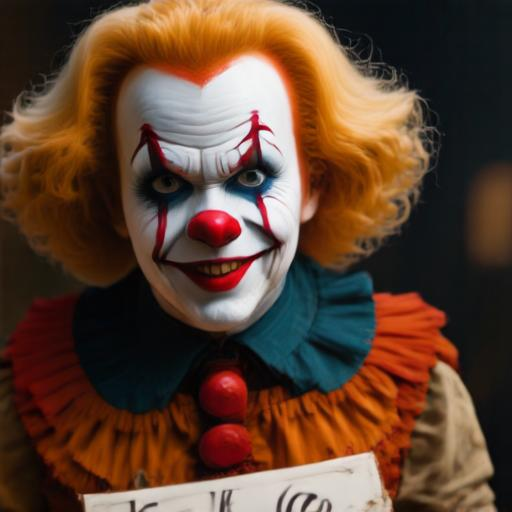}
    \end{minipage}%
    \begin{minipage}{0.20\textwidth}
        \includegraphics[width=\linewidth]{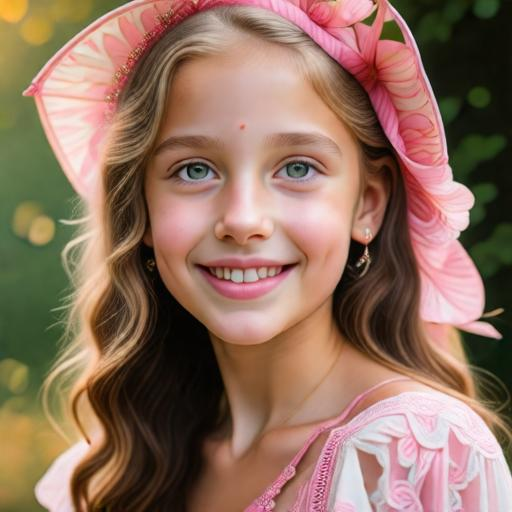}
    \end{minipage}%
    \begin{minipage}{0.2\textwidth}
        \includegraphics[width=\linewidth]{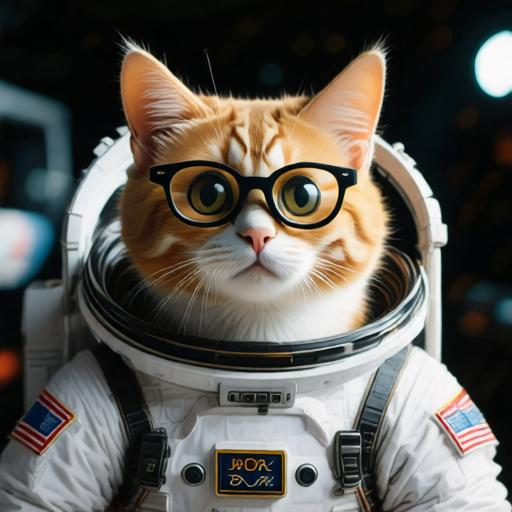}
    \end{minipage}%
    \begin{minipage}{0.2\textwidth}
        \includegraphics[width=\linewidth]{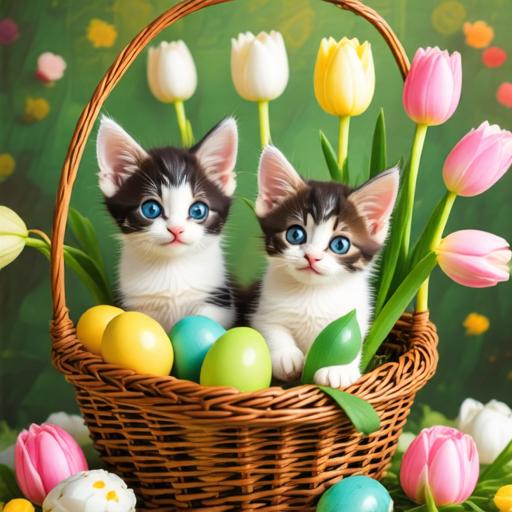}
    \end{minipage}%
        \begin{minipage}{0.20\textwidth}
        \includegraphics[width=\linewidth]{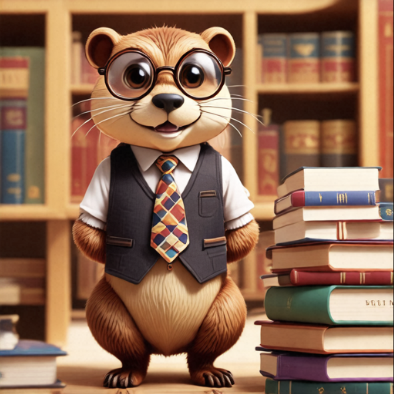}
    \end{minipage}%
\vspace{0cm}
    \begin{minipage}{0.20\textwidth}
        \includegraphics[width=\linewidth]{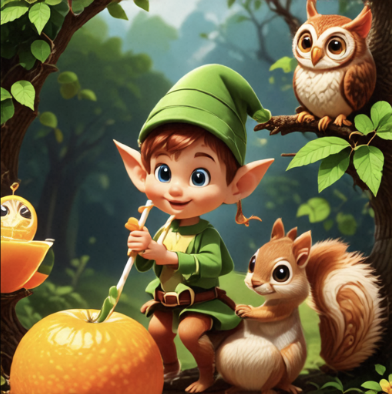}
    \end{minipage}%
    \begin{minipage}{0.20\textwidth}
        \includegraphics[width=\linewidth]{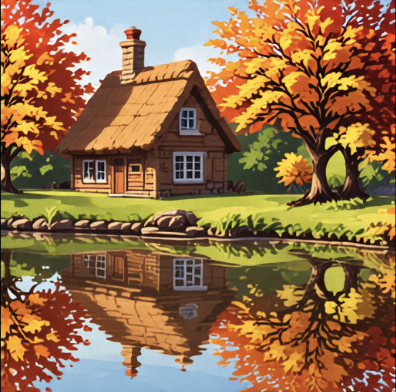}
    \end{minipage}%
    \begin{minipage}{0.2\textwidth}
        \includegraphics[width=\linewidth]{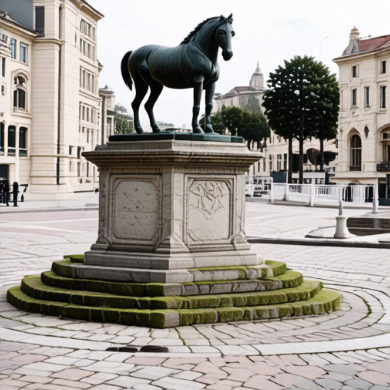}
    \end{minipage}%
    \begin{minipage}{0.2\textwidth}
        \includegraphics[width=\linewidth]{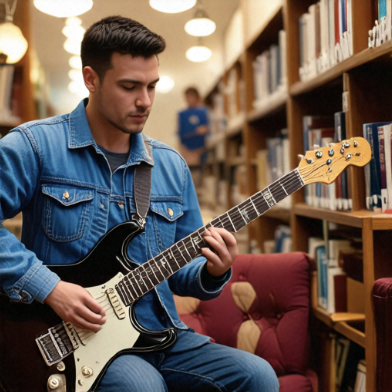}
    \end{minipage}%
        \begin{minipage}{0.20\textwidth}
        \includegraphics[width=\linewidth]{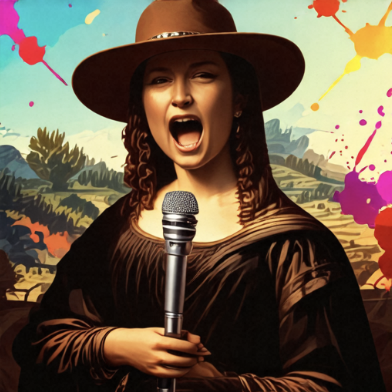}
    \end{minipage}%
    \caption{Visual generations produced by our AdvDMD method under 4 sampling steps on SD3.5. All results are obtained without applying classifier-free guidance~\citep{ho2021classifier}.}
    \label{fig:show_qwen}
\end{figure}

\begin{figure}[H]
    \centering
    \begin{minipage}{0.20\textwidth}
        \includegraphics[width=\linewidth]{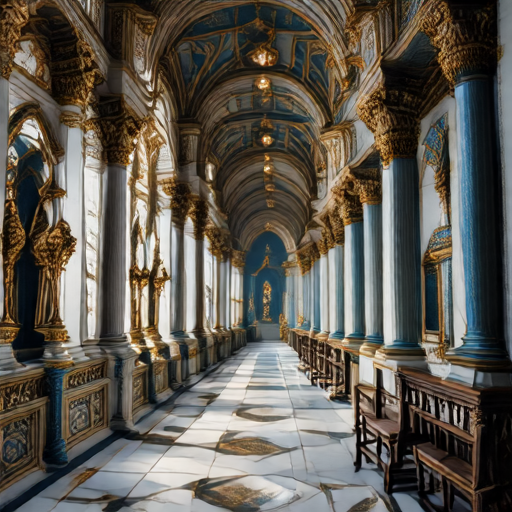}
    \end{minipage}%
    \begin{minipage}{0.20\textwidth}
        \includegraphics[width=\linewidth]{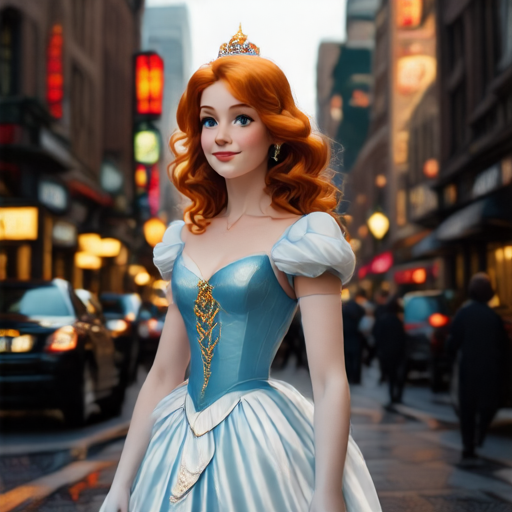}
    \end{minipage}%
    \begin{minipage}{0.20\textwidth}
        \includegraphics[width=\linewidth]{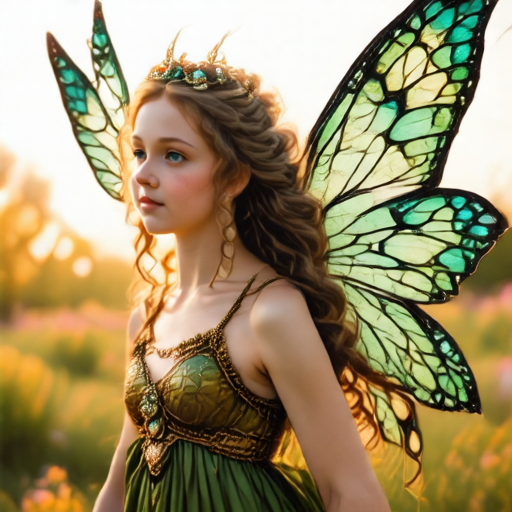}
    \end{minipage}%
    \begin{minipage}{0.20\textwidth}
        \includegraphics[width=\linewidth]{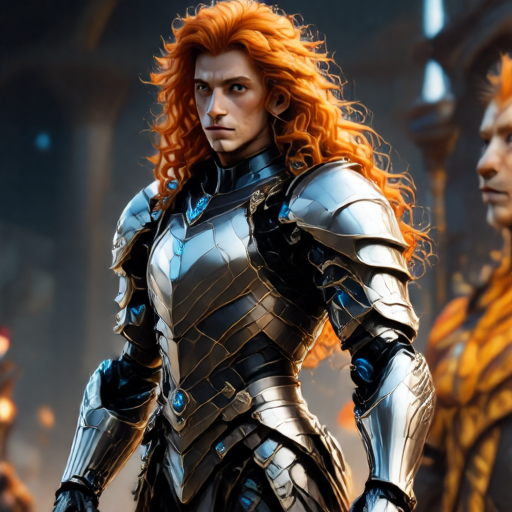}
    \end{minipage}%
    \begin{minipage}{0.20\textwidth}
        \includegraphics[width=\linewidth]{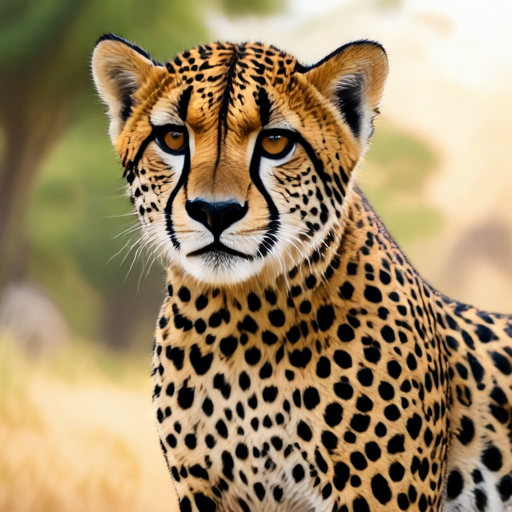}
    \end{minipage}%

    \begin{minipage}{0.2\textwidth}
        \includegraphics[width=\linewidth]{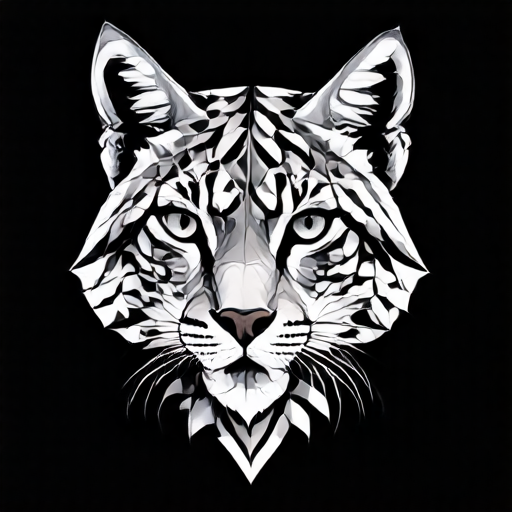}
    \end{minipage}%
    \begin{minipage}{0.2\textwidth}
        \includegraphics[width=\linewidth]{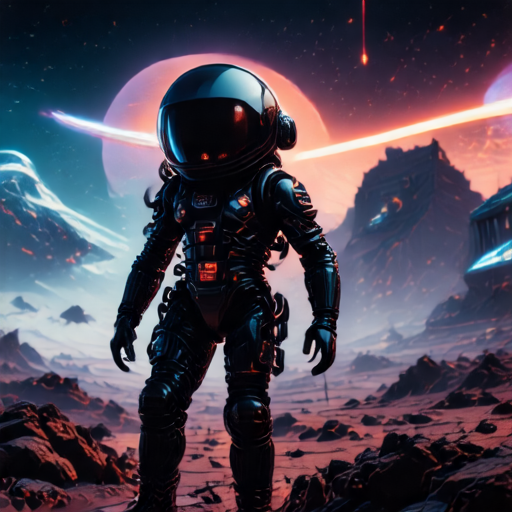}
    \end{minipage}%
    \begin{minipage}{0.2\textwidth}
        \includegraphics[width=\linewidth]{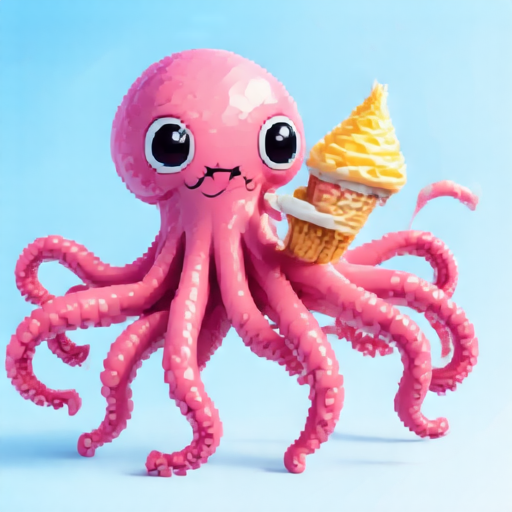}
    \end{minipage}%
    \begin{minipage}{0.2\textwidth}
        \includegraphics[width=\linewidth]{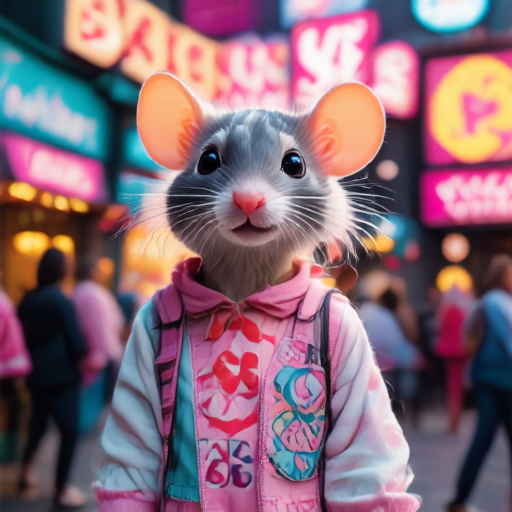}
    \end{minipage}%
    \begin{minipage}{0.20\textwidth}
        \includegraphics[width=\linewidth]{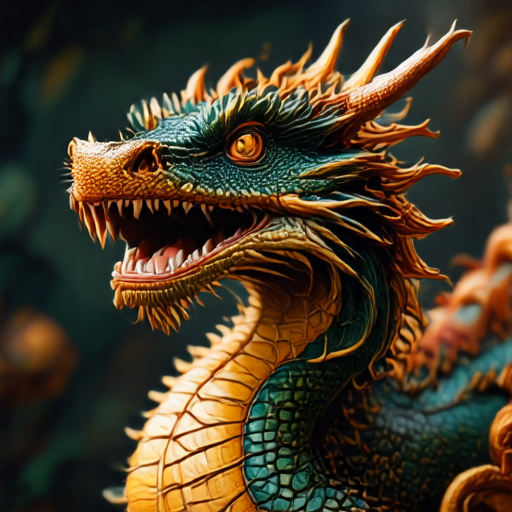}
    \end{minipage}%

    \begin{minipage}{0.2\textwidth}
        \includegraphics[width=\linewidth]{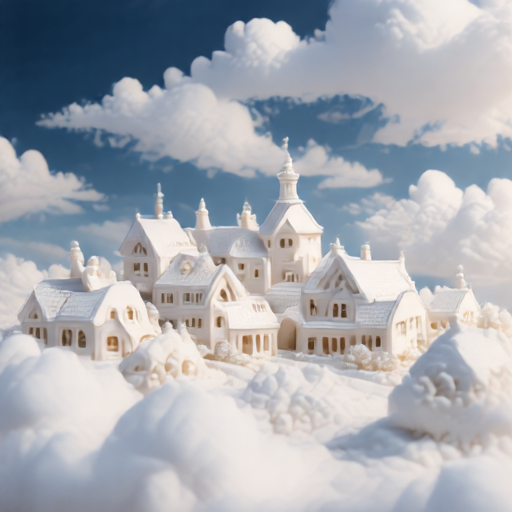}
    \end{minipage}%
    \begin{minipage}{0.2\textwidth}
        \includegraphics[width=\linewidth]{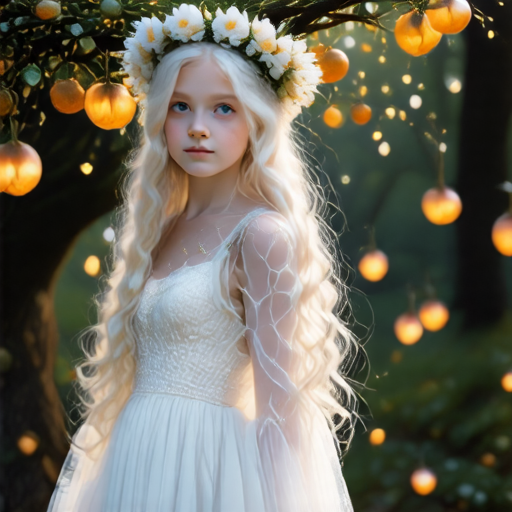}
    \end{minipage}%
    \begin{minipage}{0.2\textwidth}
        \includegraphics[width=\linewidth]{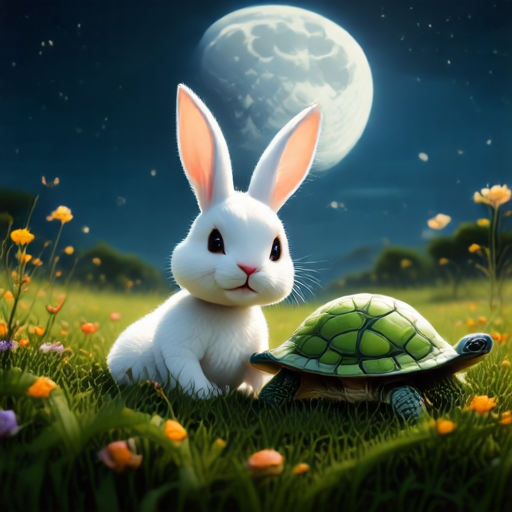}
    \end{minipage}%
    \begin{minipage}{0.2\textwidth}
        \includegraphics[width=\linewidth]{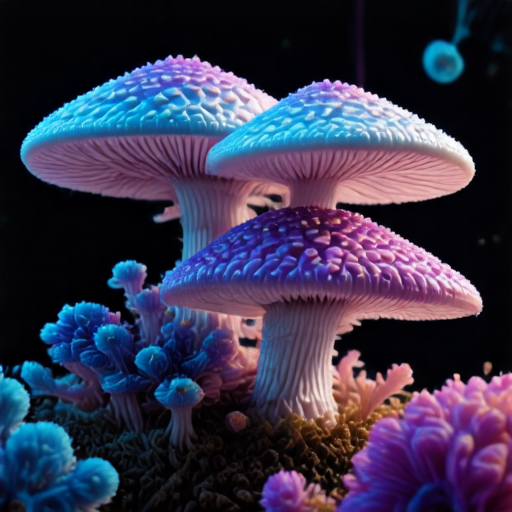}
    \end{minipage}%
    \begin{minipage}{0.20\textwidth}
        \includegraphics[width=\linewidth]{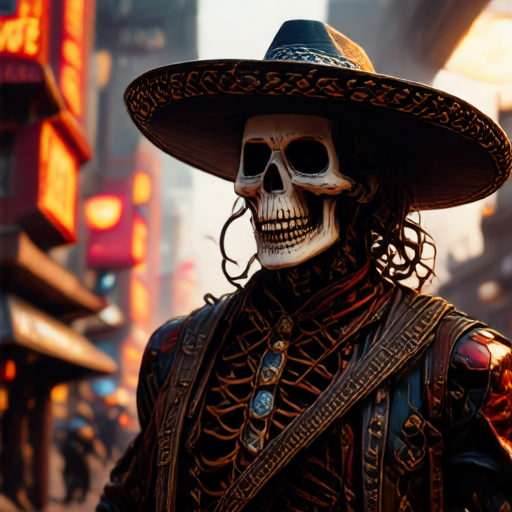}
    \end{minipage}%
    \vspace{0cm}
    \begin{minipage}{0.20\textwidth}
        \includegraphics[width=\linewidth]{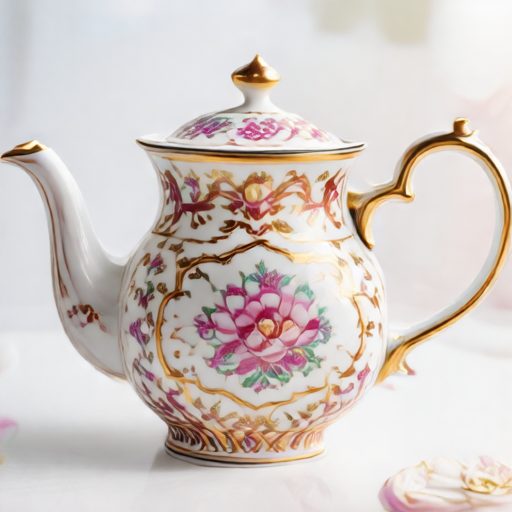}
    \end{minipage}%
    \begin{minipage}{0.20\textwidth}
        \includegraphics[width=\linewidth]{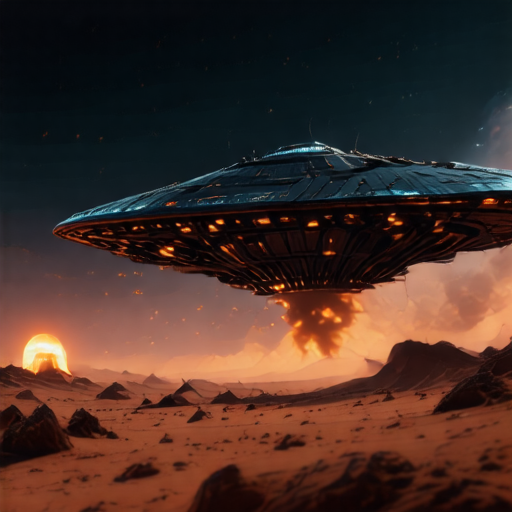}
    \end{minipage}%
    \begin{minipage}{0.2\textwidth}
        \includegraphics[width=\linewidth]{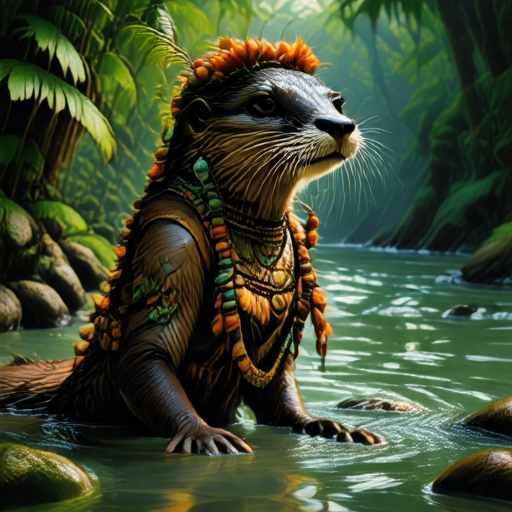}
    \end{minipage}%
    \begin{minipage}{0.2\textwidth}
        \includegraphics[width=\linewidth]{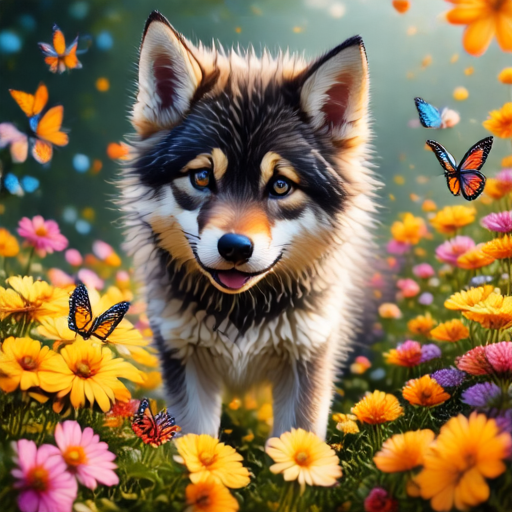}
    \end{minipage}%
        \begin{minipage}{0.20\textwidth}
        \includegraphics[width=\linewidth]{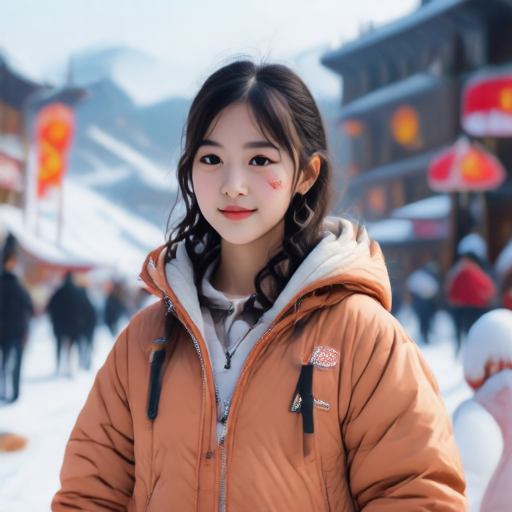}
    \end{minipage}%
\vspace{0cm}
    \begin{minipage}{0.20\textwidth}
        \includegraphics[width=\linewidth]{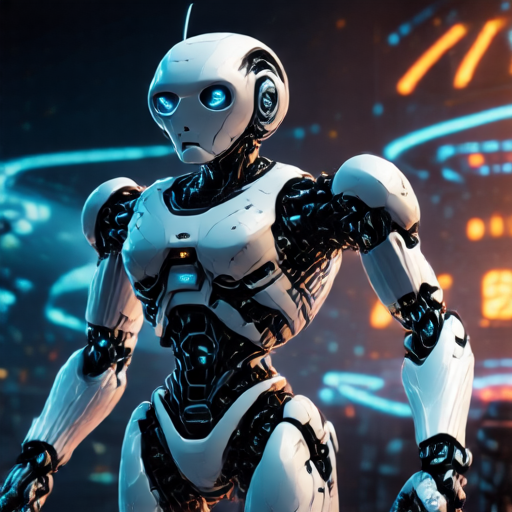}
    \end{minipage}%
    \begin{minipage}{0.20\textwidth}
        \includegraphics[width=\linewidth]{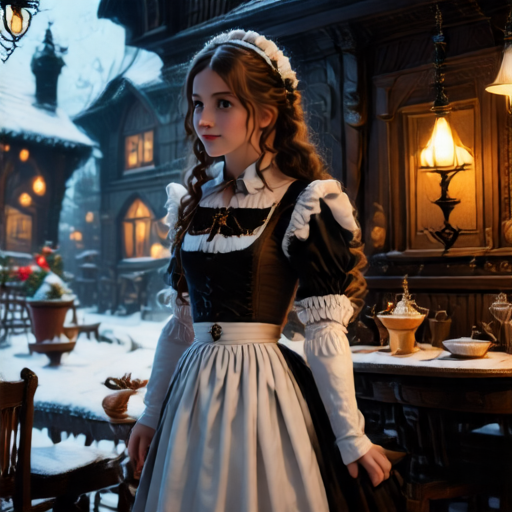}
    \end{minipage}%
    \begin{minipage}{0.2\textwidth}
        \includegraphics[width=\linewidth]{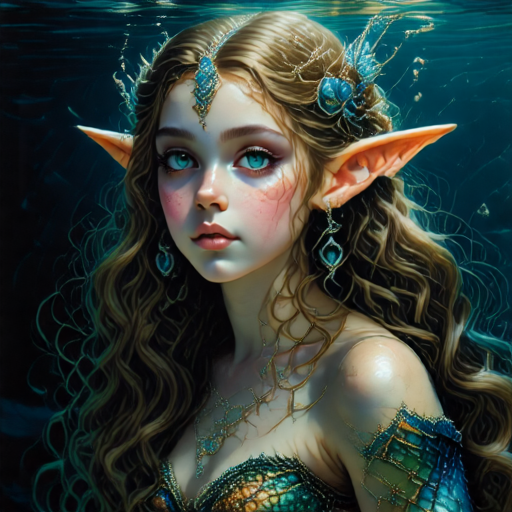}
    \end{minipage}%
    \begin{minipage}{0.2\textwidth}
        \includegraphics[width=\linewidth]{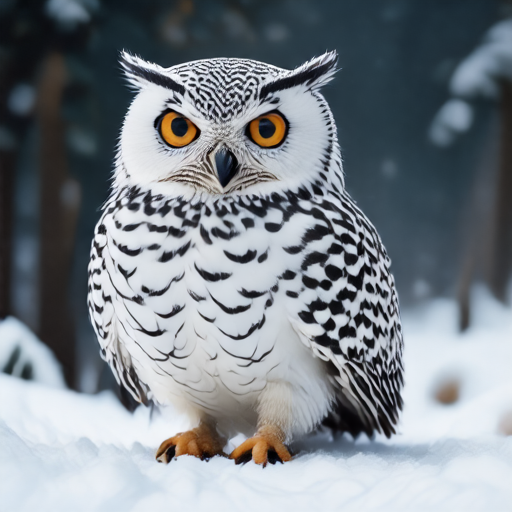}
    \end{minipage}%
        \begin{minipage}{0.20\textwidth}
        \includegraphics[width=\linewidth]{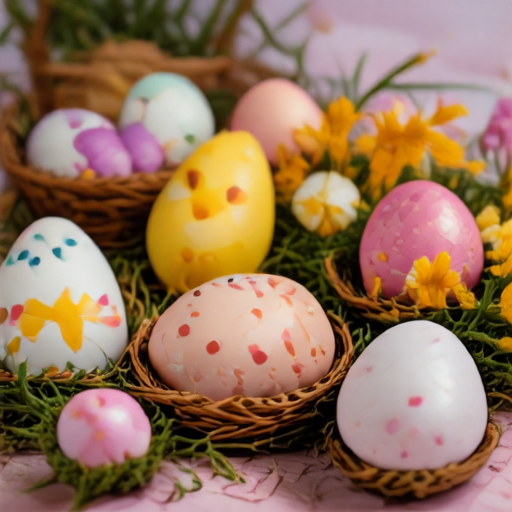}
    \end{minipage}%
    \caption{Visual generations produced by our AdvDMD method under 2 sampling steps. All results are obtained without applying classifier-free guidance~\citep{ho2021classifier}.}
    \label{fig:show_2}
\end{figure}
\begin{figure}[H]
    \centering
    \begin{minipage}{0.20\textwidth}
        \includegraphics[width=\linewidth]{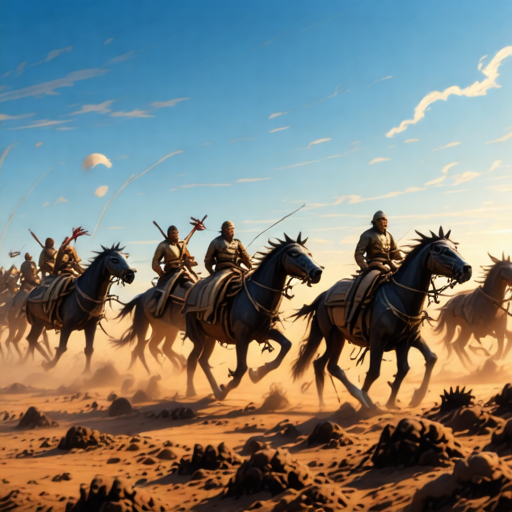}
    \end{minipage}%
    \begin{minipage}{0.20\textwidth}
        \includegraphics[width=\linewidth]{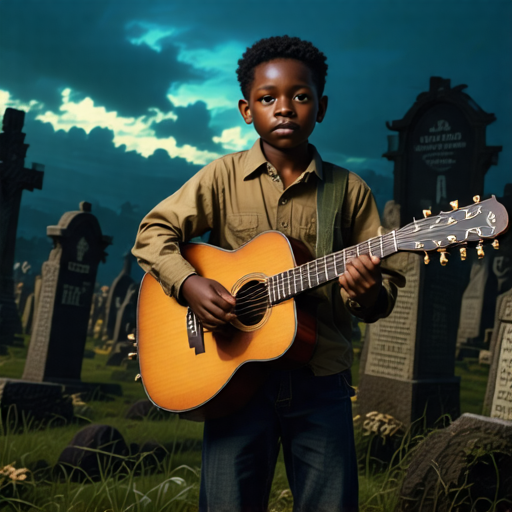}
    \end{minipage}%
    \begin{minipage}{0.20\textwidth}
        \includegraphics[width=\linewidth]{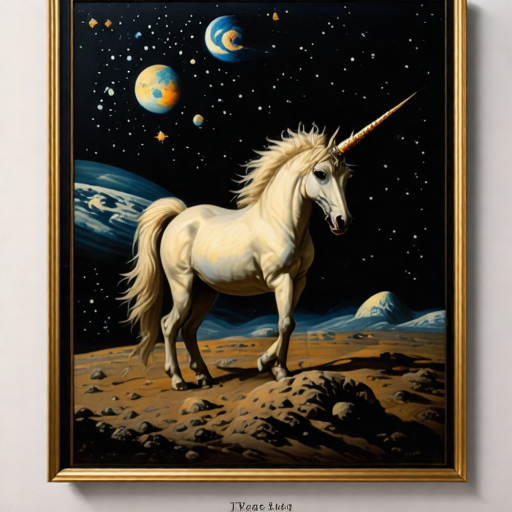}
    \end{minipage}%
    \begin{minipage}{0.20\textwidth}
        \includegraphics[width=\linewidth]{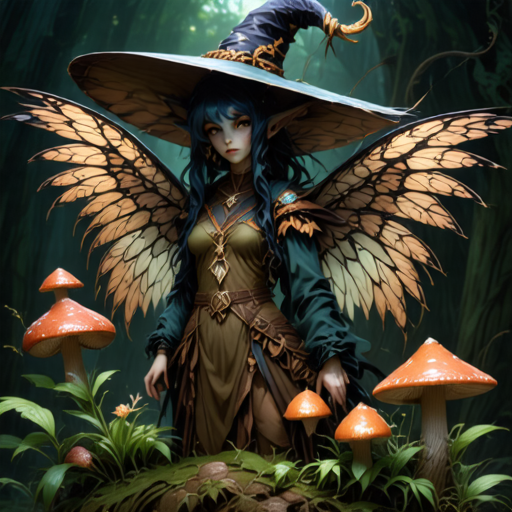}
    \end{minipage}%
    \begin{minipage}{0.20\textwidth}
        \includegraphics[width=\linewidth]{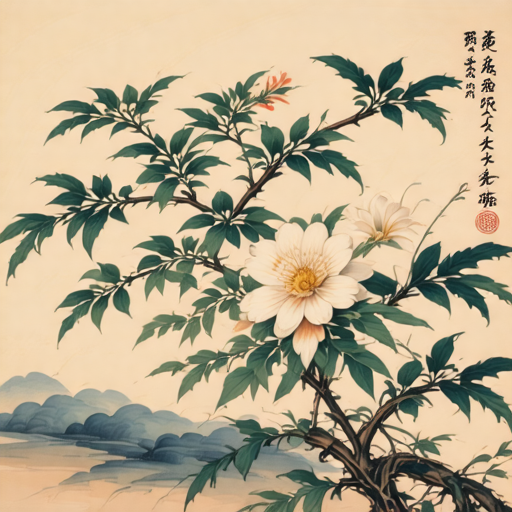}
    \end{minipage}%

    \begin{minipage}{0.2\textwidth}
        \includegraphics[width=\linewidth]{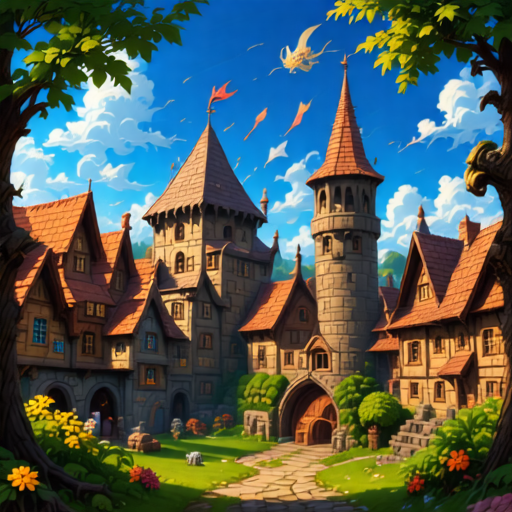}
    \end{minipage}%
    \begin{minipage}{0.2\textwidth}
        \includegraphics[width=\linewidth]{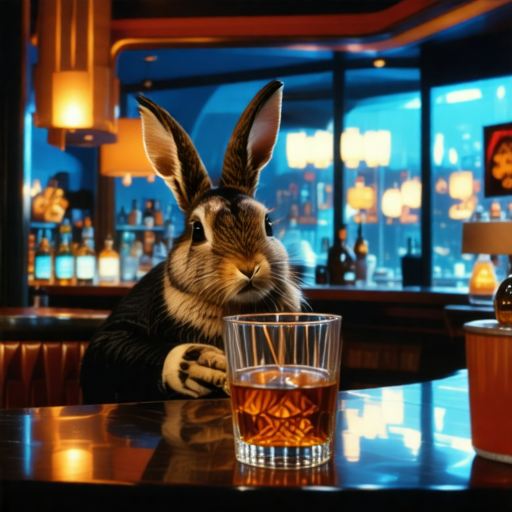}
    \end{minipage}%
    \begin{minipage}{0.2\textwidth}
        \includegraphics[width=\linewidth]{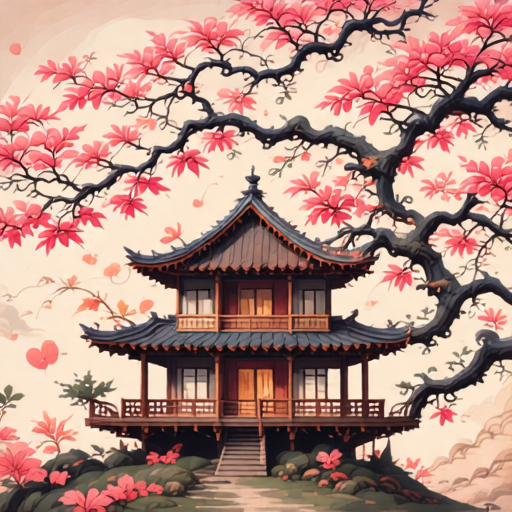}
    \end{minipage}%
    \begin{minipage}{0.2\textwidth}
        \includegraphics[width=\linewidth]{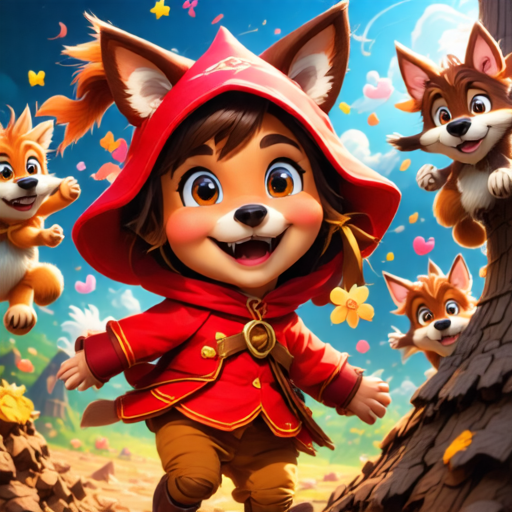}
    \end{minipage}%
    \begin{minipage}{0.20\textwidth}
        \includegraphics[width=\linewidth]{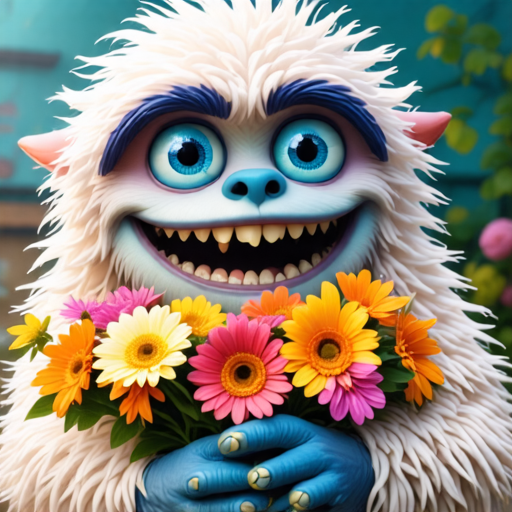}
    \end{minipage}%

    \begin{minipage}{0.2\textwidth}
        \includegraphics[width=\linewidth]{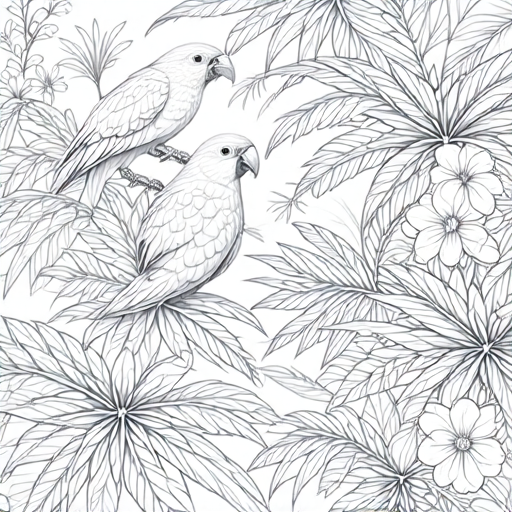}
    \end{minipage}%
    \begin{minipage}{0.2\textwidth}
        \includegraphics[width=\linewidth]{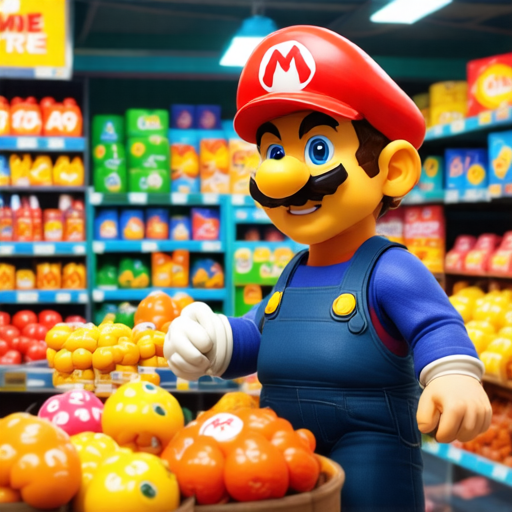}
    \end{minipage}%
    \begin{minipage}{0.2\textwidth}
        \includegraphics[width=\linewidth]{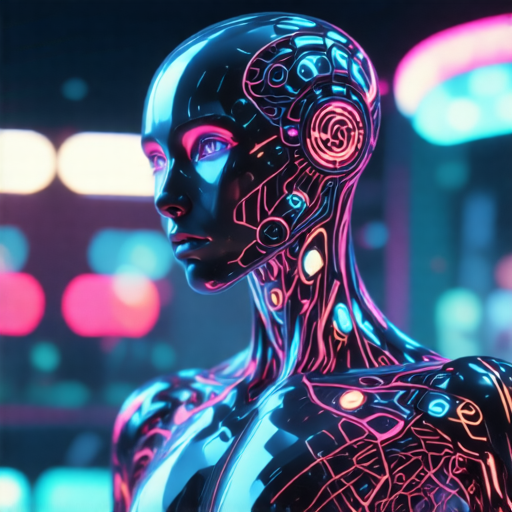}
    \end{minipage}%
    \begin{minipage}{0.2\textwidth}
        \includegraphics[width=\linewidth]{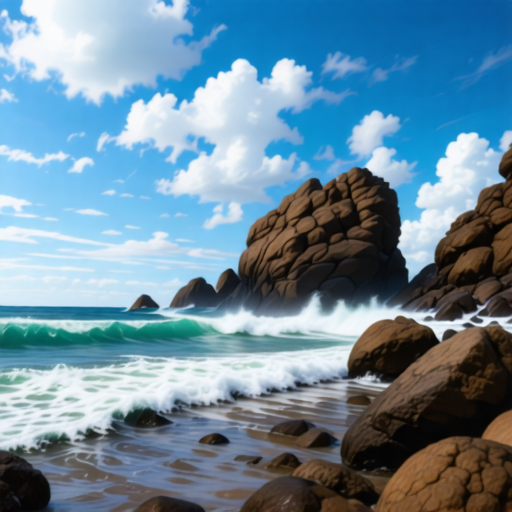}
    \end{minipage}%
    \begin{minipage}{0.20\textwidth}
        \includegraphics[width=\linewidth]{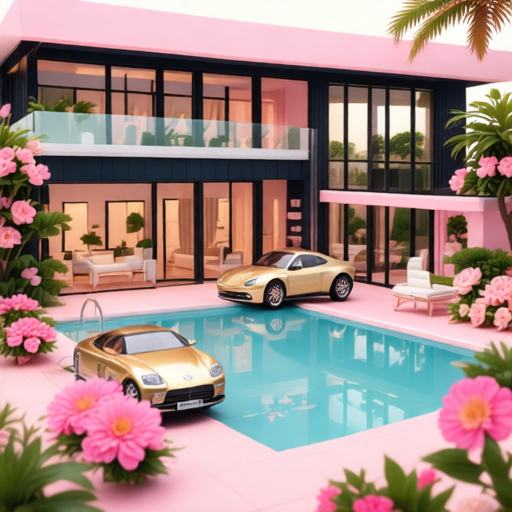}
    \end{minipage}%
    \vspace{0cm}
    \begin{minipage}{0.20\textwidth}
        \includegraphics[width=\linewidth]{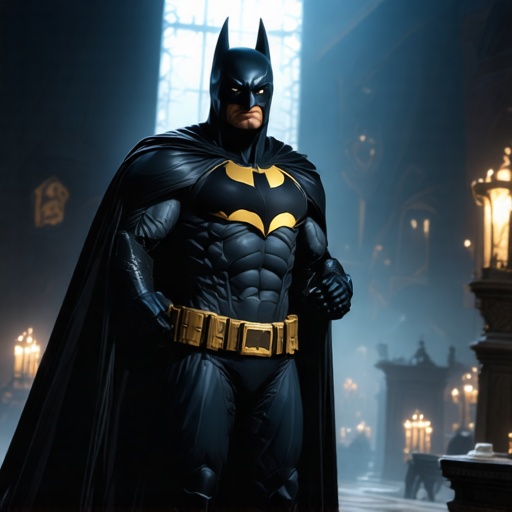}
    \end{minipage}%
    \begin{minipage}{0.20\textwidth}
        \includegraphics[width=\linewidth]{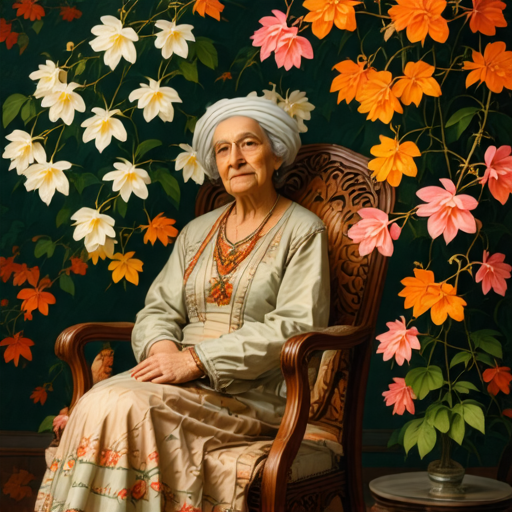}
    \end{minipage}%
    \begin{minipage}{0.2\textwidth}
        \includegraphics[width=\linewidth]{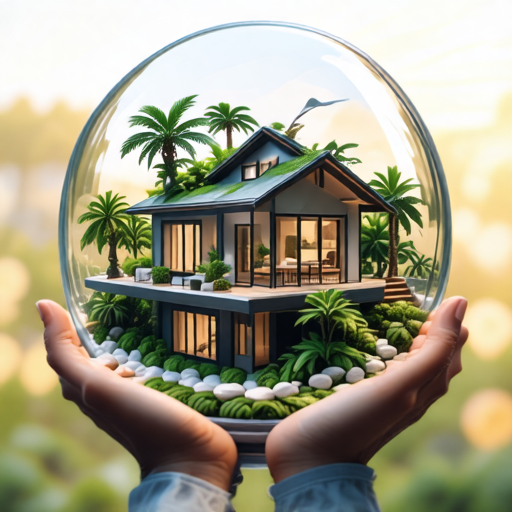}
    \end{minipage}%
    \begin{minipage}{0.2\textwidth}
        \includegraphics[width=\linewidth]{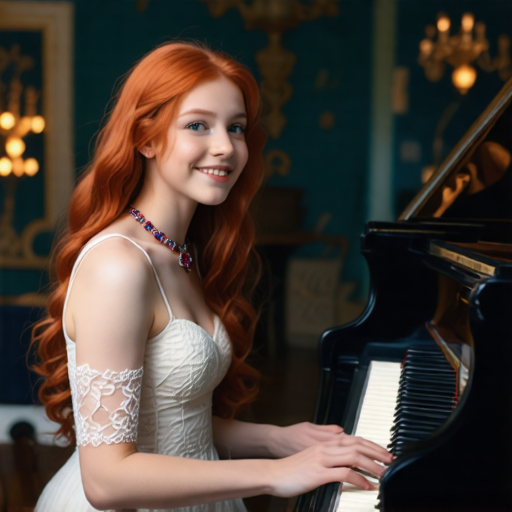}
    \end{minipage}%
        \begin{minipage}{0.20\textwidth}
        \includegraphics[width=\linewidth]{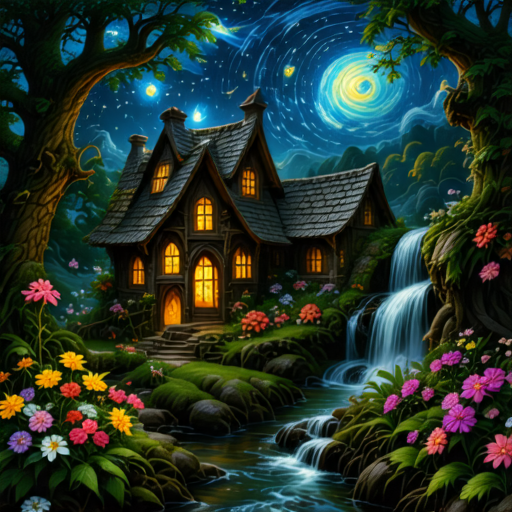}
    \end{minipage}%
\vspace{0cm}
    \begin{minipage}{0.20\textwidth}
        \includegraphics[width=\linewidth]{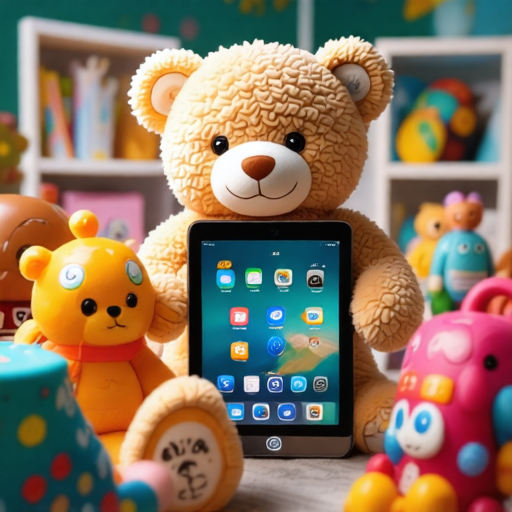}
    \end{minipage}%
    \begin{minipage}{0.20\textwidth}
        \includegraphics[width=\linewidth]{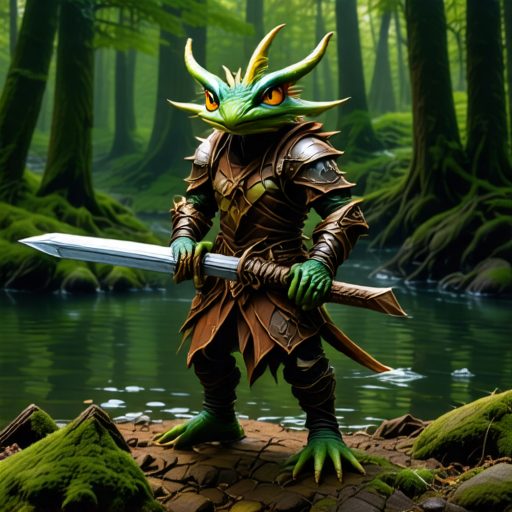}
    \end{minipage}%
    \begin{minipage}{0.2\textwidth}
        \includegraphics[width=\linewidth]{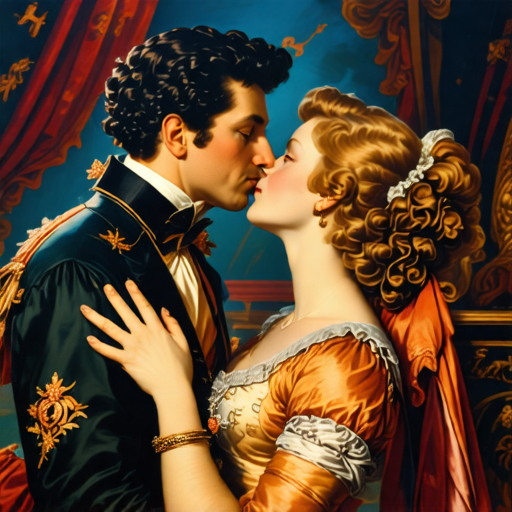}
    \end{minipage}%
    \begin{minipage}{0.2\textwidth}
        \includegraphics[width=\linewidth]{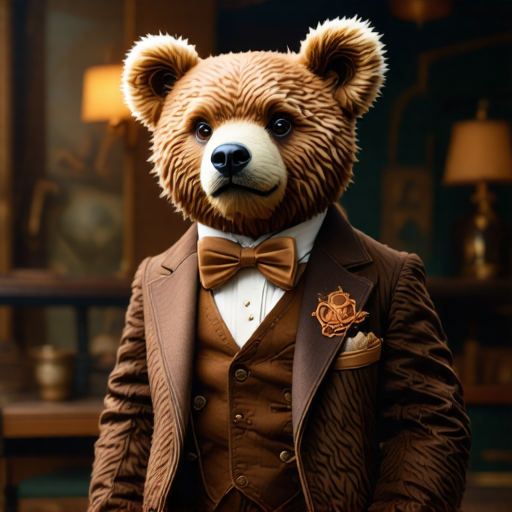}
    \end{minipage}%
        \begin{minipage}{0.20\textwidth}
        \includegraphics[width=\linewidth]{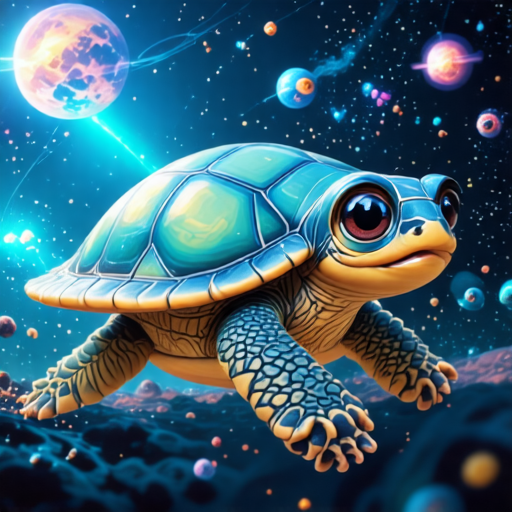}
    \end{minipage}%
    \caption{Visual generations produced by our AdvDMD method under 4 sampling steps. All results are obtained without applying classifier-free guidance~\citep{ho2021classifier}.}
    \label{fig:show_more_4}
\end{figure}
\begin{figure}[H]
    \centering
    \begin{minipage}{0.25\textwidth}  
        \centering
        \textbf{TDM}  
    \end{minipage}%
    \begin{minipage}{0.25\textwidth}  
               \centering
        \textbf{Flash}
    \end{minipage}%
    \begin{minipage}{0.25\textwidth}
                      \centering
        \textbf{Hyper-SD}
    \end{minipage}%
    \begin{minipage}{0.25\textwidth}
           \centering
        \textbf{AdvDMD}
    \end{minipage}%
\vspace{0.1cm}
    \begin{minipage}{0.25\textwidth}
     \centering
        \includegraphics[width=0.8\linewidth]{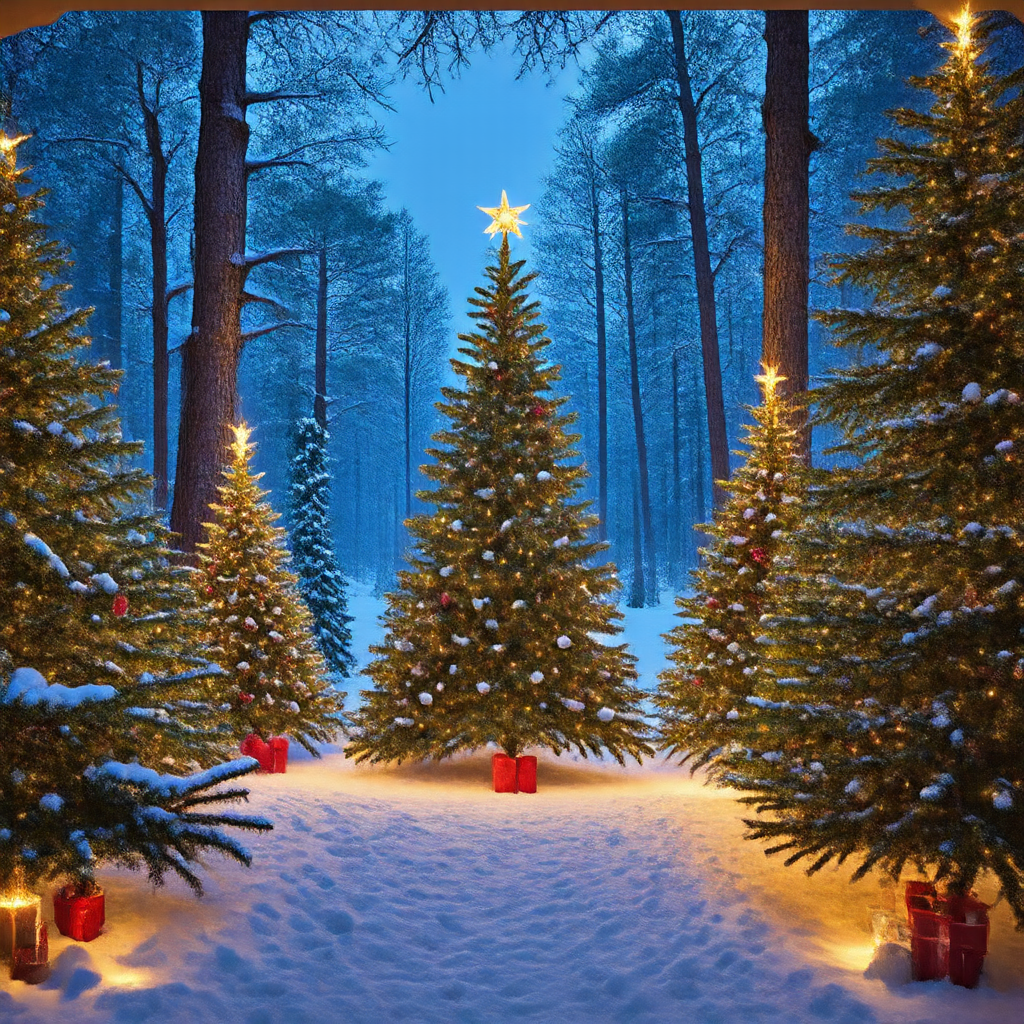}
    \end{minipage}%
    \begin{minipage}{0.25\textwidth}
     \centering
        \includegraphics[width=0.8\linewidth]{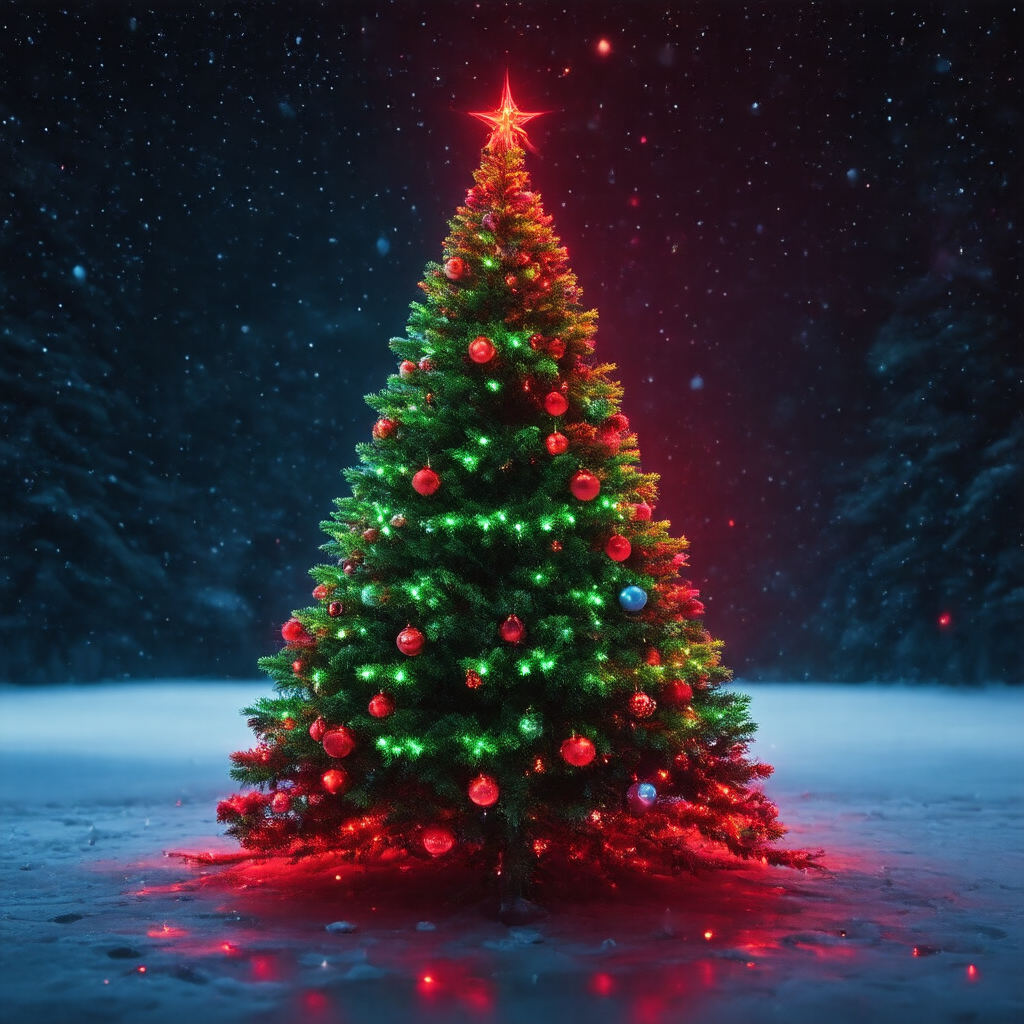}
    \end{minipage}%
    \begin{minipage}{0.25\textwidth}
     \centering
        \includegraphics[width=0.8\linewidth]{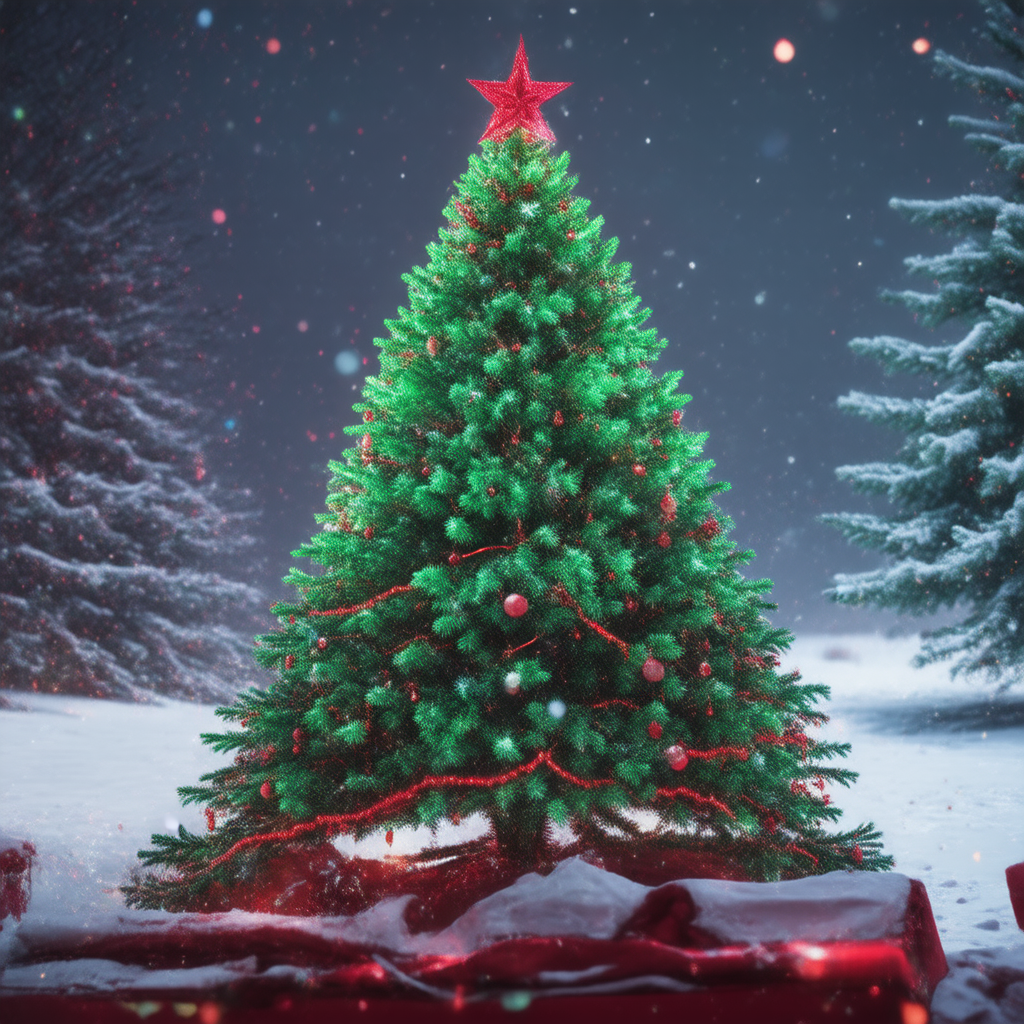}
    \end{minipage}%
    \begin{minipage}{0.25\textwidth}
     \centering
        \includegraphics[width=0.8\linewidth]{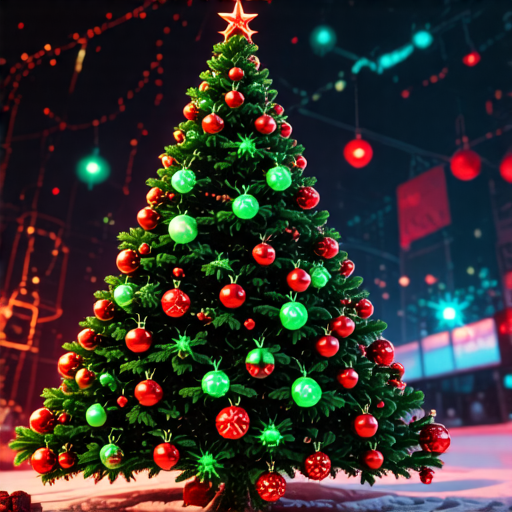}
    \end{minipage}%
\vspace{0.1cm}
\begin{minipage}[c]{\textwidth}
  \setlength{\parskip}{0pt} 
  \setlength{\parindent}{0pt} 
  \centering
  \vspace{0pt} 
  \textbf{cyberpunk christmas tree portrait, red and green bright colors...}
  \vspace{0pt} 
\end{minipage}
      \begin{minipage}{0.25\textwidth}
       \centering
        \includegraphics[width=0.8\linewidth]{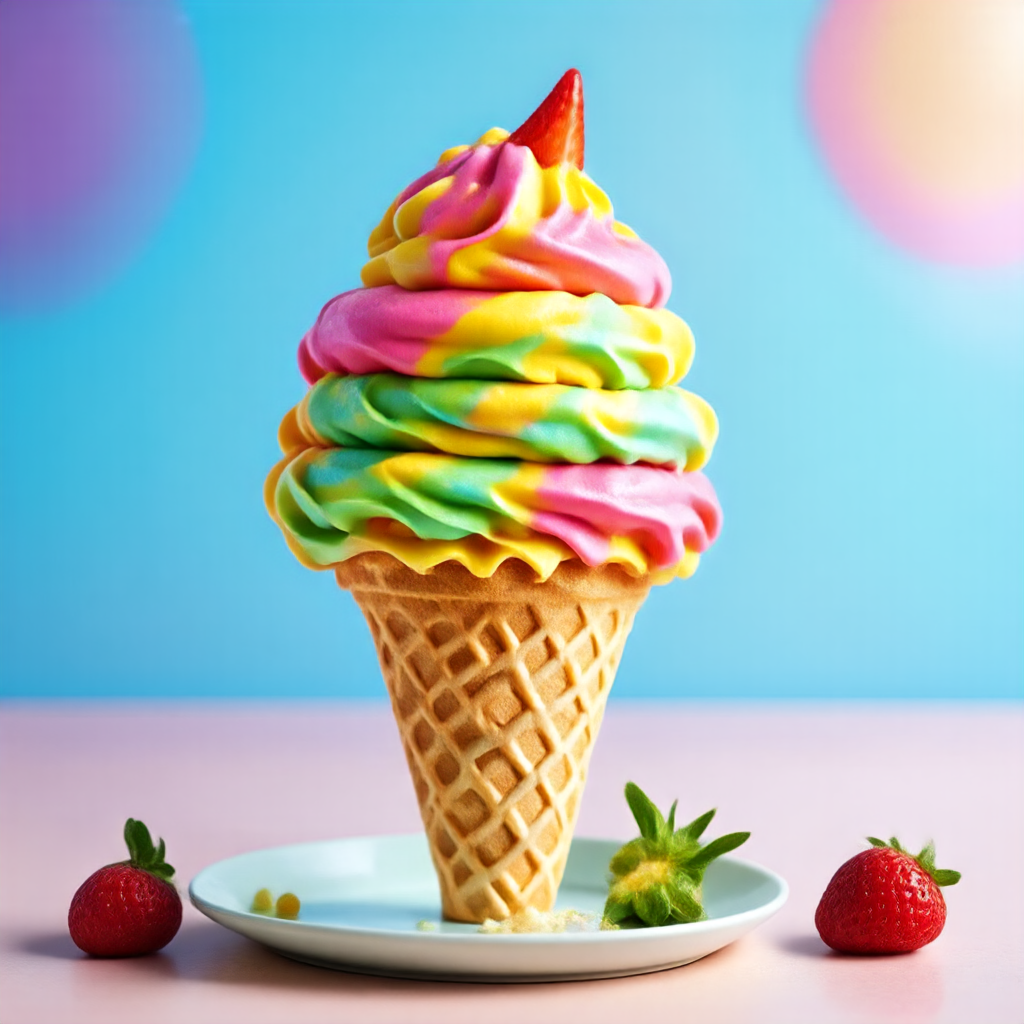}
    \end{minipage}%
    \begin{minipage}{0.25\textwidth}
     \centering
        \includegraphics[width=0.8\linewidth]{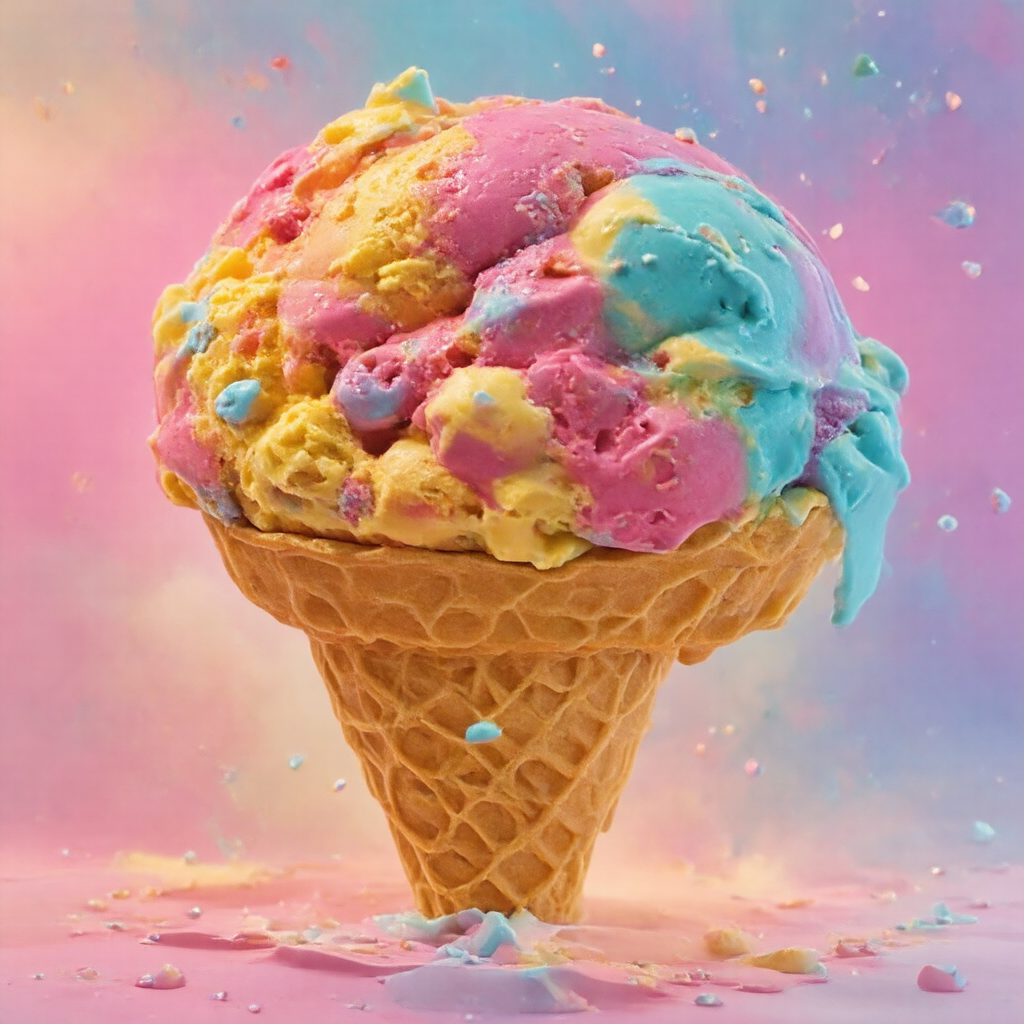}
    \end{minipage}%
    \begin{minipage}{0.25\textwidth}
     \centering
        \includegraphics[width=0.8\linewidth]{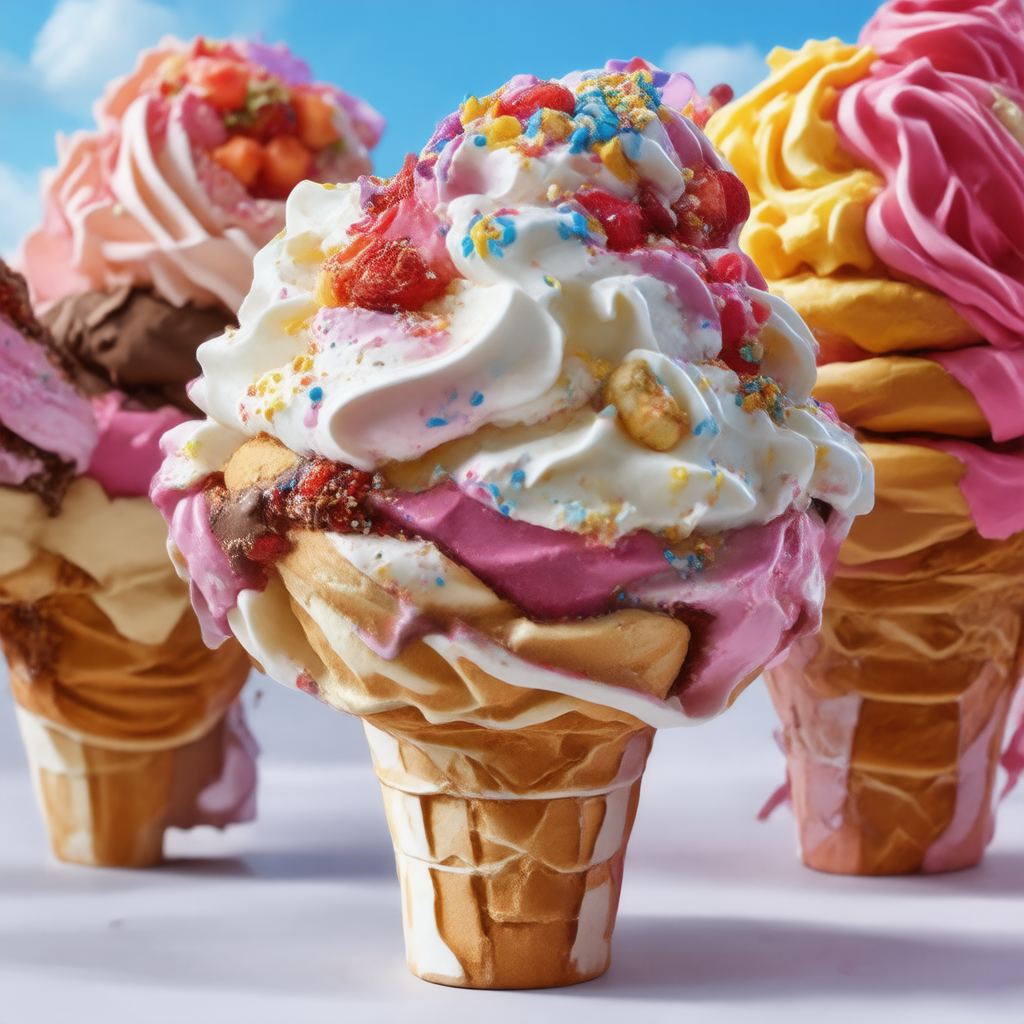}
    \end{minipage}%
    \begin{minipage}{0.25\textwidth}
     \centering
        \includegraphics[width=0.8\linewidth]{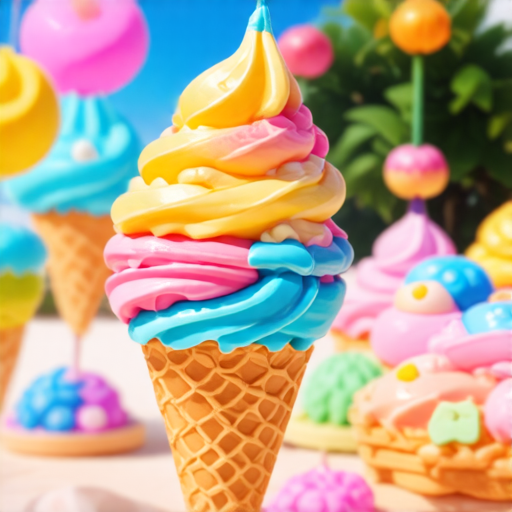}
    \end{minipage}%
    \vspace{0.1cm}
\begin{minipage}[c]{\textwidth}
  \setlength{\parskip}{0pt} 
  \setlength{\parindent}{0pt} 
  \centering
  \vspace{0pt} 
  \textbf{ice cream, ultra detailed, 8k cinematic, vivid colors, stylized, beautiful...}
  \vspace{0pt} 
\end{minipage}
      \begin{minipage}{0.25\textwidth}
       \centering
        \includegraphics[width=0.8\linewidth]{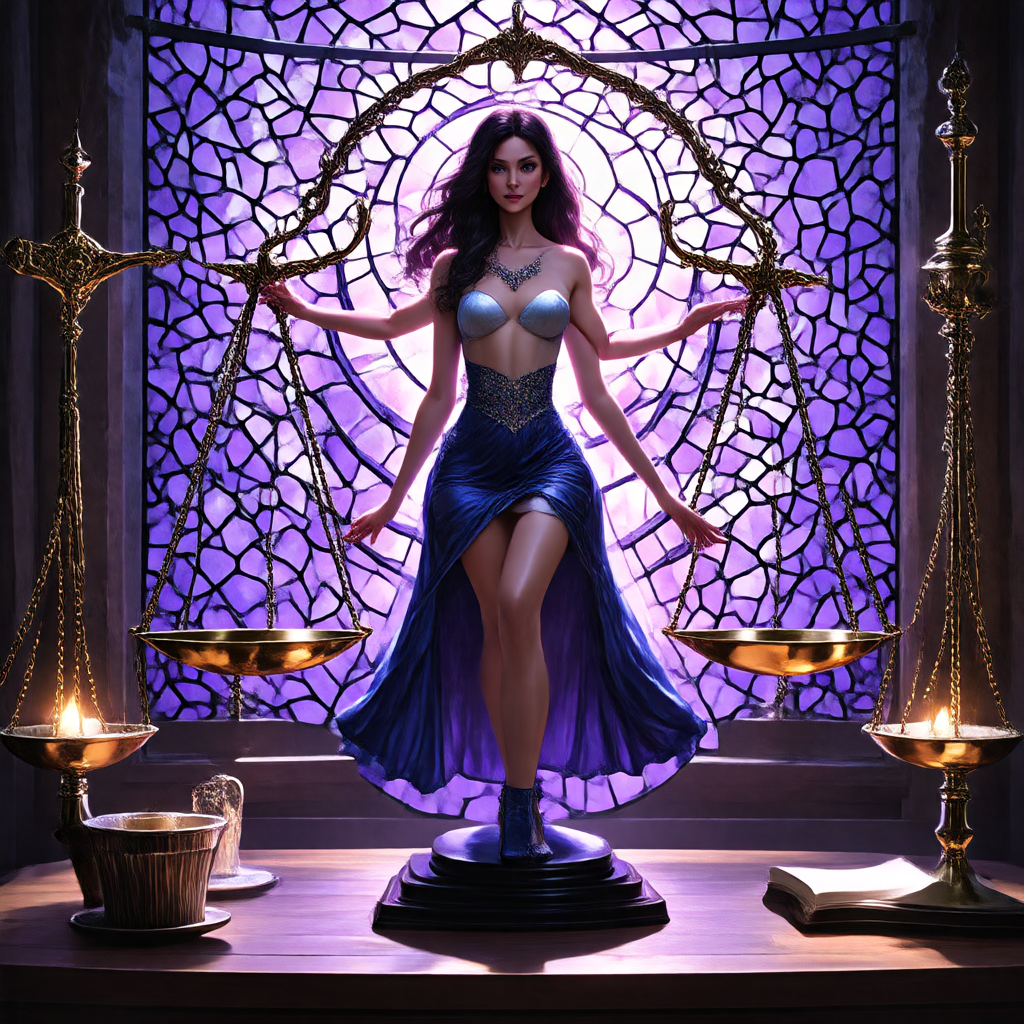}
    \end{minipage}%
    \begin{minipage}{0.25\textwidth}
     \centering
        \includegraphics[width=0.8\linewidth]{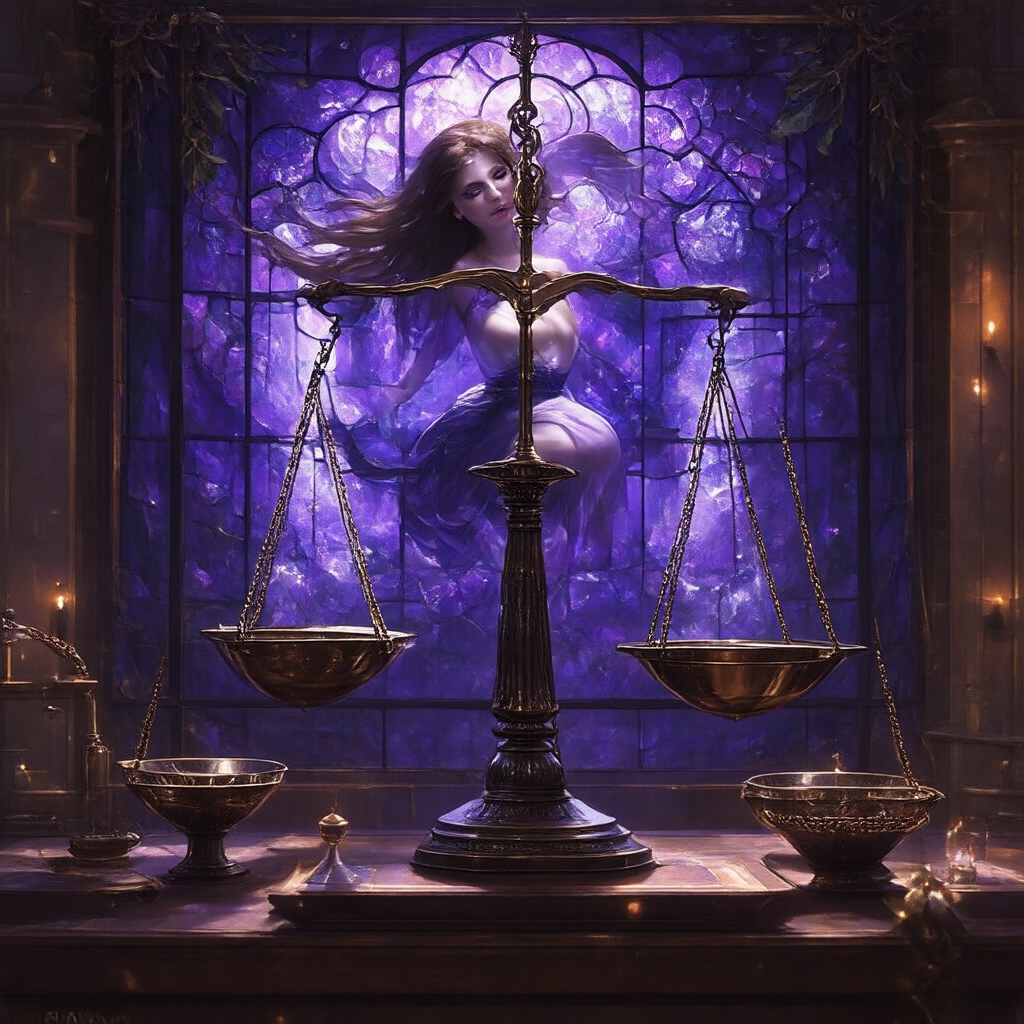}
    \end{minipage}%
    \begin{minipage}{0.25\textwidth}
     \centering
        \includegraphics[width=0.8\linewidth]{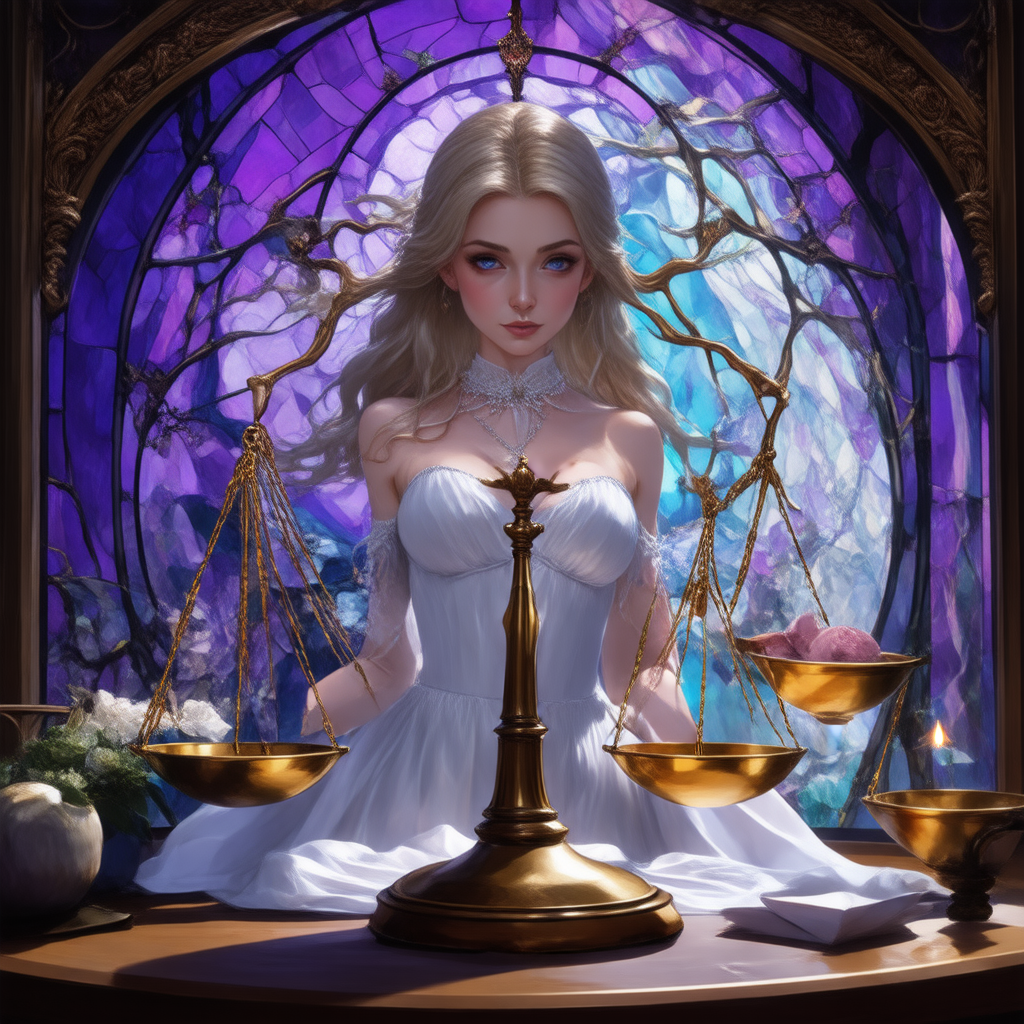}
    \end{minipage}%
    \begin{minipage}{0.25\textwidth}
     \centering
        \includegraphics[width=0.8\linewidth]{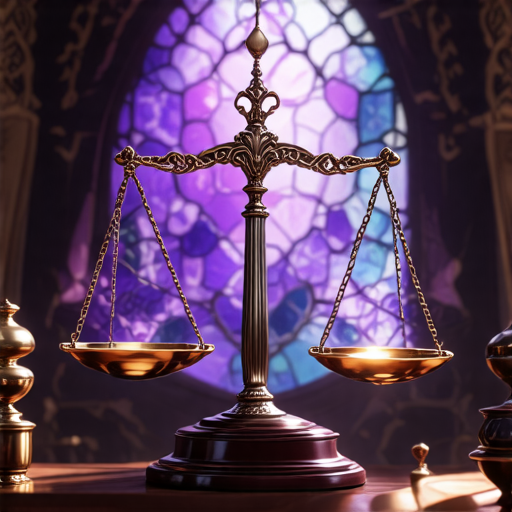}
    \end{minipage}%
   ‘\vspace{0.1cm}
\begin{minipage}[c]{\textwidth}
  \setlength{\parskip}{0pt} 
  \setlength{\parindent}{0pt} 
  \centering
  \vspace{0pt} 
  \textbf{scales and balance, scale, mystical, fantasy, highly detailed, stained glass purple royal tapestry in the background...}  \vspace{0pt} 
\end{minipage}

      \begin{minipage}{0.25\textwidth}
       \centering
        \includegraphics[width=0.8\linewidth]{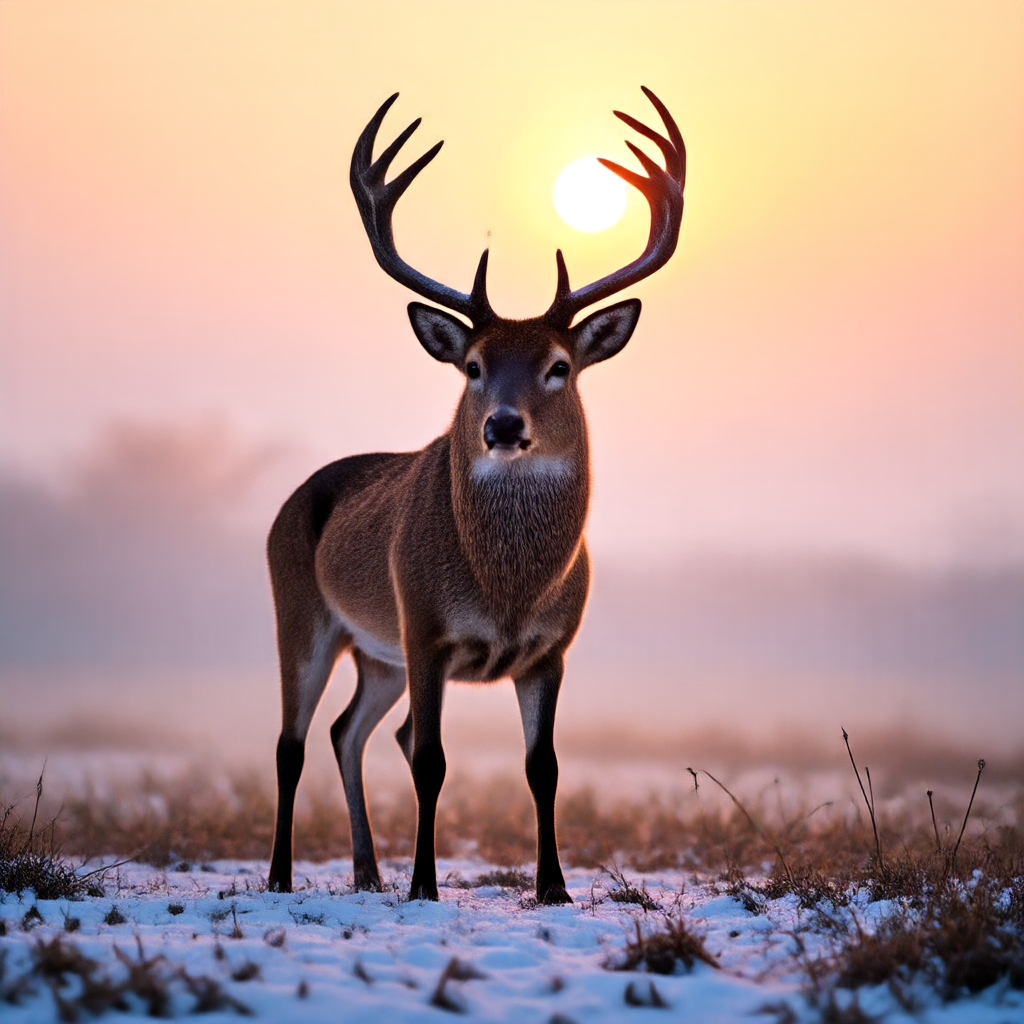}
    \end{minipage}%
    \begin{minipage}{0.25\textwidth}
     \centering
        \includegraphics[width=0.8\linewidth]{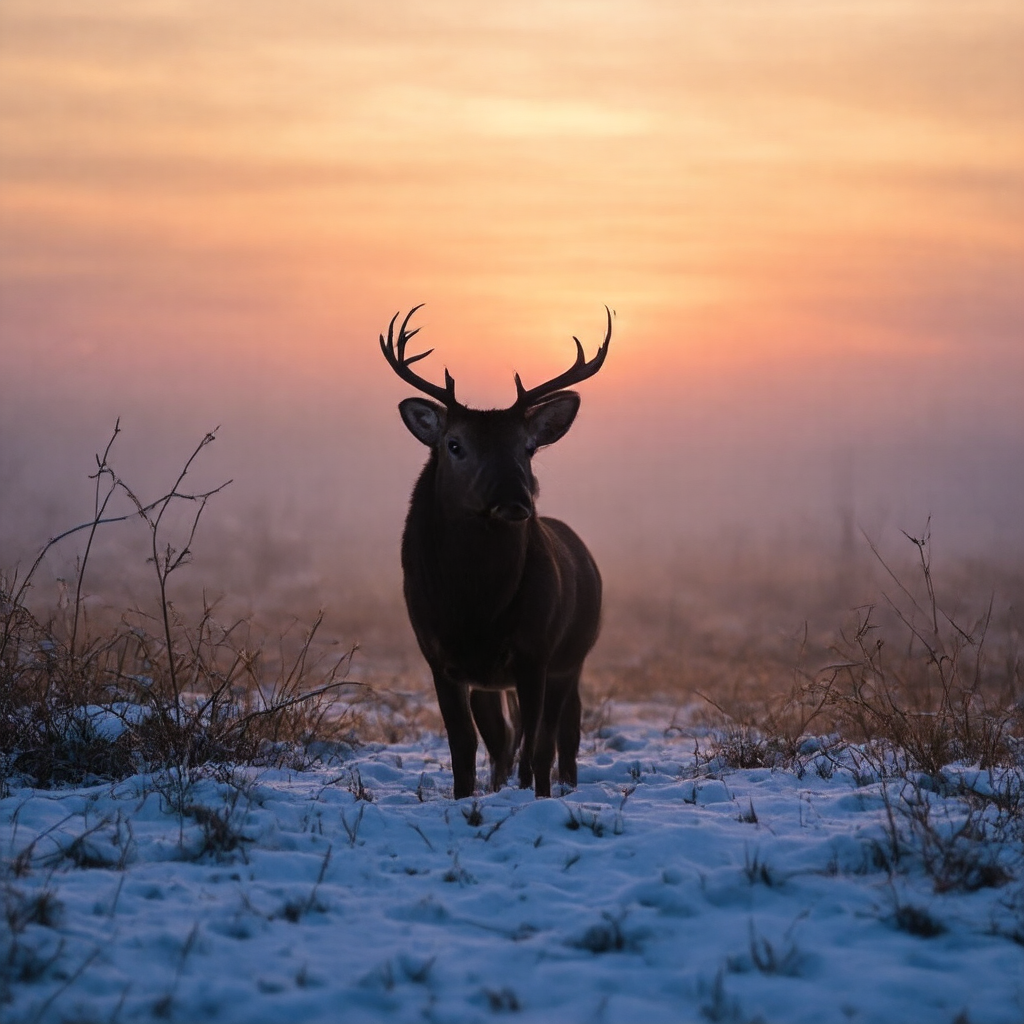}
    \end{minipage}%
    \begin{minipage}{0.25\textwidth}
     \centering
        \includegraphics[width=0.8\linewidth]{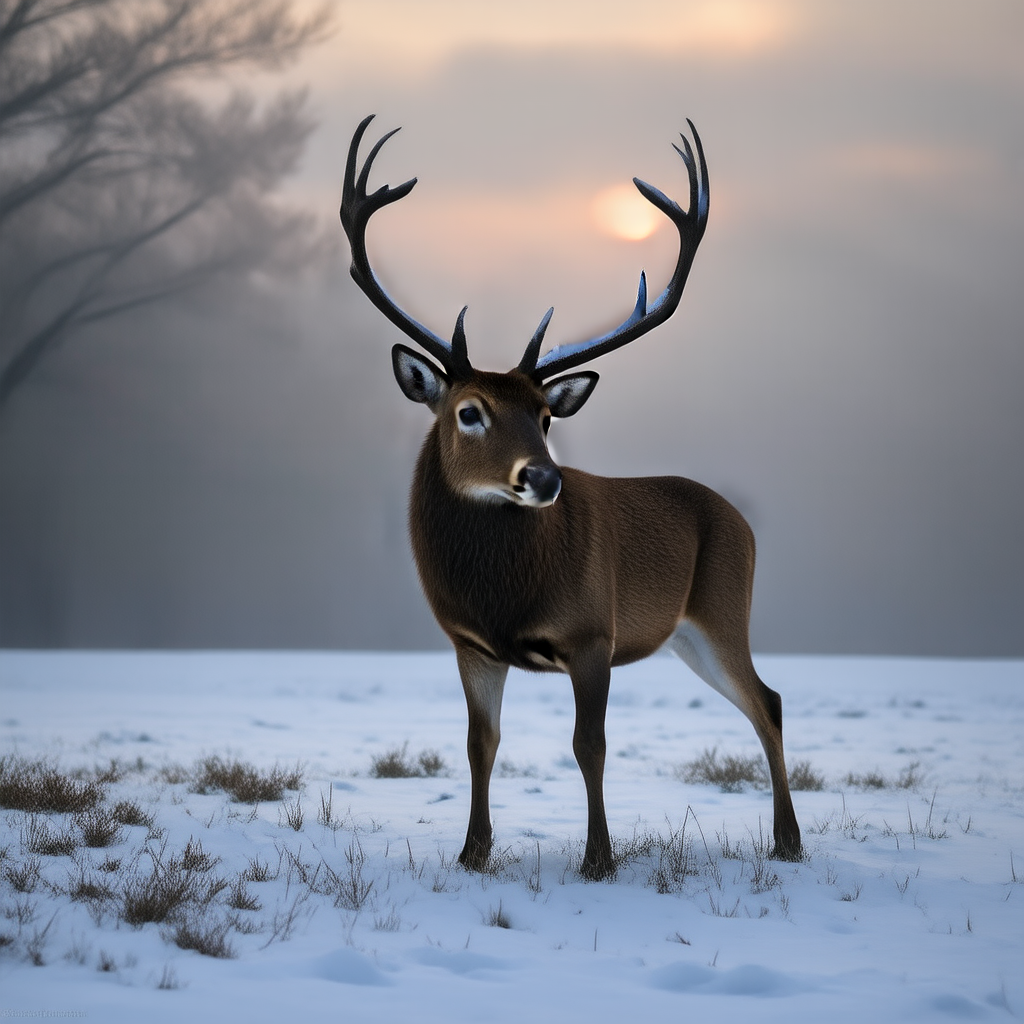}
    \end{minipage}%
    \begin{minipage}{0.25\textwidth}
     \centering
        \includegraphics[width=0.8\linewidth]{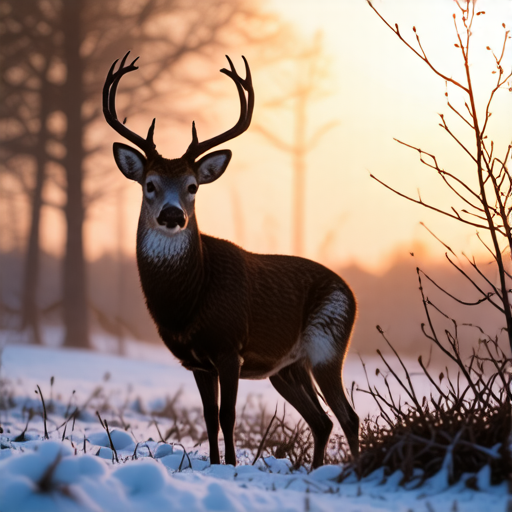}
    \end{minipage}%
    \vspace{0.1cm}
    \begin{minipage}[c]{\textwidth}
  \setlength{\parskip}{0pt} 
  \setlength{\parindent}{0pt} 
  \centering
  \vspace{0pt} 
  \textbf{buck silhouette, winter solstice sunset, snow in ground, mist and fog, strong light.}
  \vspace{0pt} 
\end{minipage}
    \vspace{0.1cm}
    \caption{Comparison between TDM, Flash, Hyper-SD and AdvDMD}
    \label{fig:compare_more}
\end{figure}

\end{document}